\documentclass[superscriptaddress]{article}

\usepackage[preprint]{corl_2022}
\usepackage{microtype}
\usepackage{graphicx}
\usepackage{caption}
\usepackage{subcaption}
\usepackage{booktabs} % for professional tables
\usepackage[export]{adjustbox}
\usepackage[wide]{sidecap}
\usepackage{array}
\usepackage{wrapfig,lipsum,booktabs}
\usepackage{url}

\usepackage{hyperref}

\usepackage[T1]{fontenc}% http://ctan.org/pkg/fontenc
\usepackage[outline]{contour}% http://ctan.org/pkg/contour

\contourlength{0.1pt}
\contournumber{10}
\usepackage[dvipsnames]{xcolor}

\newcommand{\contrastive}{JE }
\newcommand{\Contrastive}{JE }
\newcommand{\reconstruction}{REC }
\newcommand{\Reconstruction}{REC }

\usepackage{amsmath}
\usepackage{amssymb}
\usepackage{mathtools}
\usepackage{amsthm}
\usepackage{wrapfig}
\usepackage{multirow}
\usepackage[compact]{titlesec}      
\usepackage{breakcites}
\usepackage[capitalize,noabbrev]{cleveref}

\theoremstyle{plain}

\theoremstyle{definition}

\theoremstyle{remark}

\usepackage[textsize=tiny]{todonotes}

\title{Objectives Matter: Understanding the Impact of Self-Supervised Objectives on Vision Transformer  Representations}
\author{
  Shashank Shekhar$^{1}$\thanks{Work done during an AI Residency at FAIR. Corresponding author: sshkhr@meta.com}
  \hspace{0.5cm}
  Florian Bordes$^{1,2,3}$
  \hspace{0.5cm}
  Pascal Vincent$^{1,2,3}$
  \hspace{0.5cm}
  Ari S. Morcos$^{1}$\\
  $^1$Meta AI (FAIR)\hspace{0.5cm}$^2$Mila-Quebec AI Institute\hspace{0.5cm}$^3$Universit\'e de Montr\'eal, DIRO\\
}
\begin{document}

\maketitle

\begin{abstract}
Joint-embedding based learning (e.g., SimCLR, MoCo, DINO) and reconstruction-based learning (e.g., BEiT, SimMIM, MAE) are the two leading paradigms for self-supervised learning of vision transformers, but they differ substantially in their transfer performance. Here, we aim to explain these differences by analyzing the impact of these objectives on the structure and transferability of the learned representations. Our analysis reveals that reconstruction-based learning features are significantly dissimilar to joint-embedding based learning features and that models trained with similar objectives learn similar features even across architectures. These differences arise early in the network and are primarily driven by attention and normalization layers. We find that joint-embedding features yield better linear probe transfer for classification because the different objectives drive different distributions of information and invariances in the learned representation. These differences explain opposite trends in transfer performance for downstream tasks that require spatial specificity in features. Finally, we address how fine-tuning changes reconstructive representations to enable better transfer, showing that fine-tuning re-organizes the information to be more similar to pre-trained joint embedding models. 
\end{abstract}


\section{Introduction}
\label{introduction}

Self-supervised learning (SSL) methods have become the de facto approach to training large machine learning models since they don't require labeled data and learn representations that generalize to many downstream tasks \cite{brown2020language, chen2020big}. In computer vision, SSL approaches learn by optimizing proxy objectives to learn representations that are informative both out of the box \citep{caron2021emerging} and for supervised transfer learning \cite{chen2020simclr}. Transformer models \citep{vaswani2017attention} were introduced as sequence-to-sequence models for natural language translation, but were adopted for vision (\citet{dosovitskiy2020image}, ViT: Vision Transformer) by tokenizing image patches as inputs, and adding an extra \texttt{CLS} token to represent object class for training an image classifier.\looseness=-1 %Besides classification, ViTs have demonstrated state-of-the-art empirical results across a variety of visual tasks such as object detection \citep{wei2022contrastive}, semantic segmentation \citep{chen2022vision}, and strong results across a variety of other visual tasks \citep{khan2021transformers}. 

Among SSL methods for learning ViT representations, two broad categories have emerged:  joint embedding based learning \citep{chen2020simple, caron2021emerging} and reconstruction-based learning \citep{zhou2021image, he2022masked} (\textit{referred to as} \textbf{JE} \textit{and} \textbf{REC} \textit{respectively hereafter}). \Contrastive training objectives encourages representations that maximize view invariance between samples from the same image via a joint-embedding (Siamese) \cite{chen2020simsiam}\footnote{Joint-embedding learning is sometimes interchangeably used with `Contrastive' learning. However, contrastive learning also includes variable-contrastive methods like VICReg \citep{bardes2021vicreg} which do not rely on joint-embeddings, while joint-embedding methods also includes methods like BYOL \citep{grill2020byol} which do not use a contrastive objective function. In this work, we limit ourselves to joint-embedding (JE) based methods where similarity across two image views is maximized.}. In contrast, \Reconstruction objectives, which train models to reconstruct images in pixel space from a masked input, encourage representations that can accurately reconstruct local features.

% exploit the strong spatial correlation present locally in natural images. This involves masking/corruption of input tokens to learn a visual representation from this noisy input \cite{vincent2008extracting}, and then reconstructing the image in pixel space to maximize similarity between the input and the reconstruction.

Both methods have demonstrated strong empirical performance on downstream supervised tasks, but each method comes with its own strengths and weaknesses. \Contrastive learning demonstrates strong linear probe transfer, but requires careful selection of augmentations across views to learn useful invariances \citep{chen2020simple} and a suite of additional components to avoid representational collapse \citep{jing2021understanding}. \reconstruction learning does not require hand-crafted augmentations, but doesn't transfer as well when directly using pre-trained features without any fine-tuning \citep{he2022masked}, and also requires a decoder in pixel space, increasing the computational cost relative to \contrastive. Other methods \citep{el2021large, msn} have tried to combine these objectives.  For example, Masked Siamese networks \citep{msn} integrate masked image modelling with \contrastive based learning. While combining objectives leads to improved performance, the reason why these differences in transfer performance arise in the first place remains unclear.

% ari{better in RW}
% For supervised ViTs, \citet{raghu2021vision, park2022how} have shown key differences in how information is represented, distributed, and used for discriminative learning in ViTs versus CNNs. \citet{park2022how} showed how larger datasets alleviate issues with non convex losses during training, thus explaining the benefit of training ViTs on larger datasets \cite{raghu2021vision} and informing future research on bridging both architectural biases \citep{d2021convit}.

We seek to better understand the differences in representations learned across SSL ViT methods in order to diagnose what information is learned and discarded during SSL pre-training. This can help guide informed choices about which method to prefer for a downstream task, as well as to develop methods that best utilize both objectives. 
% Our aim is to develop a better understanding about the differences between \contrastive and \reconstruction learning from the lens of both representations and transfer, in order to inform future research on SSL of ViTs.
%While both methods have shown strong empirical performance on various computer vision tasks, there are several open questions in terms of \textit{how} each training learns to solve these visual tasks. Are the pre-trained representations learned by \contrastive and \reconstruction methods similar? How does supervised fine-tuning affect theirs representations? Are the similarities and differences in the representations learned by these methods affected by depth and layer-types? Understanding the answers to these questions is important to address several theoretical and practical questions about SSL; what information is present in SSL ViT representations, why \contrastive representations perform better for linear probe transfer while \reconstruction learning representations transfer better when the ViT is fine-tuned end-to-end \citep{he2022masked}. 
We study these questions by comparing the representations of a standard ViT-Base model \citep{dosovitskiy2020image} trained with 16x16 image patches (ViT-B/16) on the ImageNet \citep{deng2009imagenet} dataset across popular \contrastive (MoCo-V3 \citet{he2020momentum}, DINO \citet{caron2021emerging}) and \reconstruction methods (MAE \citet{he2022masked}) .  %using Centred Kernel Alignment (CKA) \citep{kornblith2019similarity}. -- a vector similarity index that has been demonstrated its utility for comparing neural network representations \citep{raghu2021vision, nguyen2020wide, grigg2021self}. 
We approach differences between SSL methods from the perspectives of representational (dis)similarity, accessibility of information for transfer, as well as changes that arise during fine-tuning, leading to the following contributions:

\begin{itemize}
\itemsep0em 
\item \contrastive representations are more similar to each other than \reconstruction representations and vice versa (even across architectures). These differences arise early in the network, and are concentrated in the Layer Norm and Multi-Head Self-Attention Layers (Section \ref{sec:res-representation}).
\item \contrastive models contain more linearly decodable representations because all relevant class discriminative information is available in final pre-projector layer \texttt{CLS} token. In contrast, \reconstruction models lack key invariances and distribute class discriminative information across layers, leading to poor downstream transfer without fine-tuning (Section \ref{sec:res-transfer}).
\item Training probes on multiple layers from \reconstruction models improves transfer. The downstream task also plays a role in transfer from frozen pre-trained representation, as we discover that reconstructive features transfer better to tasks requiring spatial specificity (Section \ref{sec:res-transfer}). 
\item Fine-tuning \reconstruction models makes them similar to \contrastive models by re-organizing class information into the final layer. During fine-tuning, \reconstruction models take a more efficient path through parameter space than \contrastive models (Section \ref{sec:res-ft}).
\end{itemize}

\section{Related Work} 

\paragraph{Self-Supervised Learning of Vision Transformers} 

Self-supervised learning (for ViTs) can be broadly categorized into two families of algorithms. First is the \contrastive SSL family \citep{chen2020simclr, grill2020byol, zbontar_barlow_2021, bardes2021vicreg, chen2020simsiam} which rely on training criteria that encourage the representations learned from different augmentations of a given image to be close together. Second is \reconstruction SSL family which rely on a reconstruction loss in the pixel space that doesn't require handcrafted data augmentations but instead utilizes a decoder to reconstruct from the noisy representation \citep{zhou2021image, xie2022simmim, he2022masked}. 
%Unlike CNNs, ViT architectural biases of taking in image patches allow easy implementation of input noise by masking these patches, thus making ViTs suitable to masked image modelling and \reconstruction SSL. 
%While \contrastive learning methods were quite successful before ViTs, specific methods that leverage ViTs like DINO \cite{caron2021emerging} showed that \contrastive pre-training leads to emerging properties like edge detection in ViTs. 
\citet{he2022masked} showed that a simple masking approach (MAE) tailored for ViTs followed by a pixel-level reconstruction objective outperforms all other methods for fine-tuning and scaling with dataset and ViT size. However, the performance of MAEs with linear probes was much poorer than that of \contrastive models. Research on combining both methods has focused on sample efficient learning and transfer. \citet{msn} tried to utilize masked image modelling to improve efficiency for \contrastive learning and better few-shot transfer. \citet{el2021large} combined joint-embedding learning with \reconstruction learning across disjoint subsets of patches, as well as utilizing feature space augmentations and contrastive loss to improve training sample efficiency.  \citet{park2023what} performed an extensive study on the differences in pre-trained feature diversity, scale of features, and texture versus shape bias across both methods, and showed that a simple linear combination of losses outperforms individual objectives.

% SimSiam \cite{chen2020simple} showed that contrastive learning can work even without additional components like negative samples, momentum encoders and stop gradient operations.  

% Following the resounding success of masked language modelling \citep{devlin2018bert, brown2020language} for learning natural language representations, there has been a greater push to implement a similar self-supervised approach towards learning visual representations. 

% In the past, denoising autoencoders \cite{vincent2008extracting, vincent2010stacked} introduced denoising autoencoders for learning deep representations from noisy input via a reconstruction loss objective. Unlike CNNs, ViT architectural biases of taking in image patches allow easy implementation of input noise by masking these patches, thus making ViTs suitable to masked image modelling and reconstruction-based SSL. Indeed, several methods like i-BOT \citep{zhou2021image}, SimMIM \cite{xie2022simmim}, and MAE \cite{he2022masked} have utilized this approach to successfully learn SSL ViT representations. 

\begin{figure}[t]
\centering
\includegraphics[width=\columnwidth, center]{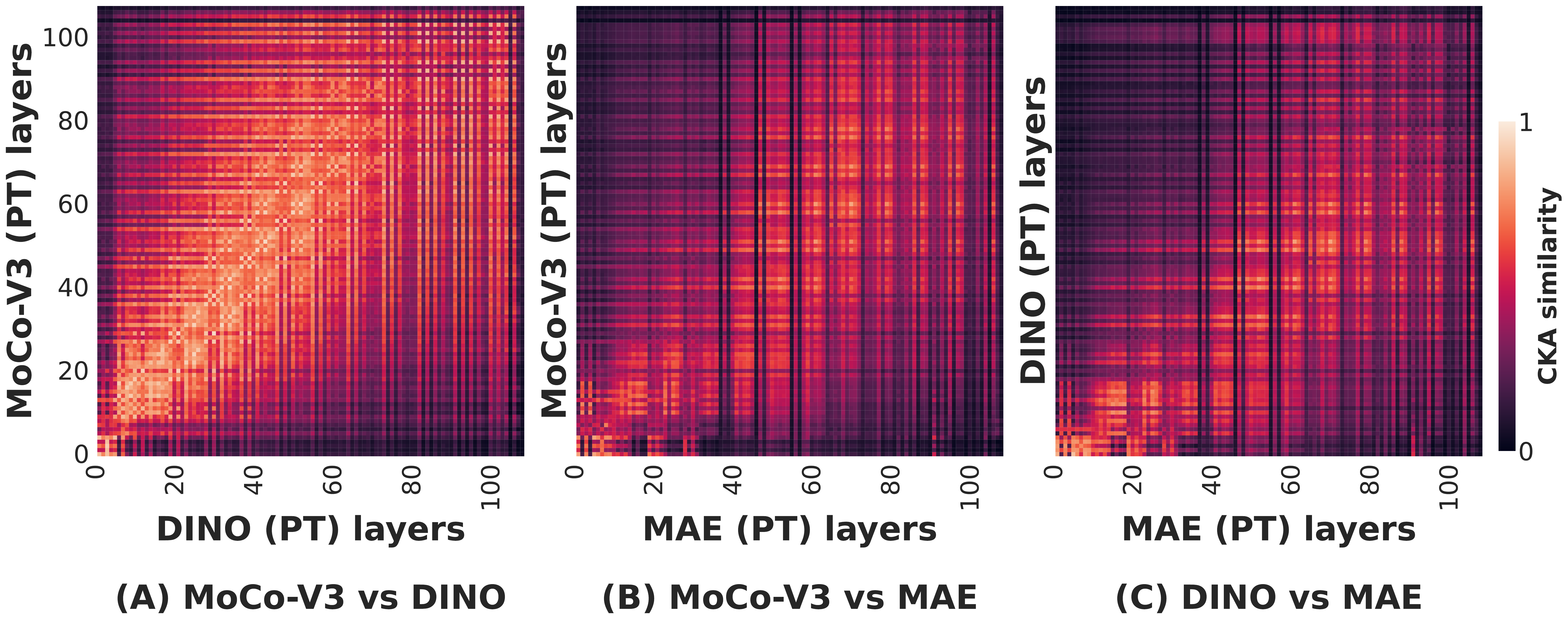}
\caption{CKA similarity between Pre-trained (PT) ViT models trained with different SSL methods. \textbf{\contrastive models show high similarity, and are less similar to \reconstruction models}. Furthermore, \contrastive ViTs demonstrate strong layer-to-layer correspondence in similarity, but groups of \contrastive layers correspond to groups of \reconstruction layers.}
\label{fig:fig_cka}
\end{figure}

\paragraph{Representation Similarity Analyses as a Lens for Model Understanding}

% In order to better understand what features each SSL method learns, we need to compare and contrast the representations learned across layer depths and types. 
Representational Similarity metrics provide a method for comparison of neural network representations across layer dimensionality, model initialization, and neural architectures \citep{raghu2017svcca, morcos2018insights, kornblith2019similarity}. Among these, Centered Kernel Alignment (CKA) \citep{kornblith2019similarity} between two representation matrices is given as the normalized Hilbert-Smith Independence Criteria \citep{gretton2007kernel} of the Gram similarity matrices. We adapt the formalization from \citep{nguyen2020wide} which approximates the linear CKA metric by averaging over $k$ minibatches to obtain the minibatch CKA metric. \citet{raghu2021vision} utilized CKA to demonstrate that information is localized and distributed differently across CNNs and ViTs, and that training set size plays an important role in the scale of features learned by supervised ViTs. 
% \citet{nguyen2020wide} utilized it to investigate how width and depth of networks affects the features learned. 
\citet{grigg2021self} used it to analyze how supervised and SSL representations defer while controlling for model architecture and training datasets. \citet{park2023what} showed low feature diversity in attention heads in pre-trained \contrastive models versus \reconstruction models, showing high CKA values across depth, attention heads, and tokens.\looseness=-1

\paragraph{Transfer Learning from SSL representations}

% While CKA values are informative of representation level similarity, they do not inform us about the information content in said representations. Hence, we need more fine-grained approaches to understand what information is available in these representations or how it transfers to downstream tasks.
Different SSL methods can have very different downstream performances. To visualize how invariances differ between SSL and supervised-trained representations, \citet{bordes2022high} trained a Representation Conditioned Diffusion Model (RCDM) to generate images conditioned on a given pretrained representation. While most SSL methods analyze how intermediate probes \citep{alain2017understanding} perform for linear transfer, \citet{pmlr-v162-evci22a} showed that probes trained on intermediate layers in addition to final layer features improve transfer performance and robustness. In the supervised setting, \citet{neyshabur2020being} showed that the scale of features being transferred during fine-tuning depends on the relation between the pre-training and transfer tasks. \citet{asano2019critical} showed that (older) SSL methods for CNNs cannot match supervised performance irrespective of amount of data and augmentation used, while \citet{el2021large} showed that \reconstruction SSL is more robust to type and size of dataset versus \contrastive learning. 

% We provide additional details on methods and different experimental setups in Appendix \ref{app:background}.

\begin{figure}[t]
\centering
\includegraphics[width=\columnwidth]{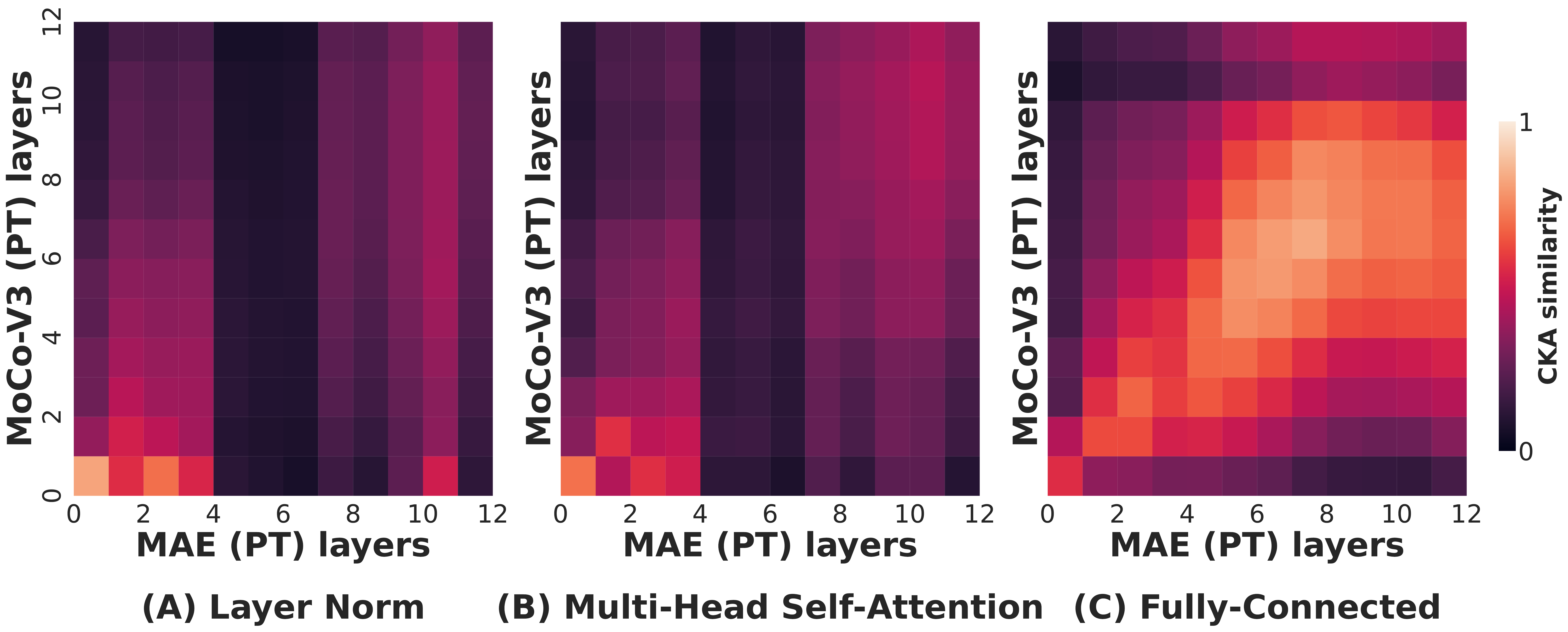}
\caption{CKA simlarity between MoCo-V3 and MAE for different type of layers. Within a ViT block, \textbf{attention and normalization layers are much more dissimilar than fully-connected layers.} NOTE: Layer indices here refer to the index by layer type (each layer appears once in a ViT-B made up of 12 ViT blocks.)}
\label{fig:fig_cka_layer}
\end{figure}

\section{Experiments and Results}

\subsection{How does representational structure of ViTs trained with different SSL objectives compare?}
\label{sec:res-representation}

To analyze why reconstructive models transfer differently than \contrastive models without any fine-tuning, we need to understand how their representations compare to each other. Is representation structure largely similar since the model and training data are same or fundamentally different due to the different training objectives? We perform pairwise comparisons of the representational structures of MoCo-V3, DINO, and MAE using CKA (Fig. \ref{fig:fig_cka}) to answer this.

We observe that the two \contrastive learning procedures (MoCo-V3 and DINO) have very similar representations (Fig. \ref{fig:fig_cka}A). Early and intermediate layers show a very strong correspondence in their representations across the two methods, but are dissimilar in the later layers (final three transformer blocks) which could be explained by how each \contrastive method learns different view invariances based on its augmentations \citep{grigg2021self}. In comparison, the \reconstruction learning method (MAE) has representations that are very dissimilar to both \contrastive methods (Fig. \ref{fig:fig_cka}b,c).
%The representations between corresponding layers of a \reconstruction and \contrastive model are much less similar. 
We also observe emergence of block-wise correspondences in layer similarities: the first quarter of MAE layers are similar to the first half of layers in MoCo-V3 and DINO, while the last three-quarters are similar to the last half. This shows that there are differences in how spatial information is aggregated and localized across the ViT layers in \contrastive and \reconstruction learning \citep{raghu2021vision}. 

We also perform representational and functional comparisons against MSN, which is a \contrastive learning method but utilizes masked image modelling like MAEs. We find that the \contrastive trianing objective governs the representational behaviour of MSN, as they are representationally and functionally more similar to \contrastive methods than \reconstruction methods. Discussion and results for MSN are provided in Section \ref{app:msn}.

\paragraph{Which layers drive these differences in representations?}   
\label{sec:res-rep-layers}

In Fig. \ref{fig:fig_cka}, we observe alternating patterns of lower and higher CKA similarity in layers across methods, suggesting that different types of layers within a ViT are similar to different degrees. Here, we focus on understanding whether the differences are higher in attention layers which encode global shape features, or in MLP layers which encode local texture features in a ViT \citep{naseer2021intriguing, park2023what}. In Fig. \ref{fig:fig_cka_layer}, we plot the CKA similarity across a subset of three layers in each ViT block: the layer normalization before the attention layer (Layer Norm), the multi-head self-attention layer (Multi-Head Self-Attention), and first linear layer after the residual connection (Fully-Connected). We observe that CKA similarity between attention and normalization layers across MAE and MoCo-V3 are much lower than fully connected layers. In fact, the first and intermediate attention layers of MoCo-V3 are entirely dissimilar from any of the attention layers in MAE, demonstrating that the objective-driven differences in representations are primarily due to changes in the attention and normalization layers.

% This suggests that the attention and normalization layers learn different orders of features in the intermediate layers of the ViT, and is consistent with the observation that bias towards shape features between \reconstruction methods and \contrastive methods is different, the latter being more shape-biased \cite{park2023what}. 

\begin{figure}[t]
\centering
\begin{subfigure}{0.49\columnwidth}

  \includegraphics[width=1.2\columnwidth, center]{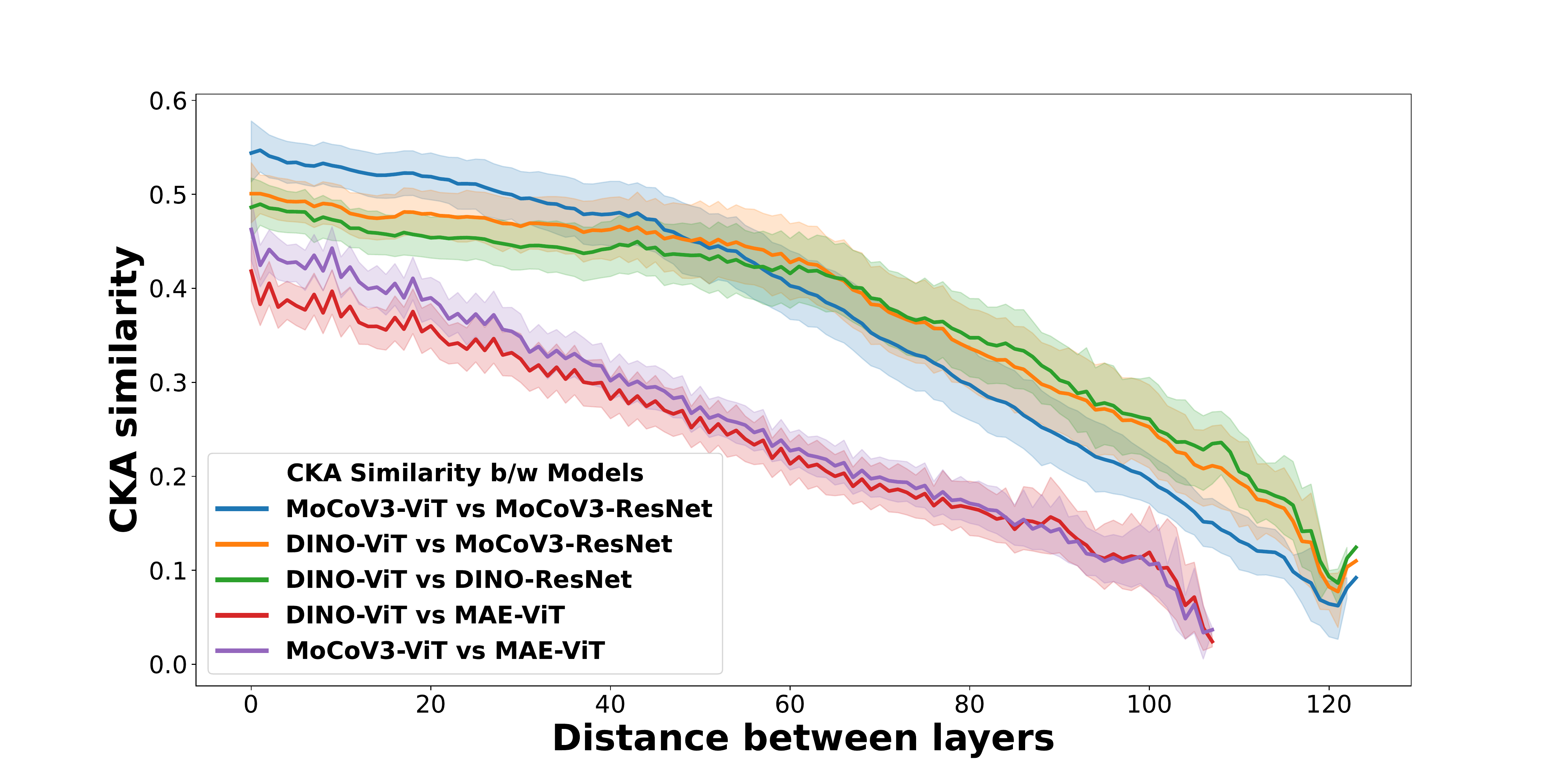}
\caption{\contrastive objectives across architectures}
\label{fig:arch_contrastive}
\end{subfigure}
\begin{subfigure}{0.49\columnwidth}
    \includegraphics[width=1.2\columnwidth, center]{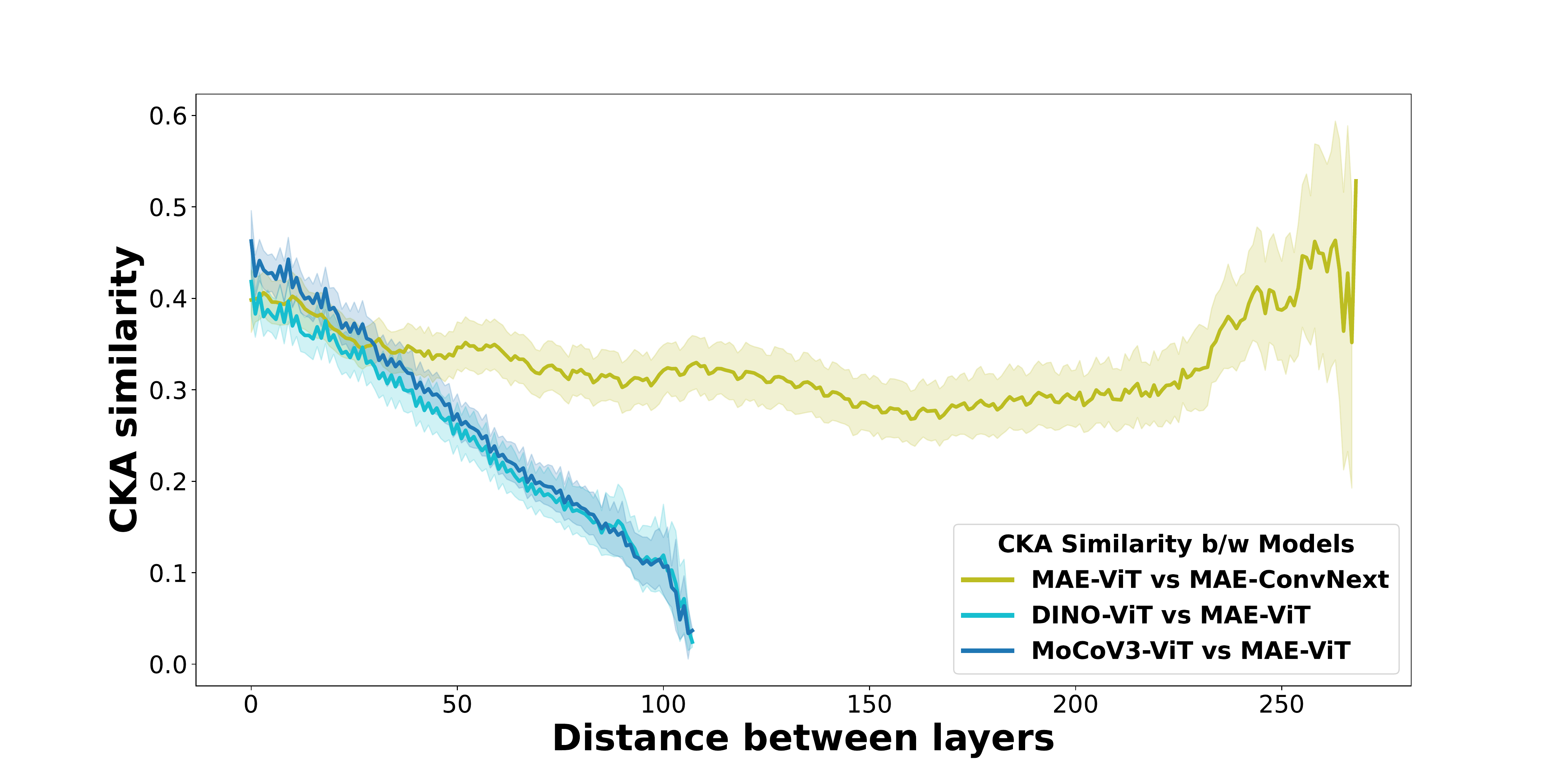}
\caption{\reconstruction objectives across architectures}
    \label{fig:arch_mae}
\end{subfigure}
\caption{CKA similarity vs layer distance across ViT and CNN architectures. \textbf{Similar SSL objectives yield similar representations even across models with very different architectures.}}
\label{fig:imagearch}
\end{figure}

\paragraph{Does objective or architecture drive representational structure?}
\label{sec:res-rep-arch}

CKA is an architecture agnostic lens to compare representation similarity, since it can be used to compare vector embeddings of different dimensions. \citet{raghu2021vision} utilized this to show that there are fundamental differences in inter and intra-model similarity across ViTs and ResNets when trained in a supervised manner on ImageNet. We want to understand if this holds true for SSL, and particularly whether the SSL objective or the architecture plays a larger role in determining representational structure. To test this, we compare the CKA values between ViT-CNN model pairs learned with the same SSL objective against ViT-ViT model pairs learned with different SSL objectives\footnote{See Appendix \ref{app:details} for details of CNN models used.} 

Our results in Fig. \ref{fig:imagearch} plot the CKA similarity between pairs of models as a function of the distance between two layers in each model pair. In Fig. \ref{fig:arch_contrastive}, we show that CKA similarity for two \contrastive models trained on different architectures is consistently higher than for a \contrastive and a \reconstruction ViT. In Fig. \ref{fig:arch_mae}, we observe that the inter-layer CKA for \reconstruction CNN-ViT models is of similar order of magnitude as the CKA for a \contrastive and a \reconstruction ViT when the layer depths are similar (layer distance lower). However, as the distances between the layers being compared increases, the CKA across \reconstruction ViT-CNN pair stays high while the CKA across pre-training objectives in a ViT-ViT pair falls off. Hence, we conclude that the SSL objective governs representational similarity more than architecture choice for both \reconstruction and \contrastive learning.

\paragraph{How do representational differences manifest when utilizing self-supervised features for class predictions?} %ari{We should include results for the resnets here if possible, but low priority}}
\label{sec:res-rep-depth-wise}

\begin{figure}[t]
\centering
\includegraphics[width=\columnwidth]{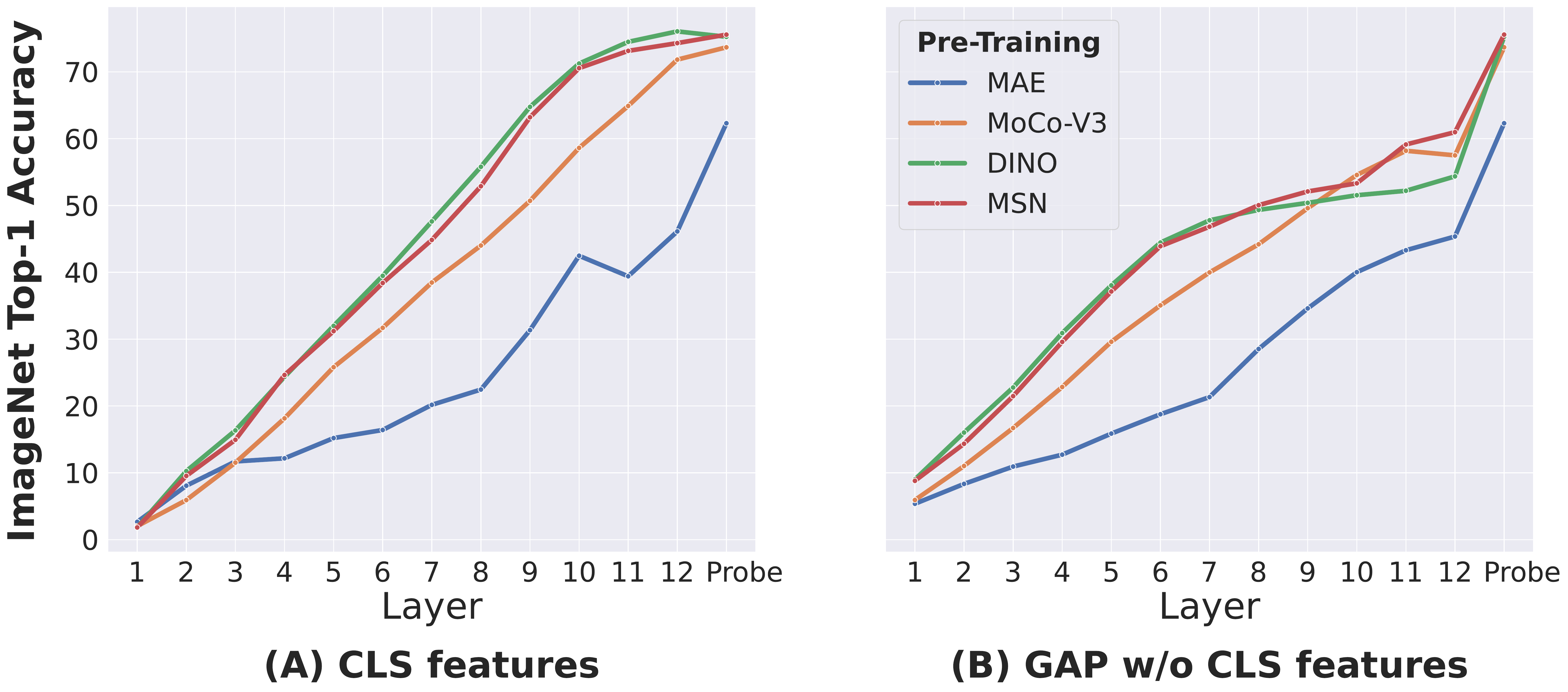}
\caption{20-Nearest Neighbour classification accuracy across ViT blocks and linear probe. \textbf{Class separability starts to decrease in MAEs after the first few ViT layers. The \texttt{CLS} token features in the last ViT layers of \contrastive models contain a significant amount of class information}.\vspace{-0.2cm}}
\label{fig:kNN}
\end{figure}

Our CKA analyses establish that representations across \contrastive and \reconstruction models are dissimilar, and vary by layer type.  Next, we consider how class discriminative information diverges between layers. Prior work has shown that linear probe transfer works better for \contrastive models than for MAE \cite{he2022masked}, but did not consider the amount of class specific information in the final ViT layer relative to the probe, and how it is distributed across layers. In order to do so, we calculate the 20 nearest-neighbour classification accuracy after each transformer block (12 in total in ViT-B) as well as for a linear probe trained on top of the SSL representation. Following \citep{raghu2021vision}, two different representations are used: the \texttt{CLS} token features, as well as Global Average Pooled features from all tokens except the \texttt{CLS} token (GAP w/o \texttt{CLS}), in order to ensure we utilize class information present in the \texttt{CLS} token as well as outside the \texttt{CLS} token. 

We plot the classification accuracy in Fig. \ref{fig:kNN}, noting two observations: first, the class separability of the MAE model starts to diverge from the MoCo-V3 model after the third transformer block, corresponding to the first quarter of MAE layers that were observed to be similar to the first half of the MoCo-V3 layers in Section \ref{sec:res-representation}. Secondly, the MoCo-V3 model already contains significant amount of class information in the last transformer layers, as the gap between 20-NN accuracy between the probe and final ViT block is negligible. For all the features besides the class token, there is less of a class information gap in the last layers of \contrastive models and \reconstruction models, implying that class information starts accumulating in the \texttt{CLS} token in the final blocks of \contrastive ViTs but not in \reconstruction ViTs.

Beyond classification accuracy, we also establish that the predictions are highly consistent across \contrastive models, demonstrating that similar pre-training objectives lead to features that represent different object classes similarly leading to class predictions which are also right and wrong in similar ways. (See Appendix \ref{app:KTrank} for details.)

\begin{figure}[t]
\centering
\includegraphics[width=0.8\columnwidth]{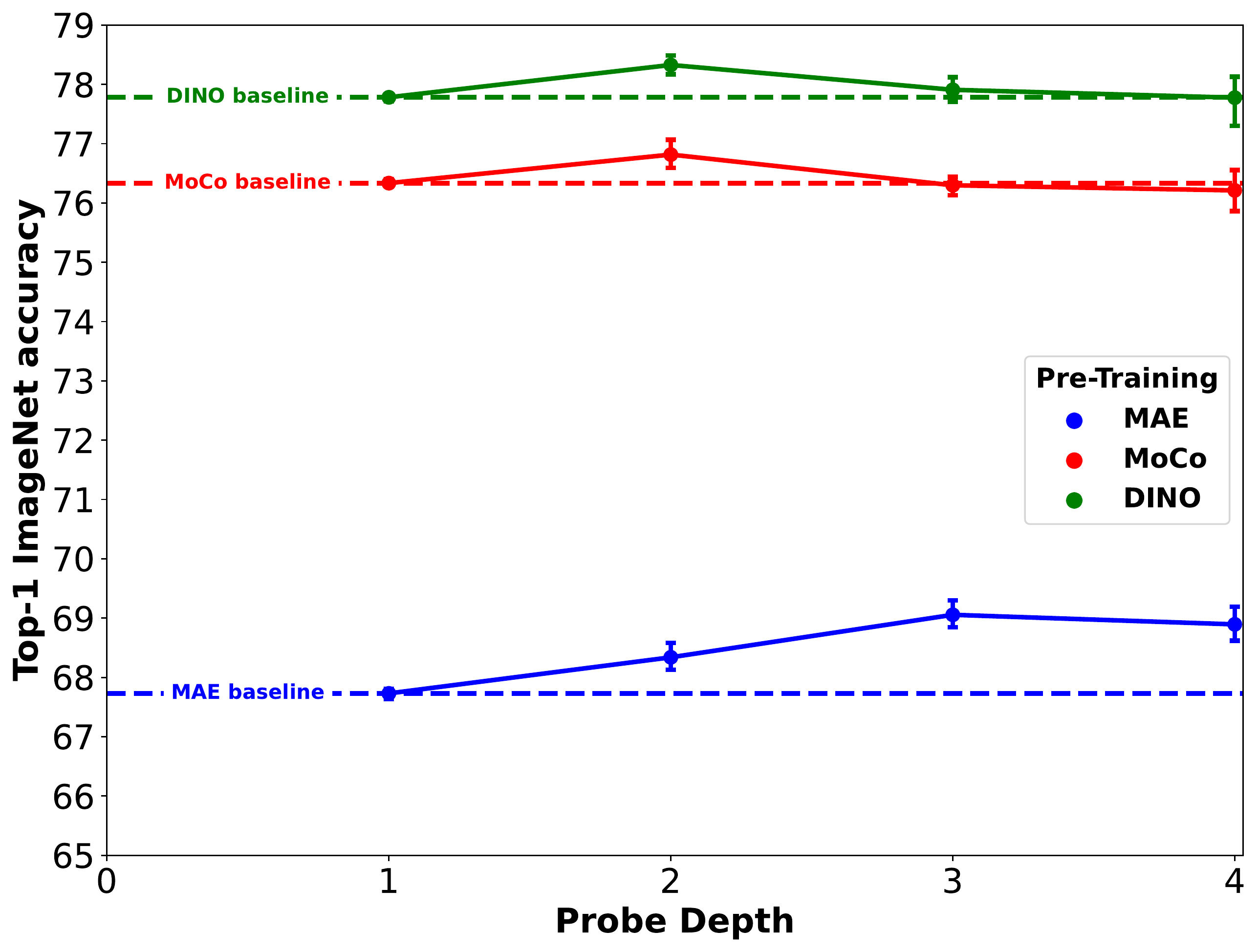}
\caption{Supervised transfer with increasing probe depth (mean ± std for best 5 runs). \textbf{MAE contains additional information about classification that can be accessed by a deeper non-linear probe, but in \contrastive models nearly all class discriminative information is already accessible to a linear probe (depth=1).}\vspace{-0.4cm}}
\label{fig:probes}
\end{figure}

\subsection{How does the SSL objective impact distribution of information in learned representations?}
\label{sec:res-transfer}

\paragraph{Is class discriminative information in pre-trained MAEs non-linearly decodable?}

While previous work on \contrastive and \reconstruction learning has consistently demonstrated that the final \texttt{CLS} token in the former contains features which can more easily be decoded by a linear probe, there has been little exploration into utilizing non-linear probes for transfer. While fine-tuning is a well-established method for transfer learning, a non-linear probe is a less expensive mechanism for supervised transfer. Furthermore, unlike fine-tuning, a non-linear probe only utilizes the features already learned by a pre-trained MAE without changing them based on a supervised signal, and thus is a more representative measure of class information in pre-trained SSL features. 

In Fig. \ref{fig:probes}, we show the results on Top-1 ImageNet accuracy when using non-linear probes of increasing depth on the final  \texttt{CLS} token features of an SSL pre-trained ViT. We perform extensive hyper-parameter sweeps, as outlined in Appendix \ref{app:details-linprobe} and report mean and standard deviation of best 5 runs in order to ensure that training stochasticity does not impact our results. Both \contrastive models see very small improvements from using a two-layer probe (0.44 \% and 0.48\% respectively), and no improvement in performance from using deeper probes. However, an MAE pre-trained ViT reports a large performance improvement of 1.33\% on top-1 accuracy when using a three layer non-linear probe. While this improvement is not enough to match the performance of \contrastive final layer features, it demonstrates that there is additional class-specific information  available in final \texttt{CLS} token features, but it is inaccessible with a linear probe.

\begin{figure}[t]
\centering
\includegraphics[width=0.9\columnwidth]{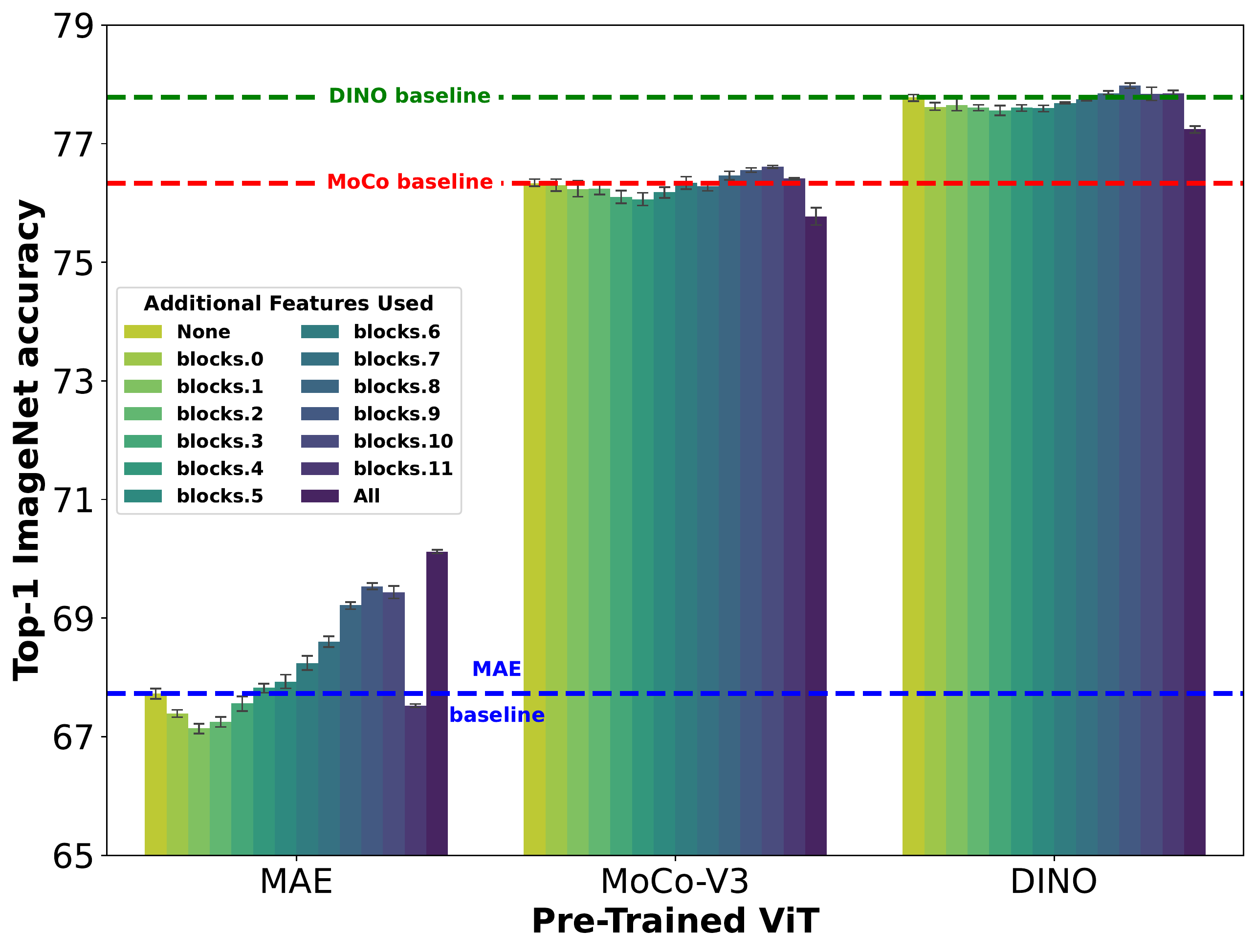}
\caption{Using additional intermediate layer \texttt{CLS} token features for linear probe transfer (mean ± std for best 5 runs). \textbf{Intermediate features provide additional information and improve accuracy for MAE, but do not provide any additional information over last layer features in DINO and MoCo-V3.}\vspace{-0.4cm}}
\label{fig:marginal}
\end{figure}

\begin{figure*}[ht]
\begin{subfigure}{.32\columnwidth}
  \centering
  \includegraphics[width=\columnwidth]{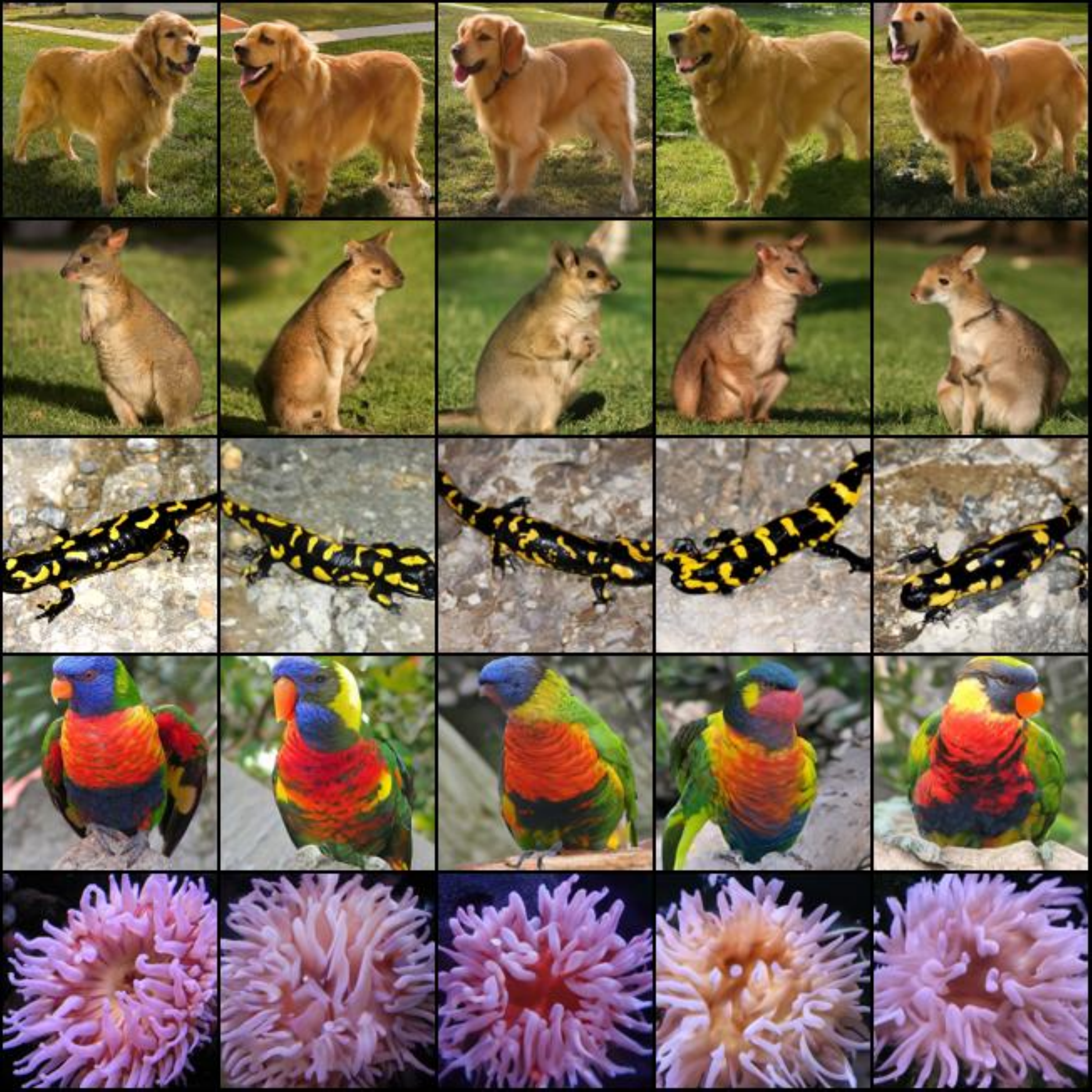}
  \caption{DINO}
  \label{fig:rcdm1}
\end{subfigure}%
\hfill
\begin{subfigure}{.32\columnwidth}
  \centering
  \includegraphics[width=\columnwidth]{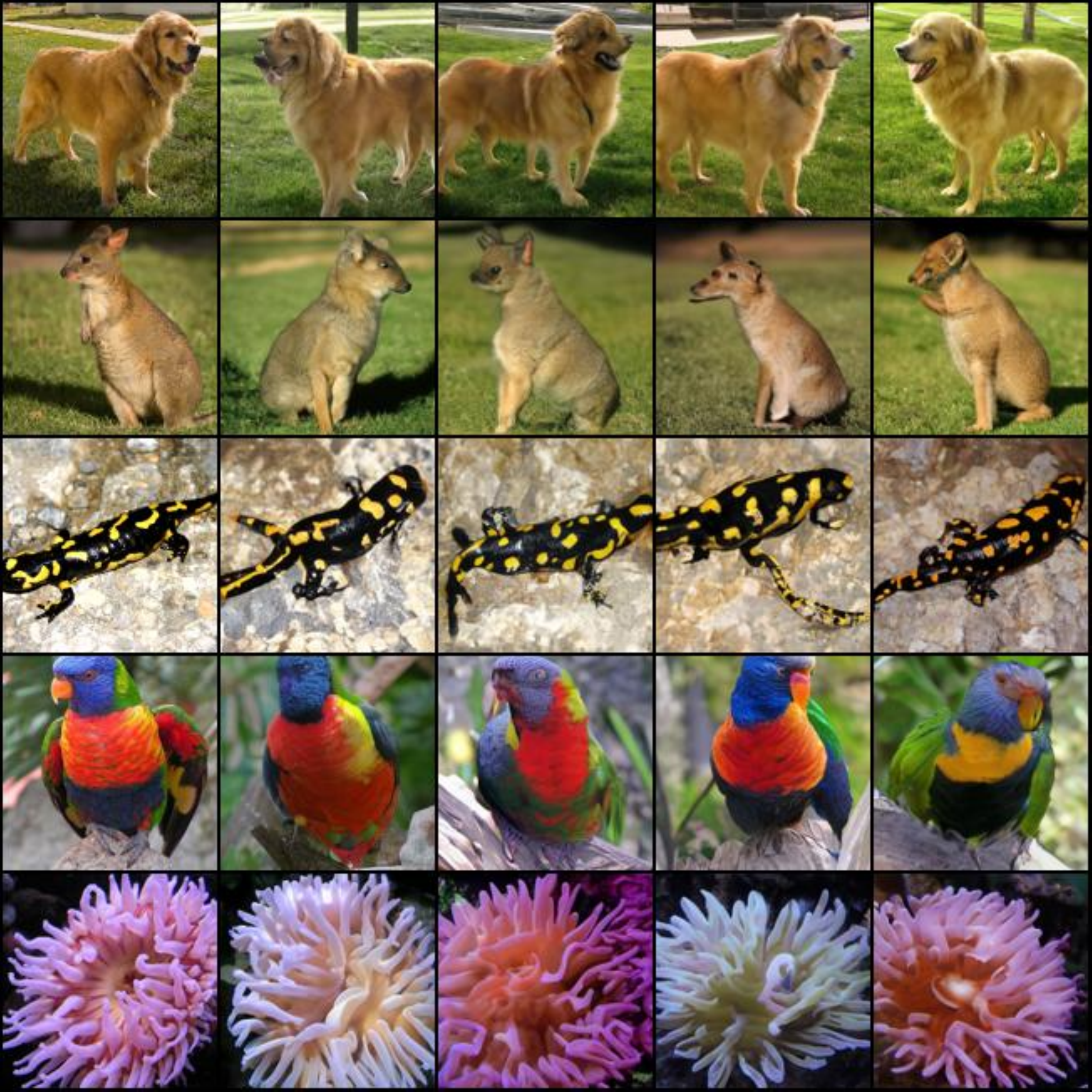}
  \caption{MoCo-V3}
  \label{fig:rcdm2}
\end{subfigure}%
\hfill
\begin{subfigure}{.32\columnwidth}
  \centering
  \includegraphics[width=\columnwidth]{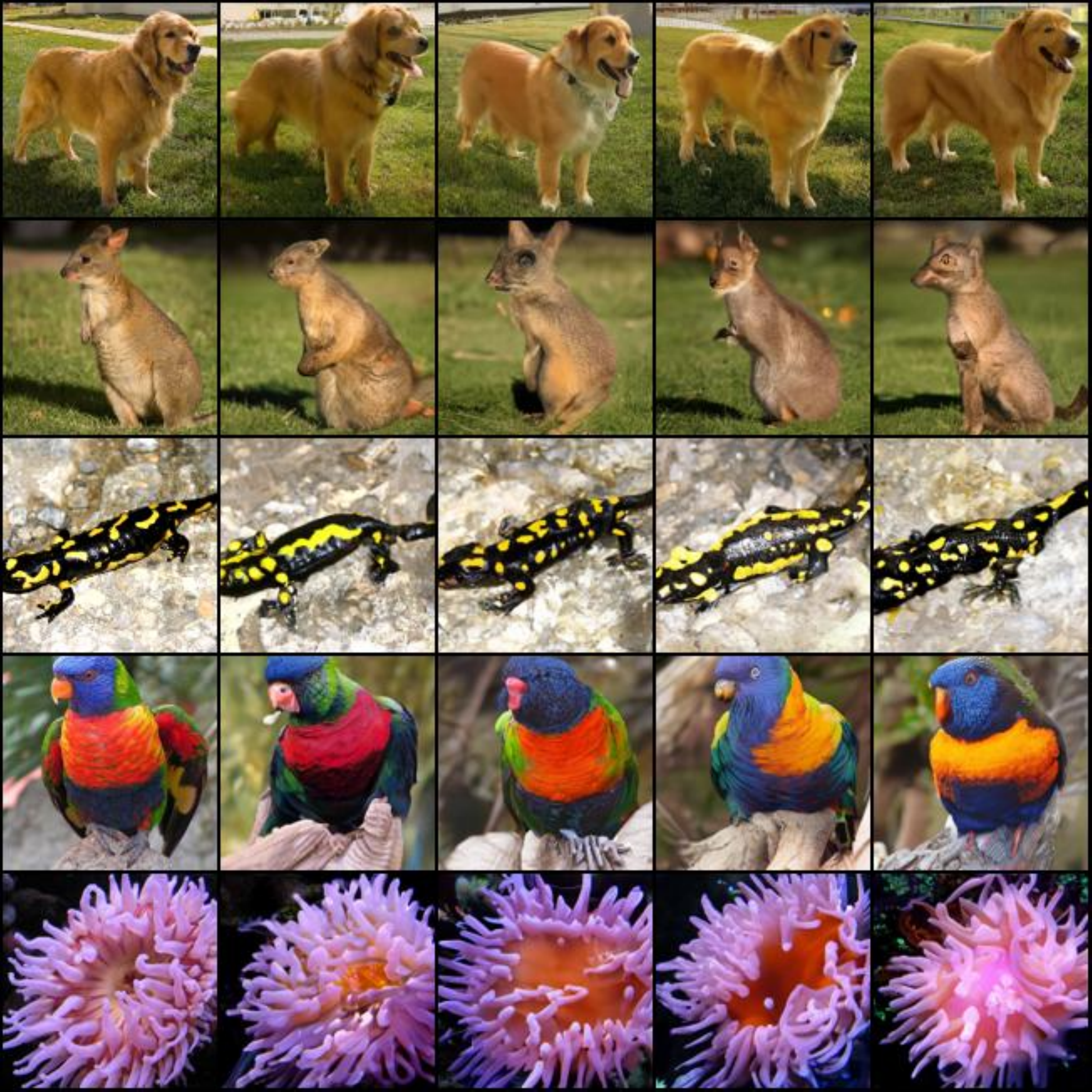}
  \caption{MAE}
  \label{fig:rcdm3}
\end{subfigure}
\caption{Visualization of samples generated by a generative 
diffusion model conditioned on pre-trained ViT features. We use RCDM  \citep{bordes2022high} that we trained on the face-blurred version of ImageNet \citep{yang2021imagenetfaces}. The information contained in the representation stays consistent across generated samples, while the missing information varies across samples. The source image is in the leftmost column, while the images generated by the RCDM model using the ViT features of the source image are in the next four columns. \textbf{\contrastive features generate images that are horizontally oriented in both the same and opposite direction to the source image, demonstrating that \contrastive features are invariant to horizontal flips. On the other hand, MAE features lack this invariance and only generate samples horizontally oriented in the same direction as the source.}}
\label{fig:rcdm}
\end{figure*}

% \usepackage{graphicx}
\begin{table}[b]
\centering
\resizebox{\columnwidth}{!}{%
\begin{tabular}{|l|l|l|l|}
\hline
        & Final Layer & Final + (Best) Intermediate & All layers        \\ \hline
DINO    & 77.78 ± 0.07\%     & 77.97 ± 0.05\% \textcolor{gray}{(+0.19\%)}           & 77.24 ± 0.07\% \textcolor{red}{(-0.54\%)} \\ \hline
MoCo-V3 & 76.33 ± 0.07\%     & 76.61 ± 0.02\% \textcolor{gray}{(+0.32\%)}           & 75.77 ± 0.18\% \textcolor{red}{(-0.56\%)} \\ \hline
MAE     & 67.73 ± 0.10\%     & 69.53 ± 0.07\% \textcolor{ForestGreen}{(+1.8\%)}            & 70.12 ± 0.04\% \textcolor{ForestGreen}{(+2.4\%)}  \\ \hline
\end{tabular}%
}
\caption{Linear Probe performance on ImageNet top-1 accuracy (mean ± std for best 5 runs) using features at different depths in pretrained ViTs. \textbf{MAE contains additional information for class discriminative learning in intermediate features, while \contrastive models do not.}}
\label{table:intermediate}
\end{table}

\paragraph{What feature invariances have been learned during self-supervised learning in the final layer of a ViT?}

We have established that the information present in the final \texttt{CLS} token is more suited for classification in \contrastive models than \reconstruction models for both linear and non-linear decoding. It becomes important then to characterize what information present in the final \texttt{CLS} token is used for transfer learning. In particular, are there a specific set of invariances to visual information that are stored in \contrastive features? To answer this question, we used RCDM \citep{bordes2022high}, a conditional diffusion model that uses pre-trained SSL representations as conditioning. For training, we used the face-blurred version of ImageNet \citep{yang2021imagenetfaces}. Since this representation is the only information about the target image that is fed into the model, RCDM is trained to extract as much information as possible from this representation in order to reconstruct the image as close as possible to the target image. As RCDM is a stochastic generative model, the information that varies across samples (because of the noise) is not contained in the representation while the information that remains constant across many samples is contained in the representation.

We visualize the samples obtained via RCDM for both \contrastive and \reconstruction models Fig. \ref{fig:rcdm}. From the conditionally generated samples, we can visualize how different pre-training approaches influence the degree of information, as well as types of invariances present in the pre-trained features. In particular, we find that \contrastive conditioned diffusion models generate images with horizontal flip invariance, whereas those conditioned on MAE representations do not. Being invariant to different horizontal alignments of objects contributes to a better linear probe transfer performance for \contrastive representations, since natural images (including those in ImageNet validation set) have objects in both orientations.

\paragraph{Does \reconstruction learning discard class discriminative information in its final layer?}

% \usepackage{graphicx}
% \usepackage{multirow}
% \usepackage{graphicx}

Our RCDM visualizations show that MAEs lack information relevant to class discriminative learning in the final \texttt{CLS} token features.  \citet{he2022masked} showed that simply fine-tuning the last few ViT blocks led to a significant improvement over linear probe transfer. This suggests that there is class discriminative information present in intermediate layers of MAE models that is accessible through fine-tuning. We explore whether this information can be accessed directly by a linear probe for classification, to establish whether class discriminative information is discarded across network depth by \reconstruction SSL. Similar to our experiments with non-linear probes, we perform an extensive hyper-parameter sweep and report mean and standard deviation of best 5 runs.

We find (Fig. \ref{fig:marginal}, Table \ref{table:intermediate}) that training a linear probe on both the final and an intermediate layer \texttt{CLS} token features leads to an improvement in classification accuracy for \reconstruction representations, but provides no marginal utility for \contrastive representations. We also control for the situation where this improvement could simply be case of using a larger dimensional probe, by training on concatenated \texttt{CLS} features from the last layer, and establish that this is not the case (Additional Features Used =  Block.11 in Fig. \ref{fig:marginal}). Furthermore, using all intermediate \texttt{CLS} token features from an MAE pre-trained model to train a linear probe gives a 2.4\% increase in top-1 accuracy vs using just the final \texttt{CLS} token features (Table \ref{table:intermediate}). These results demonstrate that class discriminative information is more distributed across \reconstruction models layers while this information is concentrated in the final layer of \contrastive models.

\paragraph{How do SSL pre-trained ViTs perform on downstream tasks requiring spatial specificity?}

In previous experiments, we examined the presence of information in SSL ViT features for linear probe transfer. However, image classification is just one possible downstream task, and does not require a model to preserve exact spatial information about object location. Other downstream transfer tasks, such as object detection and instance segmentation \citep{lin2014microsoft}, require the availability of precise object location in the pre-trained features for transfer. In order to test how SSL objectives influence ViT features for learning location-preserving information, we evaluate the performance of frozen pre-trained ViTs as backbone feature extractors on the MS-COCO detection and segmentation tasks. We utilize the ViTDet framework introduced by \citet{li2022exploring} to perform these experiments (see Appendix \ref{app:details-coco} for details). 

\begin{table}[t!]
\centering
\resizebox{\columnwidth}{!}{%
\begin{tabular}{|l|lll|lll|}
\hline
\multirow{2}{*}{} & \multicolumn{3}{c|}{\textbf{Detection}}                               & \multicolumn{3}{c|}{\textbf{Segmentation}}                             \\ \cline{2-7} 
                  & \multicolumn{1}{l|}{AP}    & \multicolumn{1}{l|}{AP\textsuperscript{large}} & AP\textsuperscript{small} & \multicolumn{1}{l|}{AP}    & \multicolumn{1}{l|}{AP\textsuperscript{large}} & AP\textsuperscript{small} \\ \hline
MAE      & \multicolumn{1}{l|}{\textbf{30.25}} & \multicolumn{1}{l|}{\textcolor{red}{39.93}}    & \textcolor{ForestGreen}{18.69}    & \multicolumn{1}{l|}{\textbf{28.56}} & \multicolumn{1}{l|}{\textcolor{red}{41.73}}     & \textcolor{ForestGreen}{14.37}    \\ \hline
MoCo-V3     & \multicolumn{1}{l|}{28.75} & \multicolumn{1}{l|}{40.21}    & 15.49    & \multicolumn{1}{l|}{26.67} & \multicolumn{1}{l|}{41.92}     & 10.89    \\ \hline
%DINO     & \multicolumn{1}{l|}{32.57} & \multicolumn{1}{l|}{44.14}    & 19.89    & \multicolumn{1}{l|}{30.04} & \multicolumn{1}{l|}{45.77}     & 14.37    \\ \hline
DINO (w/o multi-crop)     & \multicolumn{1}{l|}{29.97} & \multicolumn{1}{l|}{40.82}    & 17.55    & \multicolumn{1}{l|}{28.16} & \multicolumn{1}{l|}{42.53}     & 12.99    \\ \hline
\end{tabular}%
}
\caption{Downstream transfer for object detection and segmentation on MS COCO using a ViTDet based Mask R-CNN with a frozen backbone ViT. \textbf{MAE outperforms \contrastive models when transferring from frozen pre-trained features. Due to the scale of its features, it does worse than \contrastive models on larger objects, but performs better on small objects.}}
\label{table:coco_main}
\end{table}

Our results are shown in Table \ref{table:coco_main}. Contrary to image classification from frozen representations, we find that \texttt{CLS} token features from MAE actually outperform MoCo-V3 features on detection as well as segmentation. While DINO intially outperforms MAE (Table \ref{table:coco}), we hypothesize that it benefits from its unique multi-crop training setup since DINO is specifically trained to be invariant to both local and global scale of objects. In order to verify this, we train a DINO ViT without multi-crops and indeed find that a frozen MAE outperforms a frozen DINO for detection and segmentation. %We leave the analysis of MAE with multi-crop and augmentation based training for future works. 

For the detection and segmentation transfer, we also observe an interesting difference between \reconstruction and \contrastive models that pertains to the scale of objects. The frozen \reconstruction model performs worse than \contrastive models on localizing larger objects (AP\textsuperscript{large}), but performs better for localizing smaller objects (AP\textsuperscript{small}). Thus, the final \texttt{CLS} token features from a MAE are informative for localizing smaller objects, but lack global context to correctly detect larger objects. This observation is consistent with the \cite{park2023what} observations that the receptive field of \reconstruction models is more local in the last layer features versus \contrastive models. Our results establish that \reconstruction features can be useful out-of-the-box vs \contrastive features when the downstream task requires spatial specificity.

\paragraph{Is there information relevant to detection and segmentation in the intermediate layers of SSL pre-trained ViTs?}

\begin{figure}[t]
\centering
\includegraphics[width=0.8\columnwidth]{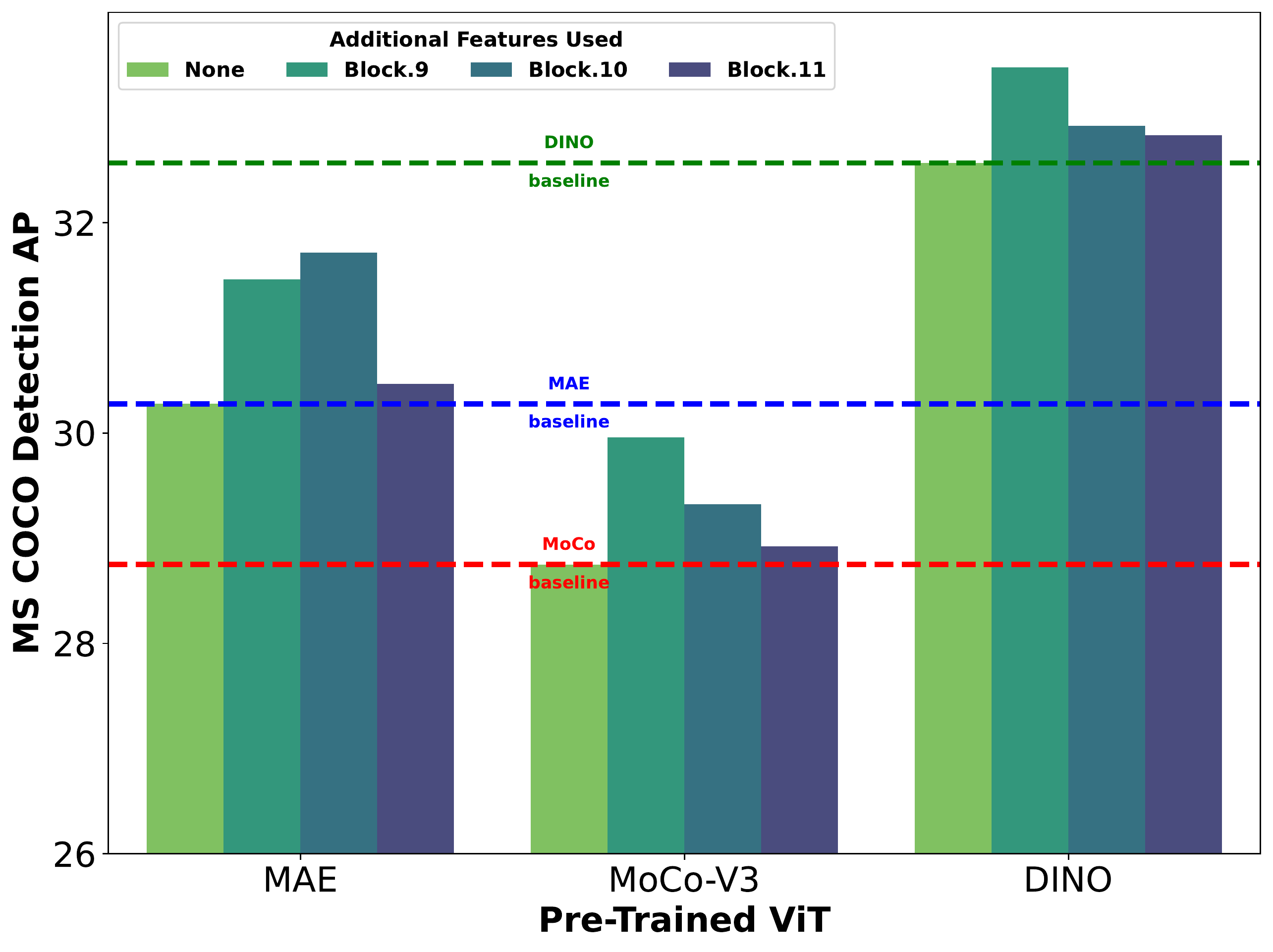}
\caption{Object detection results on MS-COCO using frozen pre-trained ViT features. \textbf{ViTs trained with both kinds of SSL objectives show improved performance when additional intermediate features are used for detection, unlike for classification.}}
\label{fig:intermediate-coco}
\end{figure}

How is information relative to spatially sensistive transfer tasks distributed across layers in SSL pre-trained models?
% Lastly, we seek to quantify whether the information relevant to downstream transfer to detection and segmentation is also concentrated in the last \texttt{CLS} token of \contrastive models, or is it distributed differently. This will help us understand whether the information distribution in intermediate layers is simply a function of the SSL objective, or whether it varies by the nature of the downstream task. 
We repeat our experiment in Table \ref{table:intermediate} and Fig. \ref{fig:marginal}, utilizing features from the final and intermediate layer \texttt{CLS} tokens, and concatenating them to build the feature pyramid for Mask R-CNN training. However, due to the computational constraints of training Mask R-CNN we limit ourselves to using intermediate features from ViT blocks 9, 10, 11.

We find that both a Mask R-CNN trained on both final and intermediate \texttt{CLS} token features outperforms a similar model trained only on the final layer features for both \reconstruction models as well as \contrastive models. While intermediate features offered no marginal utility in linear probe classification for \contrastive models, they contain information relevant to detection/segmentation which is not present in the final \texttt{CLS} token.  While the \contrastive objective lends itself better to classification by concentrating useful information in its last pre-projector layer, it loses some relevant spatial information.

\begin{figure}[t]
\centering
\includegraphics[width=0.9\columnwidth, center]{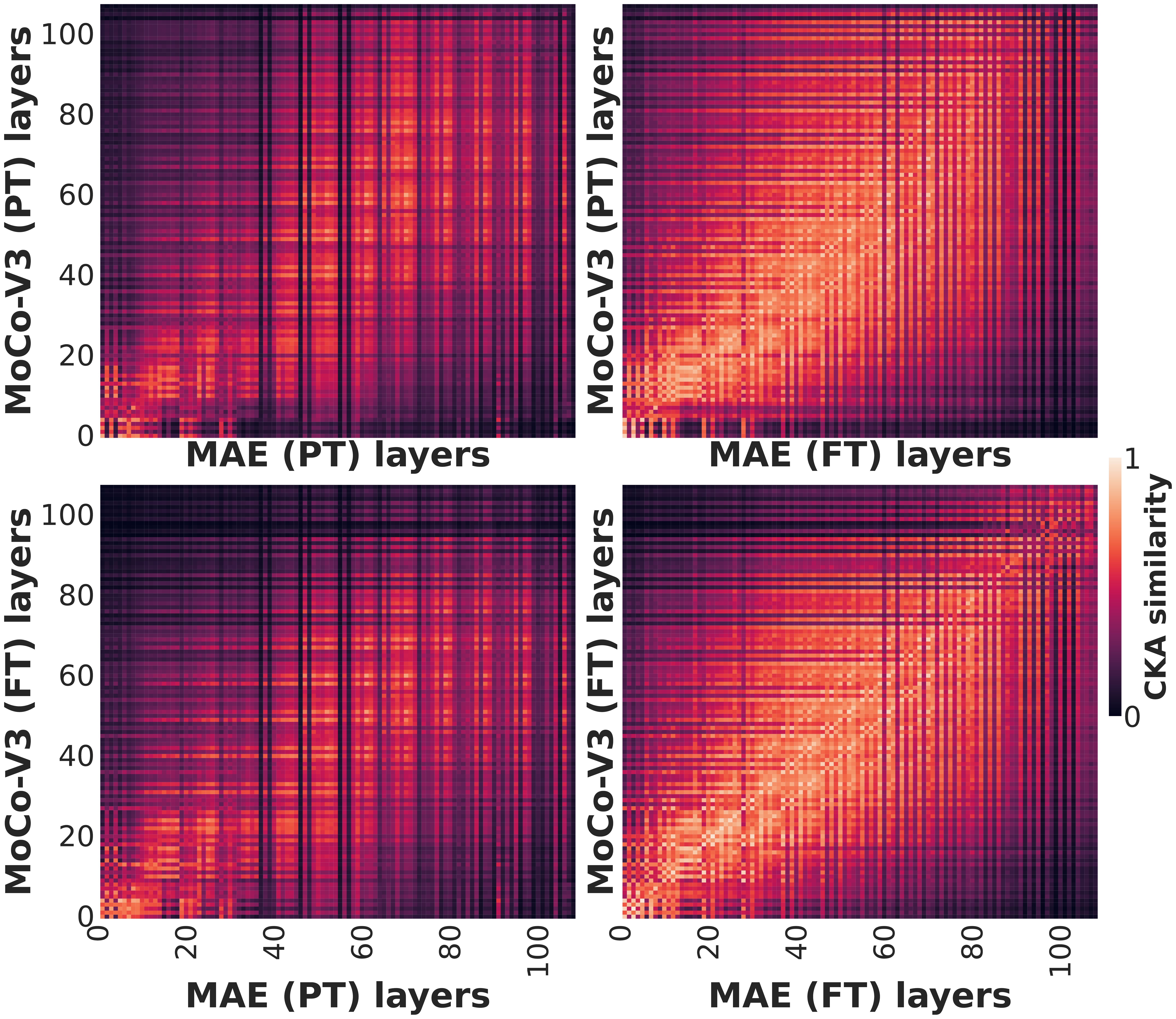}
\caption{CKA similarity between MoCo-V3 and MAE before (PT) and after fine-tuning (FT). \textbf{Fine-tuning a MAE makes it more representationally similar to a pre-trained MoCo-V3}.}
\label{fig:imageft}
\end{figure}

\subsection{What happens to self-supervised ViT representations post fine-tuning?}
\label{sec:res-ft}

Thus far, we have focused on pre-trained representations, but how does fine-tuning impact representational structure? Given the importance of fine-tuning to the downstream performance of \reconstruction models, this is a critical question. 

% Our experiments in Section \ref{sec:res-transfer} clarify several aspects of linear and non-linear probe transfer for classification as well as detection/segmentation from pretrained ViTs. However, it is also important to understand why fine-tuning of MAEs is important to their downstream transfer performance, and in fact how does it contribute so significantly to their improvement in transfer learning versus linear probes.

\begin{figure}[t]
\centering
\includegraphics[width=\columnwidth, center]{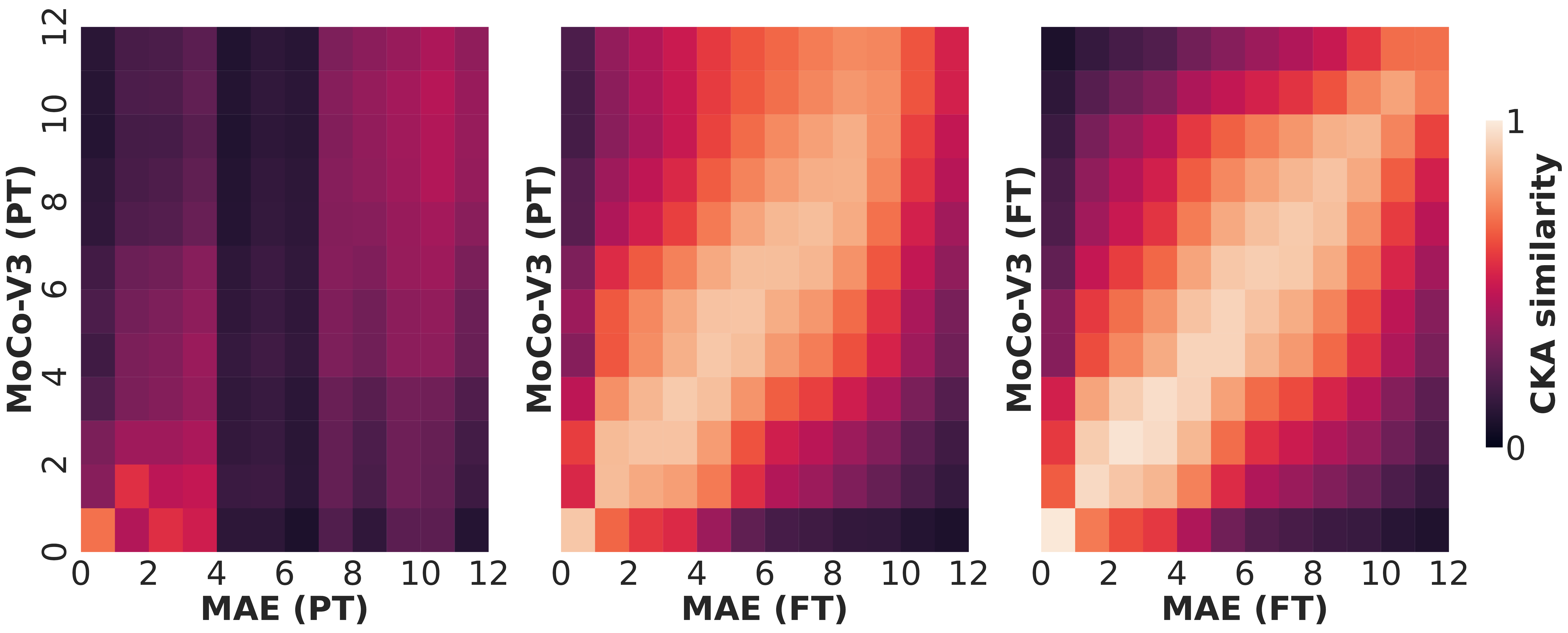}
 \caption{\textbf{Attention layers show strong linear correspondence as well as block correspondence in CKA similarity after fine-tuning \contrastive and \reconstruction models.} NOTE: Attention layer indices shown (one per ViT block, ViT-B made up of 12 ViT blocks.)}
\label{fig:imageftlayer}
\end{figure}

\paragraph{How does representational similarity change post fine-tuning?}   
\label{sec:res-ft-layers}
We first consider how the layer-wise CKA similarity changes as a result of fine-tuning. We find (Fig. \ref{fig:imageft}) that fine-tuned MAE features are highly similar to that of a pre-trained MoCo-V3, implying that instance discriminative \contrastive pre-training learns very similar representations to class discriminative fine-tuning. This correspondence remains after fine-tuning MoCo-V3 except in later layers (See Fig. \ref{fig:imagefta} for qualitatively similar results with DINO). 

\begin{figure}[]
\centering
\begin{subfigure}{.45\columnwidth}
    \centering
    \includegraphics[width=\columnwidth]{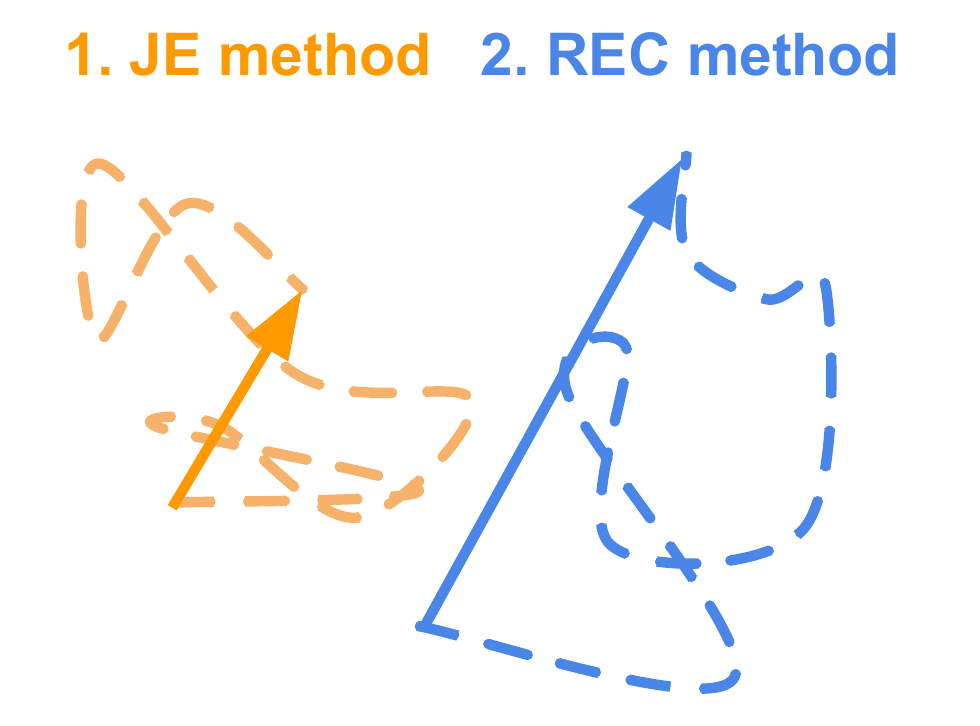}
    \caption{Sketch visualizing difference in fine-tuning dynamics of \contrastive and \reconstruction models}
    \label{fig:imageftd1}
\end{subfigure}
\hfill
\begin{subfigure}{.48\columnwidth}
    \centering
    \includegraphics[width=\columnwidth]{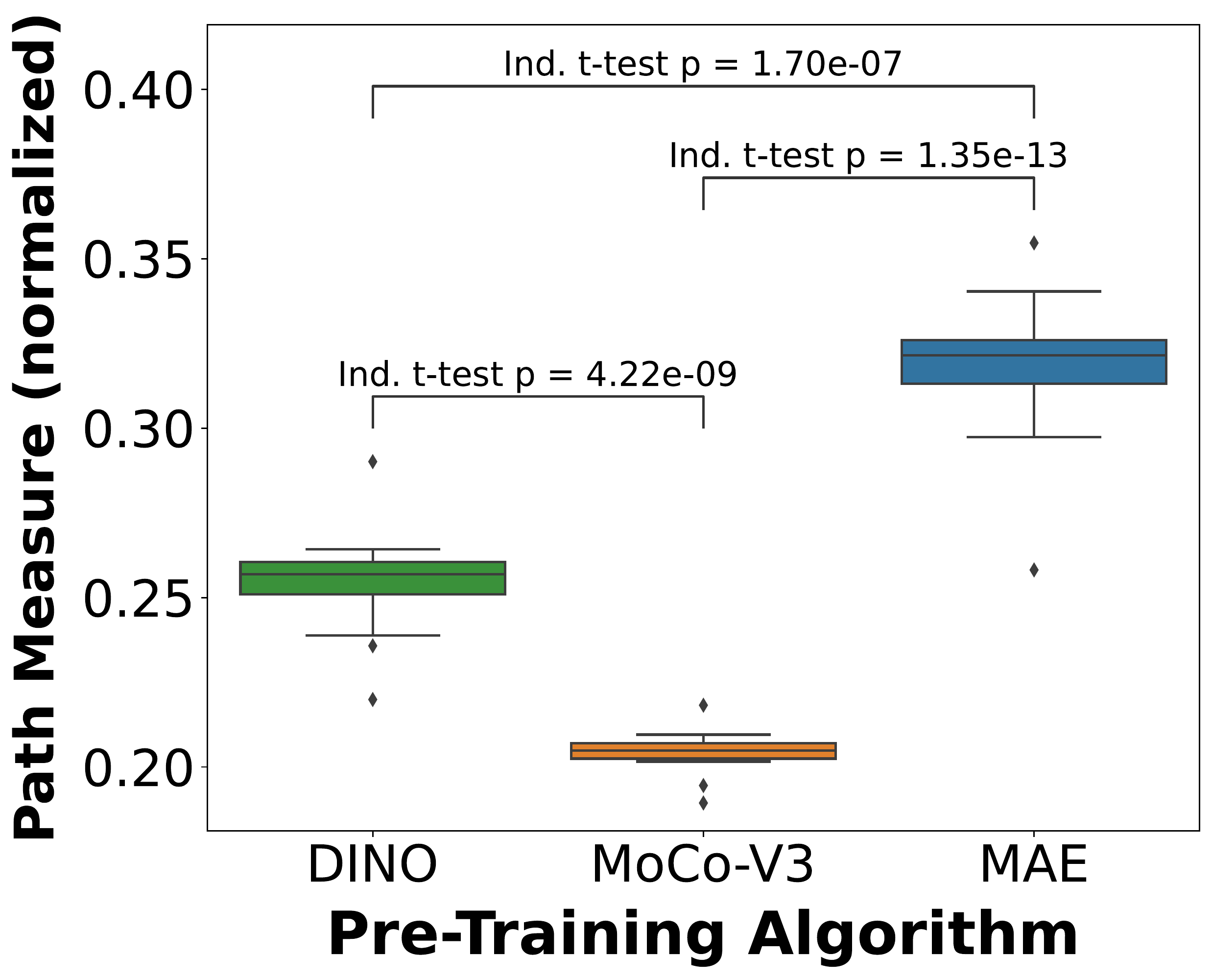}
    \caption{Fine-Tuning Path efficiency}
    \label{fig:imageftd2}
\end{subfigure}
 
\caption{Fine-tuning dynamics of attention layers of SSL ViTs. \textbf{MAE fine-tuning follows a more efficient path integral, and attention layer parameters converge more directly towards new values. On the contrary, \contrastive parameters do not follow an efficient path, and go through much higher displacement relative to the actual distance covered in parameter space.}}
\end{figure}
\label{fig:imageftd}

We repeat the experiment from Section \ref{sec:res-rep-layers} to analyze which layers drive this increase in similarity after fine-tuning. An increase in multi-head self-attention similarity would imply that fine-tuning enables MAEs to attend to spatial information differently, focusing more on global context instead of local context. Indeed, we find that the layers which were initially most dissimilar after SSL (multi-head self-attention and layer normalization) become the most similar after fine-tuning (Fig. \ref{fig:imageftlayer}, Fig. \ref{fig:image-app} in Appendix \ref{app:cka}), while the most similar pre-trained layers (fully-connected) become more similar only in the initial and intermediate layers, but become more dissimilar in the later layers of the ViT (Fig. \ref{fig:image-app} in Appendix \ref{app:cka}). We conclude that during fine-tuning, the way spatial features are attended to in ViTs pre-trained with \reconstruction learning changes significantly to align with how ViTs pre-trained with \contrastive learning attends to spatial features; which is largely consistent before and after fine-tuning.

\paragraph{A mechanistic understanding of increased similarity}   
\label{sec:mechanistic}

In the previous section, we found that the attention layers in the middle blocks (3-10) of \reconstruction models become highly similar to \contrastive models post fine-tuning. However, we cannot infer from CKA alone how the representations are transformed mechanistically during fine-tuning. In order to understand why fine-tuned MAEs can quickly exceed transfer performance of fine-tuned \contrastive models even with partial fine-tuning \citep{he2022masked}, we need to develop a mechanistic understanding of what happens during fine-tuning. 
%To do so, we look at the training dynamics during fine-tuning to understand how 'efficient' of a path the features follow in the ViT parameter space starting from different pre-trained representations. 
In order to do so we look at the the L${_2}$ norm of the total difference between the pre-trained and fine-tuned model parameters, and normalize it by the L${_2}$ norm of the difference between parameters after each epoch. This gives us a measure of the efficiency of the path taken by the ViT model during fine-tuning, a score of 1 implies a perfectly straight path from pre-trained to fine-tuned model versus a score closer to 0 implies a very inefficient path. A visualization of the quantity we measure is shown in Fig. \ref{fig:imageftd1}. We hypothesize that the relative displacement of the attention layer parameters of MAE would be low if the layers quickly converge towards new parameters instead of converging in a zig-zag fashion. On the other hand we expect the \contrastive attention layers to remain relatively stationary in the parameter space since they do not show dramatic performance changes in transfer upon fine-tuning. We indeed find that the relative displacement of the MAE pre-trained model attention layers is the noticeably lower than the MoCo-V3 and DINO models when we observe the fine-tuning dynamics of the attention layers in Fig. \ref{fig:imageftd2}.

\paragraph{Does representational similarity translate to functional similarity for finetuned MAEs?}   
\label{sec:res-ft-layers-func}

How does the information distribution across layers change during fine-tuning? To answer this, we repeat the experiments from Section \ref{sec:res-transfer} on a fine-tuned MAE model to visualize how the invariances learned in the final \texttt{CLS} token features change, as well as to quantify the marginal utility of information present in the intermediate layers versus final layer. 

In Fig. \ref{fig:rcdm_mae_ft}, we visualize the samples from an RCDM trained on the final \texttt{CLS} token features of a fine-tuned MAE model. We observe that the samples generated from a particular horizontal orientation of the object do not preserve this information i.e. during fine-tuning the ViT learns to be invariant to horizontal flip. Thus, fine-tuning enables MAE pre-trained ViTs to learn invariances that are similar to pre-trained \contrastive models and informative for classification.

\sidecaptionvpos{figure}{c}
\begin{SCfigure}[][t]
  \includegraphics[height=5cm]{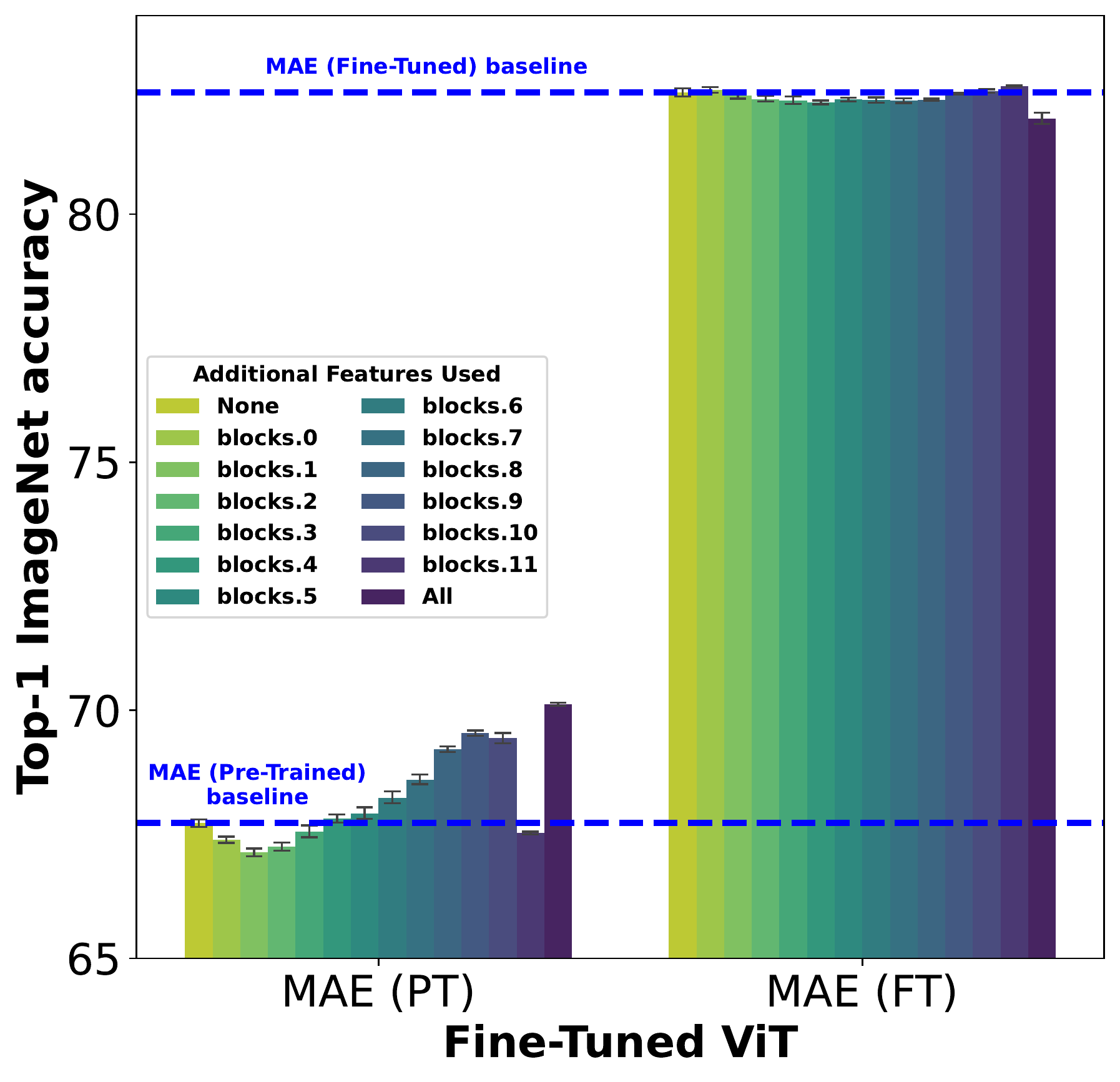}
  \captionsetup{width=\textwidth}
\caption{Using additional intermediate layer \texttt{CLS} token features for linear probe transfer from MAE before vs after fine-tuning (mean ± std for best 5 runs). \textbf{Unlike a pre-trained MAE, intermediate features provide no additional information for a fine-tuned MAE, thus demonstrating that fine-tuning restructures information to be more readily available in the last layer \texttt{CLS} token features.}\looseness=-1}
  \label{fig:mae_ft}
\end{SCfigure}

%\begin{SCfigure}[]
%  \includegraphics[height=4cm]{figures/ICML/arxiv_mae_ft_marginal_features.pdf}
% \caption{Using additional intermediate layer \texttt{CLS} token features for linear probe transfer from MAE before vs after fine-tuning (mean ± std for best 5 runs). \textbf{Unlike a pre-trained MAE, intermediate features provide no additional information for a fine-tuned MAE, thus demonstrating that fine-tuning restructures information to be more readily available in the last layer \texttt{CLS} token features.}\looseness=-1}
%  \label{fig:mae_ft}
%\end{SCfigure}

%\begin{figure}
%\centering
%    \includegraphics[width=0.5\columnwidth]{figures/ICML/mae_ft_marginal_features.pdf}
%    \caption{Using additional intermediate layer \texttt{CLS} token features for linear probe transfer. \textbf{Unlike a pre-trained MAE, intermediate features provide no additional information for a fine-tuned MAE, thus demonstrating that fine-tuning restructures information to be more readily available in the last layer \texttt{CLS} token features.}}

%\label{fig:mae_ft}
%\end{figure}

In Fig. \ref{fig:mae_ft}, we show that the marginal utility of training a linear probe on the intermediate \texttt{CLS} token features in a fine-tuned MAE. Unlike a pre-trained MAE (Fig. \ref{fig:marginal}) where not all class specific information was available in the final layer features, we find that the fine-tuned MAE does not perform any better when a linear probe is additionally trained on its intermediate features. Thus, fine-tuning with supervision leads to a re-organization of information in the ViT layers, and the class discerning information becomes readily available in the final \texttt{CLS} features.

\section{Discussion}

\paragraph{Conclusion} We analyzed ViT representations and their transferability when trained via two popular self-supervised approaches: (1) Joint-Embedding (JE) methods (MoCo-V3, DINO), and (2) Reconstruction-Based (REC) methods (MAE). We reveal key differences learned across both representations and how these differences are localized by layer types while being distributed across network depth. We explained why \contrastive models transfer better with a linear probe, as their final layer \texttt{CLS} tokens contain all pertinent information for class discriminative learning. We also presente ways to extract the relevant information distributed across \reconstruction layers \emph{without fine-tuning}. 
% We also demonstrated downstream transfer tasks where the local-scale \reconstruction features transfer better than \contrastive features without fine-tuning.
Finally, we show how fine-tuning modifies \reconstruction features to be more linearly decodable by re-organizing class information into the final layer. 
% Finally, we explained how \reconstruction methods transfer better than \contrastive methods on fine-tuning by analyzing changes to \reconstruction models during and after-tuning from the lens of representation, training dynamics, and feature invariances. %Through this work, we have addressed how the representations across \contrastive and \reconstruction methods differ, and also de-mystified the differences in transfer performance across both methods.

\paragraph{Limitations and Future Work} Our pre-training dataset, ImageNet is a balanced large-scale dataset, SSL ViT methods have demonstrated poor empirical performance and transfer when trained on imbalanced datasets \citep{assran2022hidden}. We also focused on understanding the ViT-Base model representations in this study. Seeing how different SSL pre-training methods scale with model size and dataset size and diversity is an interesting avenue for future research. Another potential future study could look into quantifying the notion behind `information' available in SSL representations from a mathematical perspective instead of our treatment of representational information as its impact on downstream transfer. It would also be interesting to see how both \contrastive and \reconstruction representations transfer to other downstream tasks beyond classification, detection, and segmentation. Lastly, we are very interested in exploring the impact of SSL objectives on multi-modal representation learning methods such as CLIP \citep{radford2021learning} and Omni-MAE \citep{girdhar2022omnimae}.

% Note use of \abovespace and \belowspace to get reasonable spacing
% above and below tabular lines.

%\clearpage
\bibliography{references}
%\bibliographystyle{unsrtnat}
%\bibliographystyle{icml2023}

%%%%%%%%%%%%%%%%%%%%%%%%%%%%%%%%%%%%%%%%%%%%%%%%%%%%%%%%%%%%%%%%%%%%%%%%%%%%%%%
%%%%%%%%%%%%%%%%%%%%%%%%%%%%%%%%%%%%%%%%%%%%%%%%%%%%%%%%%%%%%%%%%%%%%%%%%%%%%%%
% APPENDIX
%%%%%%%%%%%%%%%%%%%%%%%%%%%%%%%%%%%%%%%%%%%%%%%%%%%%%%%%%%%%%%%%%%%%%%%%%%%%%%%
%%%%%%%%%%%%%%%%%%%%%%%%%%%%%%%%%%%%%%%%%%%%%%%%%%%%%%%%%%%%%%%%%%%%%%%%%%%%%%%
\newpage

\appendix
\renewcommand{\thefigure}{A\arabic{figure}}
\setcounter{figure}{0}
\renewcommand\thetable{A\arabic{table}}
\setcounter{table}{0}
\onecolumn

\section{Additional Experimental Details}
\label{app:details}

\subsection{Mini-batch CKA details}
\label{app:details-arch}

The CKA value between two $p_1$ and $p_2$ dimensional representational matrices of $m$ examples $\mathbf{X} \in \mathbb{R}^{m \times p_1}$ and $\mathbf{Y} \in \mathbb{R}^{m \times p_2}$ is the normalized Hilbert-Smith Independence Criteria \citep{gretton2007kernel} of the Gram similarity matrices $\mathbf{K=XX}^T$ and $\mathbf{L=YY}^T$ given as:

\begin{equation}
\operatorname{CKA}(\mathbf{K}, \mathbf{L})=\frac{\mathrm{HSIC}(\mathbf{K}, \mathbf{L})}{\sqrt{\mathrm{HSIC}(\mathbf{K}, \mathbf{K}) \ \mathrm{HSIC}(\mathbf{L}, \mathbf{L})}}
\end{equation}

We adapt the formalization from \citep{nguyen2020wide} which approximates the linear CKA metric by averaging over $k$ minibatches to obtain the minibatch CKA metric. Minibatch CKA over two sets of activation matrices $\mathbf{X}_i \in \mathbb{R}^{n \times p_1}$ and $\mathbf{Y}_i \in \mathbb{R}^{n \times p_2}$ of the $i^{th}$ minibatch of $n$ examples is given as:

\begin{equation}
    \mathrm{CKA}_{\text {minibatch }}=\frac{\frac{1}{k} \sum_{i=1}^k \mathrm{HSIC}_1\left(\mathbf{X}_i \mathbf{X}_i^{\top}, \mathbf{Y}_i \mathbf{Y}_i^{\top}\right)}{\sqrt{\frac{1}{k} \sum_{i=1}^k \mathrm{HSIC}_1\left(\mathbf{X}_i \mathbf{X}_i^{\top}, \mathbf{X}_i \mathbf{X}_i^{\top}\right)} \sqrt{\frac{1}{k} \sum_{i=1}^k \mathrm{HSIC}_1\left(\mathbf{Y}_i \mathbf{Y}_i^{\top}, \mathbf{Y}_i \mathbf{Y}_i^{\top}\right)}}
\end{equation}

where HSIC\textsubscript{1} is an unbiased estimator of the Hilbert-Smith Independence Criteria such that the CKA value is independent of batch size. The HSIC\textsubscript{1} between two similarity matrices $\mathbf{K}$ and $\mathbf{L}$ ($\tilde{\mathbf{K}}$ and $\tilde{\mathbf{L}}$ are obtained by setting the respective diagonal entries to zeros) is given as:

\begin{equation}
    \mathrm{HSIC}_1(\mathbf{K}, \mathbf{L})=\frac{1}{n(n-3)}\left(\operatorname{tr}(\tilde{\mathbf{K}} \tilde{\mathbf{L}})+\frac{\mathbf{1}^{\top} \tilde{\mathbf{K}} \mathbf{1 1}{ }^{\top} \tilde{\mathbf{L}} \mathbf{1}}{(n-1)(n-2)}-\frac{2}{n-2} \mathbf{1}^{\top} \tilde{\mathbf{K}} \tilde{\mathbf{L}} \mathbf{1}\right)
\end{equation}

\begin{figure}
 \captionsetup{justification=centering}

\centering
\begin{subfigure}{.45\columnwidth}
    \centering
    \includegraphics[width=\columnwidth]{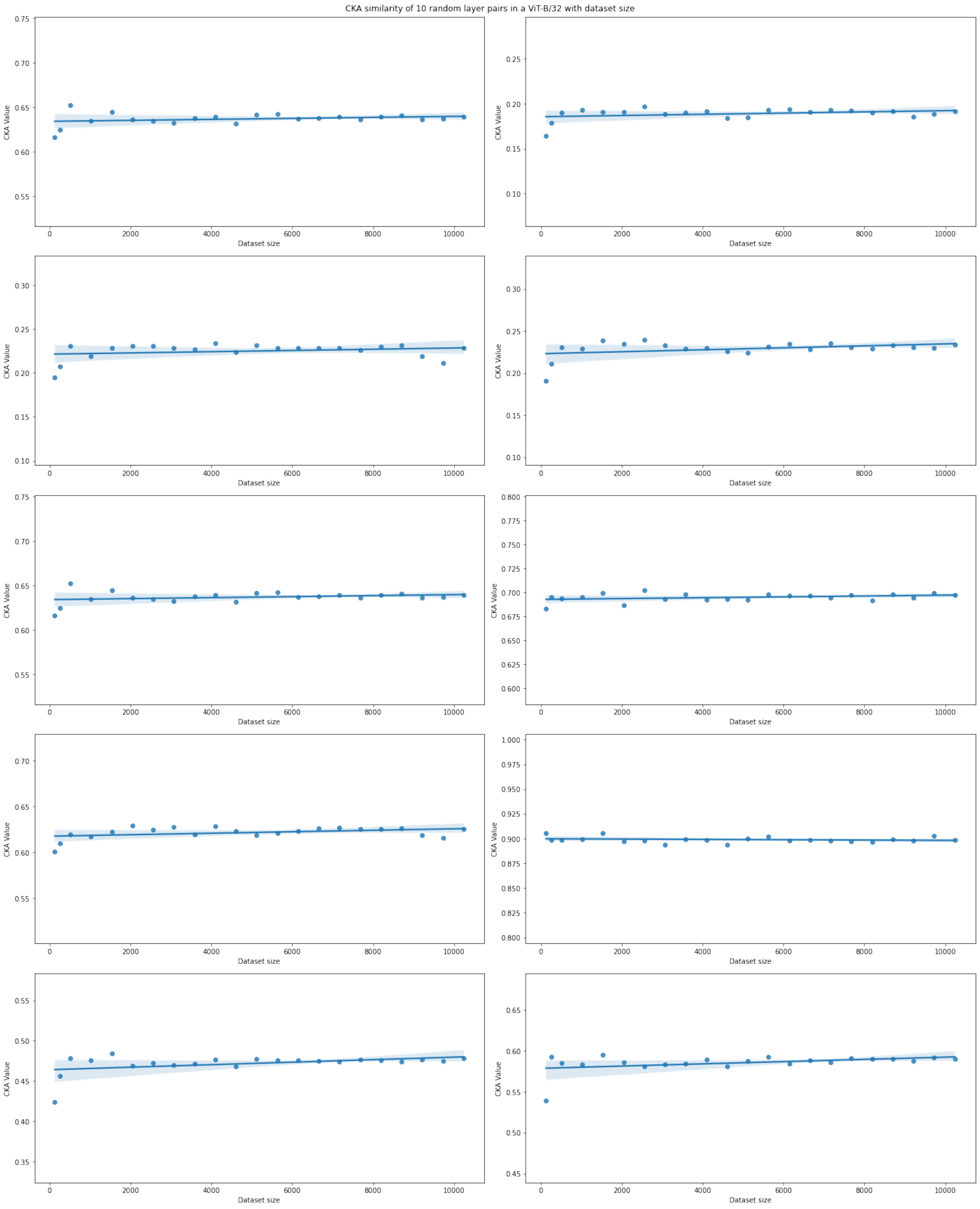}
    \caption{Minibatch CKA vs samples used}
    \label{fig:minicka1}
\end{subfigure}
\hfill
\begin{subfigure}{.45\columnwidth}
    \centering
    \includegraphics[width=\columnwidth]{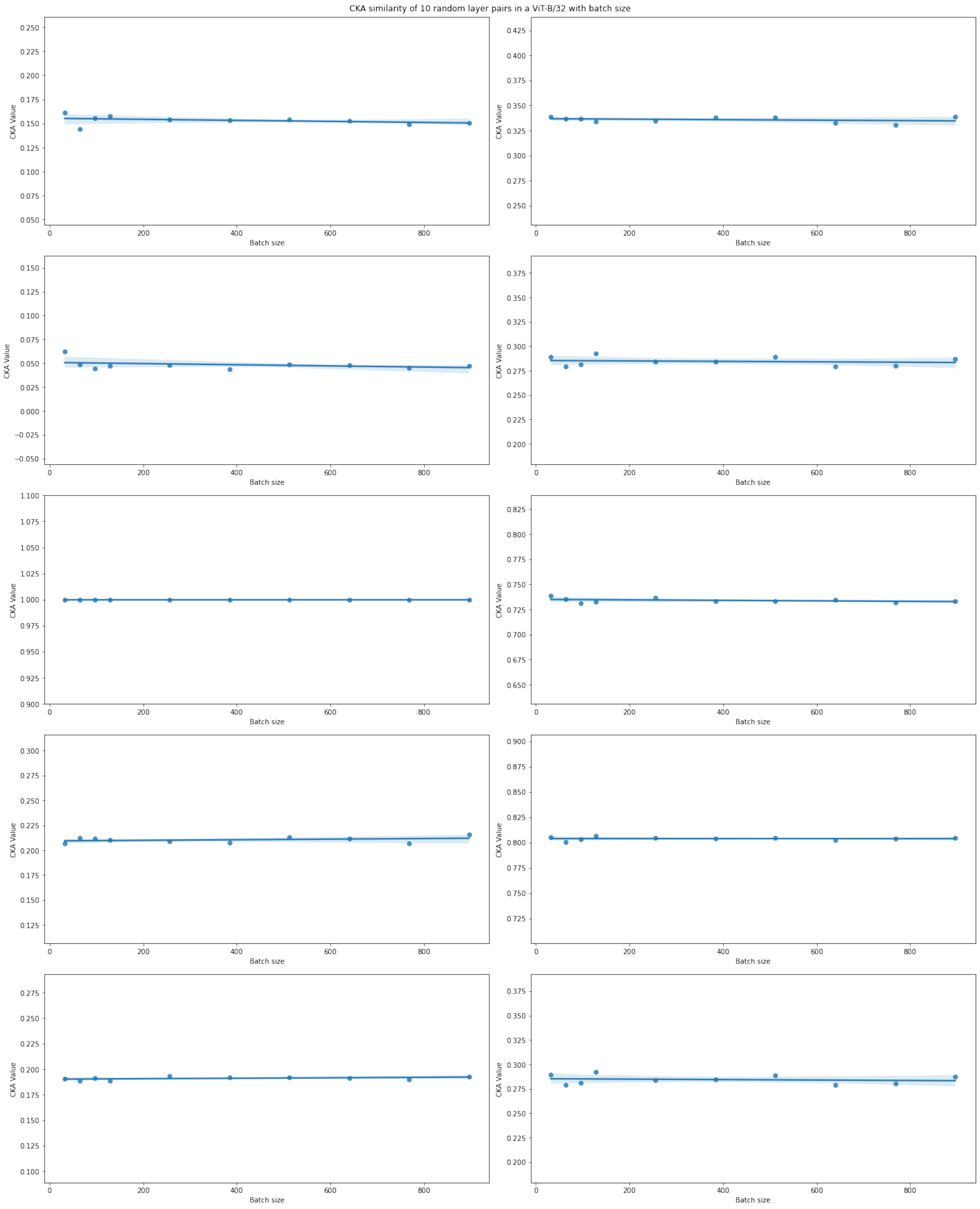}
    \caption{Minibatch CKA vs batch size}
    \label{fig:minicka2}
\end{subfigure}
\caption{Comparisons showing the impact of batch size and number of data samples used to compute minibatch CKA values across a random subset of 10 layers in ViT-B. Minibatch CKA values remain consistent above batch sizes of $2^5$ and sample sizes of 1024.}
\label{fig:mini-batch-cka}
\end{figure}

 For our mini-batch CKA computations, we use a batch size of 32 and sample a total of 1024 examples without replacement for computing the representations. Like \cite{raghu2021vision}, we compared our mini-batch CKA values across a large range of mini-batch sizes ($2^5$ to $2^{10}$) as well as a large range of examples ($10^3$ to $10^6$) and found no noticeable differences (Fig. \ref{fig:mini-batch-cka}).

\subsection{SSL pre-training details}
\label{app:details-pretrain}

For each of MoCo-V3 \cite{chen2020simple}, DINO \citep{caron2021emerging}, MAE \cite{he2022masked} we utilize pre-trained models provided by the original authors with the exact pre-training setup as mentioned in the original papers. We summarize these detials in Table \ref{app:pt_mae_details} for \reconstruction model MAE, and in Tables \ref{app:pt_moco_details} and \ref{app:pt_dino_details} for \contrastive models MoCo-V3 and DINO respectively. We pretrain replicate models for all three ourself to verify that our observations also hold on replicates. During this pre-training, Linear \textit{lr scaling} rule is used for large batch training where = \textit{lr = base.lr x batch size / 256}.

\begin{table}[t]
\centering
\begin{tabular}{l|l} 
Config & Value \\
\hline optimizer & AdamW \\
base learning rate & $1.5 \mathrm{e}-4$ \\
weight decay & $0.05$ \\
optimizer momentum & $\beta_1, \beta_2=0.9,0.95$ \\
batch size & 4096 \\
learning rate schedule & cosine decay \\
warmup epochs & $40$ \\
training epochs & $800$ \\
augmentation & RendomResizedCrop
\end{tabular}
\caption{Pre-Training Details for MAE.}\label{app:pt_mae_details}
\end{table} 

\begin{table}[t]
\setlength\extrarowheight{-3pt}
\captionsetup{justification=centering}
\centering
\resizebox{0.65\columnwidth}{!}{%
\begin{tabular}{l|l} 
Config & Value \\
\hline optimizer & AdamW \\
base learning rate & $1.5 \mathrm{e}-4$ \\
weight decay & $0.1$ \\
optimizer momentum & $\beta_1, \beta_2=0.9,0.95$ \\
batch size & 4096 \\
learning rate schedule & cosine decay \\
warmup epochs & $40$ \\
training epochs & $300$ \\
\begin{tabular}{@{}c@{}}momentum encoder\\momentum \end{tabular} & $0.99$ \\
\begin{tabular}{@{}c@{}}momentum rate\\schedule \end{tabular} & cosine \\
augmentation & \begin{tabular}{@{}c@{}}RandomResizedCrop \\ ColorJitter \\ RandomGrayscale \\ GaussianBlur \\ Solarize \\ RandomHorizontalFlip\end{tabular}  \\
\end{tabular}%
}
\captionof{table}{Pre-training details for MoCo-V3}
\label{app:pt_moco_details}

%\caption{Hyper-parameter settings for linear probe transfer and fine-tuning transfer for supervised ImageNet classification.}
%\label{app:transfer_details}
\end{table}

\begin{table}[ht!]
\setlength\extrarowheight{-3pt}
\captionsetup{justification=centering}
\centering
\resizebox{0.65\columnwidth}{!}{%
\begin{tabular}{l|l} 
Config & Value \\
\hline optimizer & AdamW \\
base learning rate & $5 \mathrm{e}-4$ \\
weight decay & $0.04$ \\
optimizer momentum & $\beta_1, \beta_2=0.9,0.95$ \\
batch size & 4096 \\
learning rate schedule & cosine decay \\
warmup epochs & $10$ \\
training epochs & $300$ \\
teacher momentum & $0.996$ \\
teacher temperature & $0.07$ \\
\begin{tabular}{@{}c@{}}teacher temperature\\warmup epochs \end{tabular} & $30$ \\
augmentation & \begin{tabular}{@{}c@{}} RandomResizedCrop-\\-Multi (96x96 and 224x244) \\ ColorJitter \\ RandomGrayscale \\ GaussianBlur \\ Solarize \\ RandomHorizontalFlip\end{tabular}  \\
\end{tabular}%
}
\captionof{table}{Pre-training details for DINO}
\label{app:pt_dino_details}
%\caption{Hyper-parameter settings for linear probe transfer and fine-tuning transfer for supervised ImageNet classification.}
%\label{app:transfer_details}
\end{table}

\begin{table}[t]
\setlength\extrarowheight{-3pt}
\captionsetup{justification=centering}
\centering
\resizebox{0.65\columnwidth}{!}{%
\begin{tabular}{l|l} 
Config & Value \\
\hline optimizer & LARS \\
base learning rate & $\{0.1,1 \mathrm{e}-2, 1 \mathrm{e}-3\}$ \\
weight decay & $\{0, 5 \mathrm{e}-2, 0.1\}$ \\
L-1 regularization $\alpha$ & $\{0, 1\mathrm{e}-\{1,2,3,4\}, 5 \mathrm{e}-4\}$ \\
optimizer momentum & $0.9$ \\
batch size & 4096 \\
learning rate schedule & cosine decay \\
warmup epochs & $\{10, 40\}$ \\
training epochs & $\{100, 200\}$ \\
augmentation & RandomResizedCrop \\
\end{tabular}%
}
\captionof{table}{Linear and Non-Linear Probe Transfer details. A hyper-parameter grid search was performed on the cross-product of config values within \{\}.}
\label{table:linprobe}
\centering
\end{table}

\subsection{Linear Probe details}
\label{app:details-linprobe}

Details of our linear probe transfer settings are given in Table \ref{table:linprobe}. Similar to \citet{he2022masked}, we utilize an extra BatchNorm layer without affine transformation before the linear classifier to calibrate feature magnitudes across different layer features for our experiments involving intermediate features. We perform extensive hyper-parameter sweeps by performing a grid search over cross product of values given in Table \ref{table:linprobe}, and report the mean and standard deviation in accuracy for the 5 best performing models in each experiment.

\subsection{Non-Linear Probe details}
\label{app:details-nonlin}

For our experiments with non-linear probes trained on the last layer \texttt{CLS} token features, each non-linear probe layer block is made up of a linear layer (same as linear probe), followed by a BatchNorm layer, followed by a non-linear ReLU activation. The training related hyperparameters remain the same as given in Table \ref{table:linprobe}. Similar to Section \ref{app:details-linprobe}, we perform extensive hyper-parameter sweeps, and report the mean and standard deviation in accuracy for the 5 best performing models in each experiment.

\begin{table}[t]
\setlength\extrarowheight{-3pt}
\captionsetup{justification=centering}
\centering
\resizebox{0.65\columnwidth}{!}{%
\begin{tabular}{l|l} 
Config & Value \\
\hline optimizer & AdamW \\
base learning rate & $1 \mathrm{e}-3$ \\
weight decay & $0.05$ \\
optimizer momentum & $\beta_1, \beta_2=0.9,0.999$ \\
layer-wise lr decay $[10,2]$ & $0.75$ \\
batch size & 1024 \\
learning rate schedule & cosine decay \\
warmup epochs & 5 \\
training epochs & $100$ \\
augmentation & RandAug $(9,0.5)[12]$ \\
label smoothing [52] & $0.1$ \\
mixup [69] & $0.8$ \\
cutmix [68] & $1.0$ \\
drop path [30] & $0.1$
\end{tabular}%
}
\captionof{table}{Fine-Tuning Transfer details}
\label{table:finetuning}
%\caption{Hyper-parameter settings for linear probe transfer and fine-tuning transfer for supervised ImageNet classification.}
%\label{app:transfer_details}
\end{table}

\subsection{Fine-Tuning details}
\label{app:details-ft}

Details of our fine-tuning transfer settings are given in Table \ref{table:finetuning}.

\subsection{Alternate neural architecture details}
\label{app:details-cnnarch}

For our comparisons of representation similarity across types of architectures, we require convolutional models pre-trained with similar objectives as our ViT models. For this purpose, we take a standard ResNet50 \citep{he2016deep} pre-trained with DINO and MoCo-V3 objectives as our candidate \contrastive CNN model, and a ConvNextv2-Base \citep{woo2023convnext} pre-trained with MAE objective as our candidate \reconstruction CNN model. The readers may refer to the original papers for exact model specifications.

\subsection{MS COCO Object Detection and Segmentation}
\label{app:details-coco}

We utilize the ViTDet framework introduced by \citet{li2022exploring}, which uses the final \texttt{CLS} token features and then uses strided convolutions and deconvolutions to upsample/downsample the single-scale features into a simple hierarchical feature pyramid. The feature pyramid generate uses strides of 4, 8, 16, and 32-consistent with ResNet based detection/segmentation models.

Once this feature pyramid is built from a ViT backbone, a standard Mask R-CNN \citep{he2017mask} is applied on top of the feature pyramid to perform bounding box regression, classification, as well as instance segmentation. In order to evaluate the utility of pre-trained ViT representations for detection and segmentation, we keep the backbone model parameters frozen when we train our Mask R-CNN in Section \ref{sec:res-transfer}. Further detection and segmentation results are provide in Section \ref{app:seg_det}.

\clearpage

\section{Segmentation and Detection Results} \label{app:seg_det}

As mentioned in Section \ref{sec:res-transfer}, MAE performs better than MoCo-V3 and DINO (without multi-crop training) when performing segmentation and detection on MS COCO with a frozen pre-trained backbone ViT. The results are repeated below in Table \ref{table:coco}, where we also include results from a DINO model trained with multi-crops.

\begin{table}[ht!]
\centering
\resizebox{\columnwidth}{!}{%
\begin{tabular}{|l|lll|lll|}
\hline
\multirow{2}{*}{} & \multicolumn{3}{c|}{\textbf{Detection}}                               & \multicolumn{3}{c|}{\textbf{Segmentation}}                             \\ \cline{2-7} 
                  & \multicolumn{1}{l|}{AP}    & \multicolumn{1}{l|}{AP\textsuperscript{large}} & AP\textsuperscript{small} & \multicolumn{1}{l|}{AP}    & \multicolumn{1}{l|}{AP\textsuperscript{large}} & AP\textsuperscript{small} \\ \hline
MAE      & \multicolumn{1}{l|}{30.25} & \multicolumn{1}{l|}{39.93}    & 18.69    & \multicolumn{1}{l|}{28.56} & \multicolumn{1}{l|}{41.73}     & 14.37    \\ \hline
MoCo-V3     & \multicolumn{1}{l|}{28.75} & \multicolumn{1}{l|}{40.21}    & 15.49    & \multicolumn{1}{l|}{26.67} & \multicolumn{1}{l|}{41.92}     & 10.89    \\ \hline
DINO     & \multicolumn{1}{l|}{32.57} & \multicolumn{1}{l|}{44.14}    & 19.89    & \multicolumn{1}{l|}{30.04} & \multicolumn{1}{l|}{45.77}     & 14.37    \\ \hline
DINO (w/o multi-crop)     & \multicolumn{1}{l|}{29.97} & \multicolumn{1}{l|}{40.82}    & 17.55    & \multicolumn{1}{l|}{28.16} & \multicolumn{1}{l|}{42.53}     & 12.99    \\ \hline
\end{tabular}%
}
\caption{Downstream transfer for object detection and segmentation on MS COCO using a ViTDet based Mask R-CNN with a frozen backbone ViT (same as Table \ref{table:coco_main} but includes results from DINO with multi-crop training).}
\label{table:coco}
\end{table}

For completeness, we also provide the corresponding detection and segmentation results when the Mask R-CNN with a ViT backbone is fine-tuned end-to-end. With supervised fine-tuning, the discrepancy noted in the detection and segmentation for \reconstruction models on larger objects vanishes, since they acquire global context and perform better on AP\textsuperscript{large} versus \contrastive models.

\begin{table}[ht!]
\centering
\resizebox{\columnwidth}{!}{%
\begin{tabular}{|l|lll|lll|}
\hline
\multirow{2}{*}{} & \multicolumn{3}{c|}{\textbf{Detection}}                               & \multicolumn{3}{c|}{\textbf{Segmentation}}                             \\ \cline{2-7} 
                  & \multicolumn{1}{l|}{AP}    & \multicolumn{1}{l|}{AP\textsuperscript{large}} & AP\textsuperscript{small} & \multicolumn{1}{l|}{AP}    & \multicolumn{1}{l|}{AP\textsuperscript{large}} & AP\textsuperscript{small} \\ \hline
MAE      & \multicolumn{1}{l|}{51.57} & \multicolumn{1}{l|}{66.36}    & 35.27    & \multicolumn{1}{l|}{45.84} & \multicolumn{1}{l|}{63.84}     & 27.27    \\ \hline
MoCo-V3     & \multicolumn{1}{l|}{48.81} & \multicolumn{1}{l|}{64.83}    & 32.79    & \multicolumn{1}{l|}{43.18} & \multicolumn{1}{l|}{62.86}     & 23.78    \\ \hline
DINO     & \multicolumn{1}{l|}{47.73} & \multicolumn{1}{l|}{62.49}    & 32.17    & \multicolumn{1}{l|}{30.04} & \multicolumn{1}{l|}{60.03}     & 24.28    \\ \hline

\end{tabular}%
}
\caption{Downstream transfer for object detection and segmentation on MS COCO using a ViTDet based Mask R-CNN with a backbone ViT and end-to-end fine-tuning.\textbf{With fine-tuning, MAE outperforms both MoCo-V3 and DINO on both large and small objects, implying that \reconstruction models learn global scale features with fine-tuning.}}
\label{table:coco_ft}
\end{table}
\clearpage

\section{Additional CKA plots}
\label{app:cka}

\subsection{CKA between pre-trained and fine-tuned MAE and DINO}

For completeness, we provide the plots comparing CKA between DINO and MAE both before and after fine-tuning, analogous to Fig. \ref{fig:imageft} in the main text with MoCo-V3.

\begin{figure}[ht]
\centering
\begin{subfigure}{.4\linewidth}
    \centering
    \includegraphics[width=\linewidth]{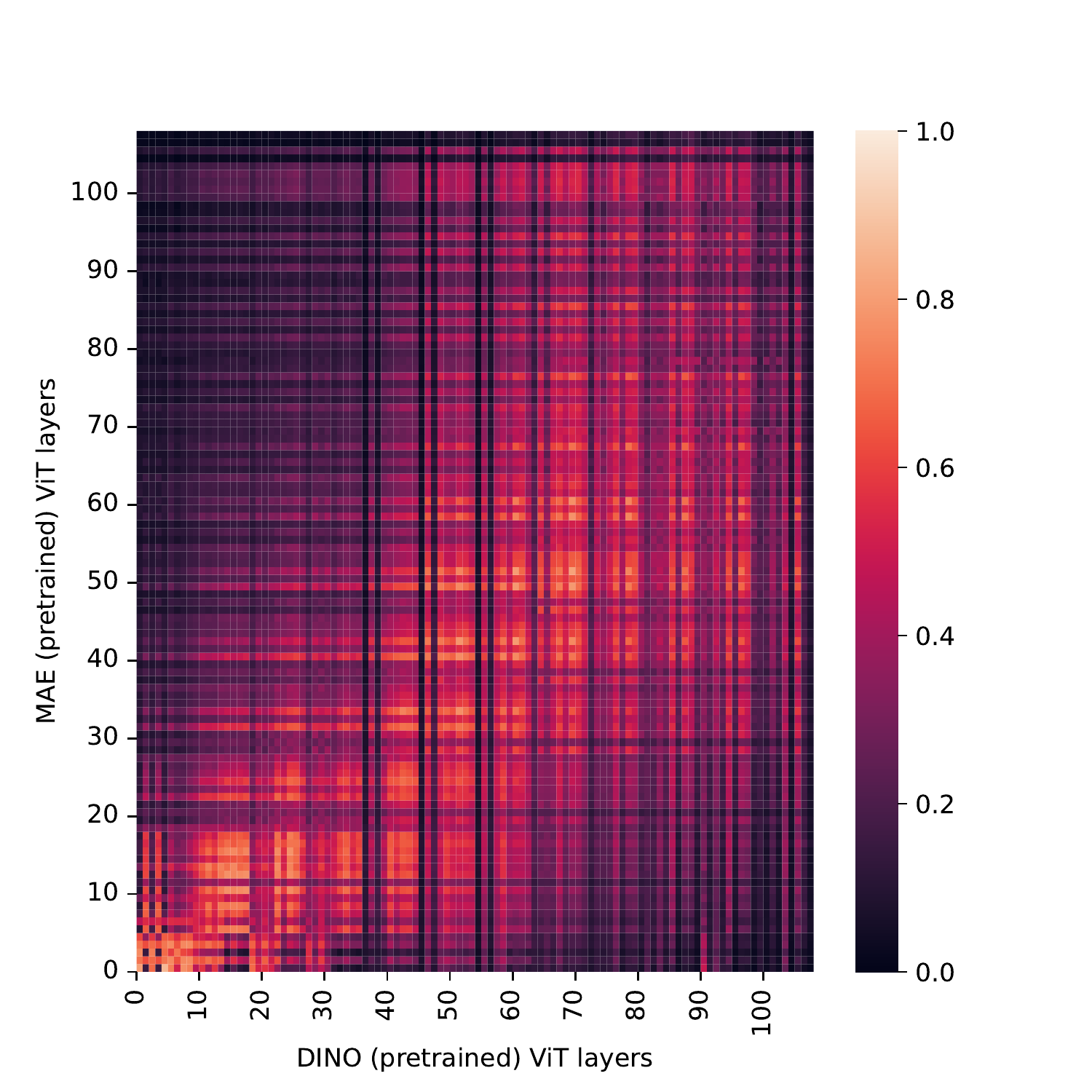}
    \caption{DINO (PT) \& MAE (PT)}
    \label{fig:imagefta1}
\end{subfigure}
\begin{subfigure}{.4\linewidth}
    \centering
    \includegraphics[width=\linewidth]{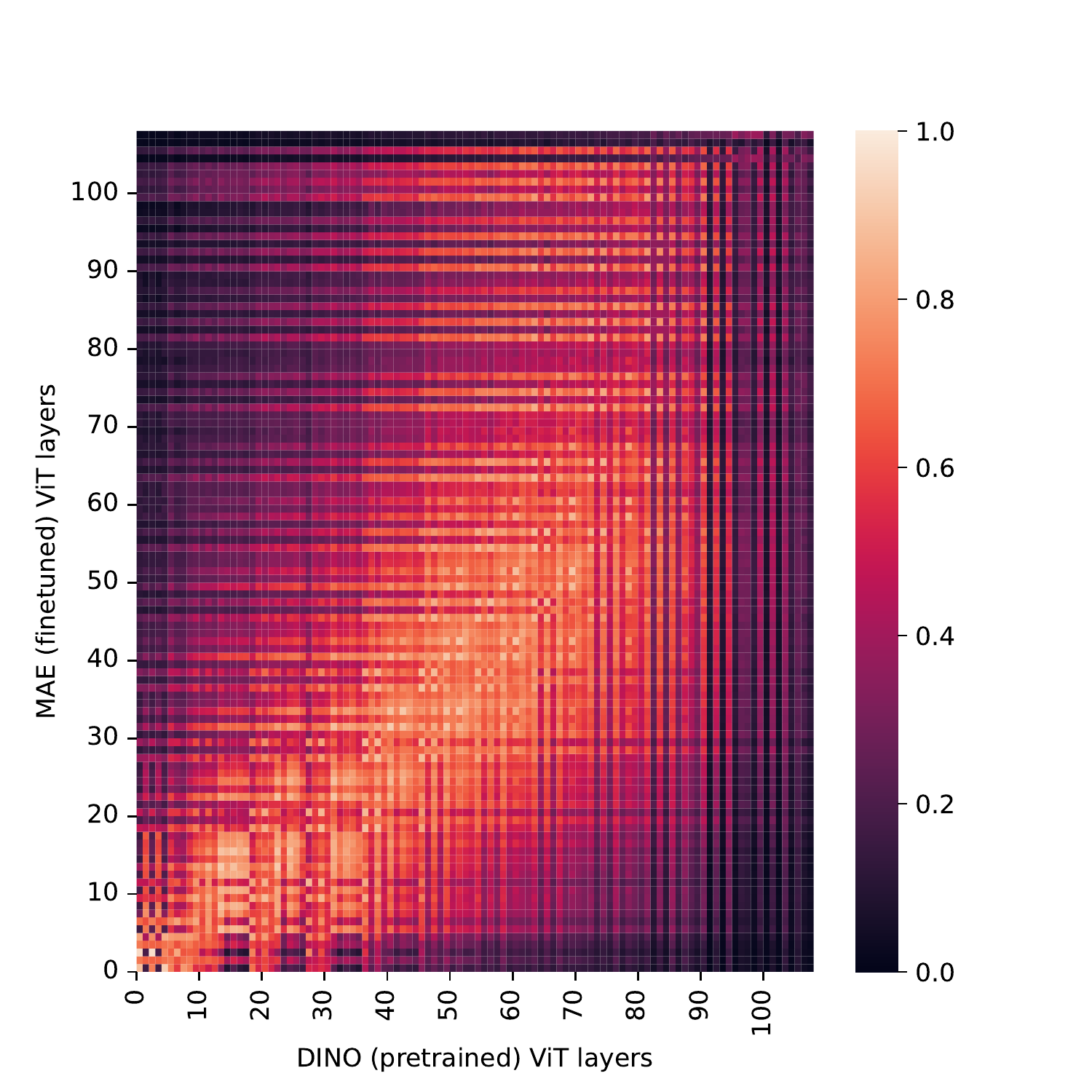}
    \caption{DINO (PT) \& MAE (FT)}
    \label{fig:imagefta2}
\end{subfigure}
\begin{subfigure}{.4\linewidth}
    \centering
    \includegraphics[width=\linewidth]{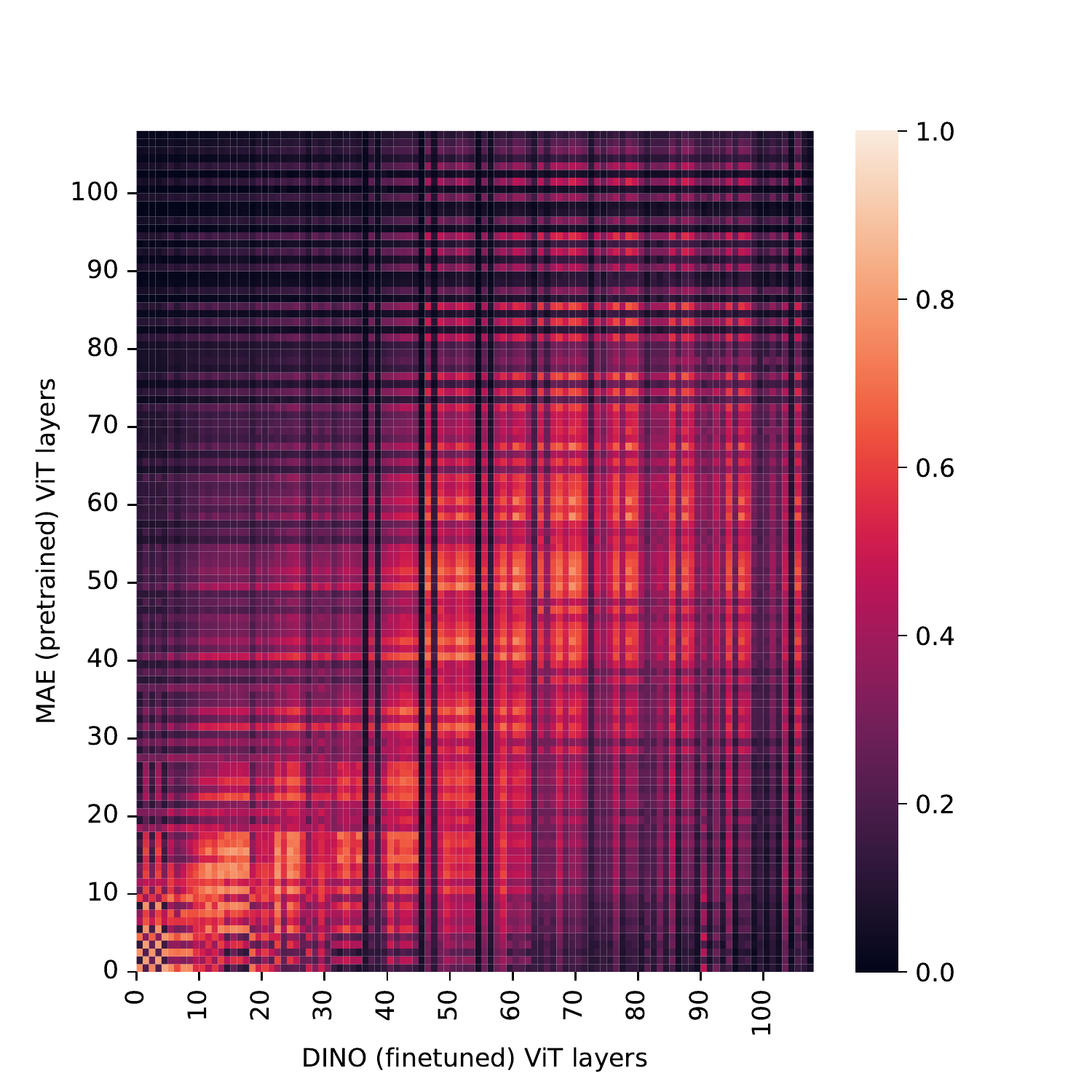}
    \caption{DINO (FT) \& MAE (PT)}
    \label{fig:imagefta4}
\end{subfigure}
\begin{subfigure}{.4\linewidth}
    \centering
    \includegraphics[width=\linewidth]{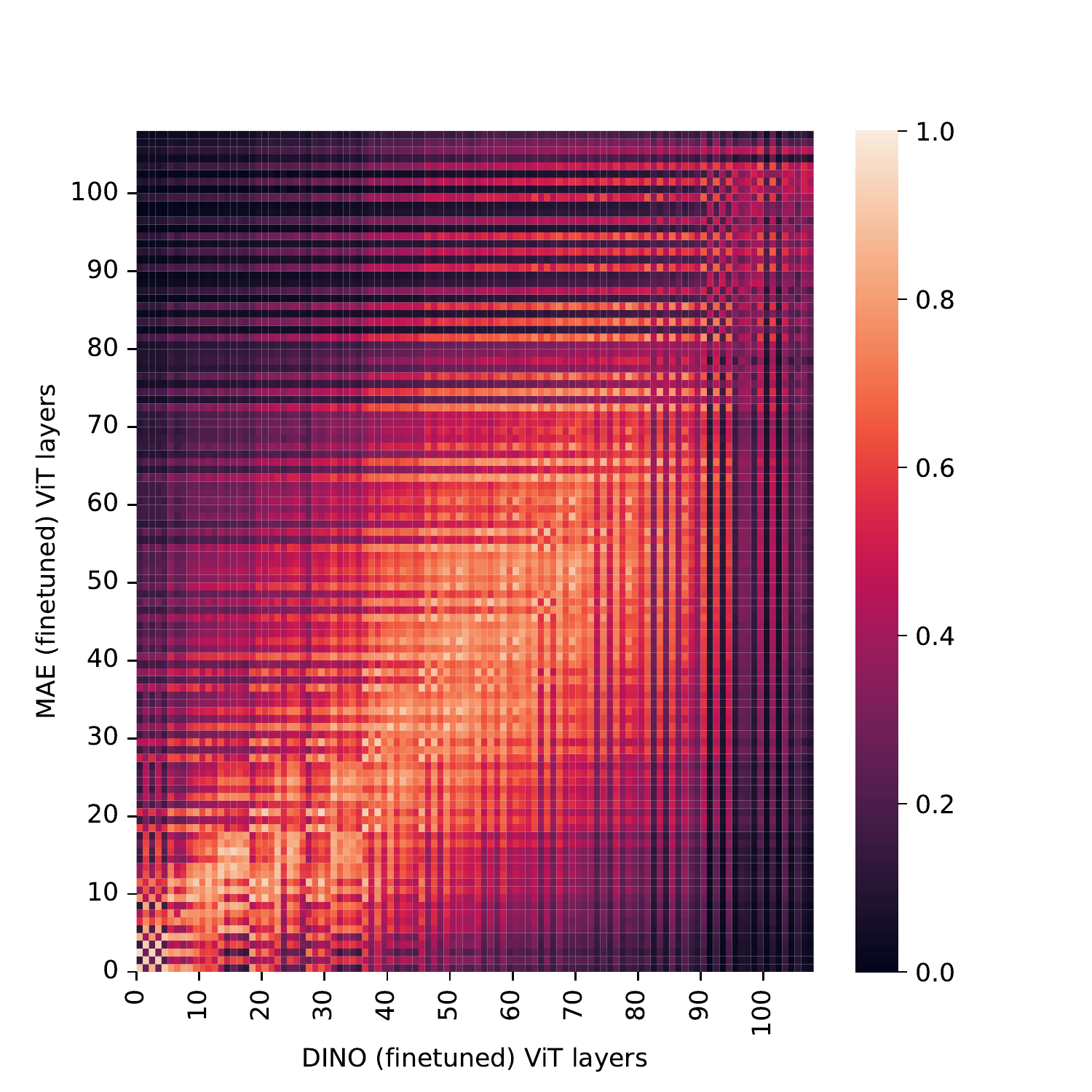}
    \caption{DINO (FT) \& MAE (FT)}
    \label{fig:imagefta5}
\end{subfigure}
\setlength{\belowcaptionskip}{-0.5cm}
\caption{CKA similarity between DINO and MAE before (PT) and after fine-tuning (FT). Similar to the MoCo-V3 comparisons \ref{fig:imageft}, an MAE (FT) ViT-B becomes very similar to a DINO (PT), (\ref{fig:imagefta4}), and the similarity persists with the DINO (FT) ViT-B/16 (\ref{fig:imagefta5}).}
\label{fig:imagefta}
\end{figure}

\clearpage

\subsection{CKA between pre-trained and fine-tuned \Contrastive and \Reconstruction models by layer type}

\begin{figure}[ht]
\centering
\begin{subfigure}{.3\linewidth}
    \centering
    \includegraphics[width=\linewidth]{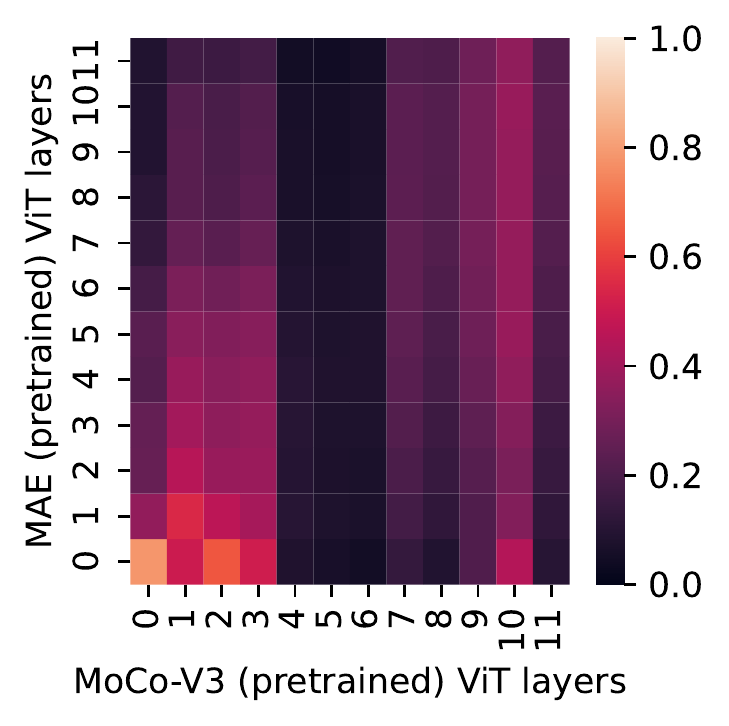}
    \caption{Layer Norm (PT)}
    \label{fig:imagel1}
\end{subfigure}
    \hfill
\begin{subfigure}{.3\linewidth}
    \centering
    \includegraphics[width=\linewidth]{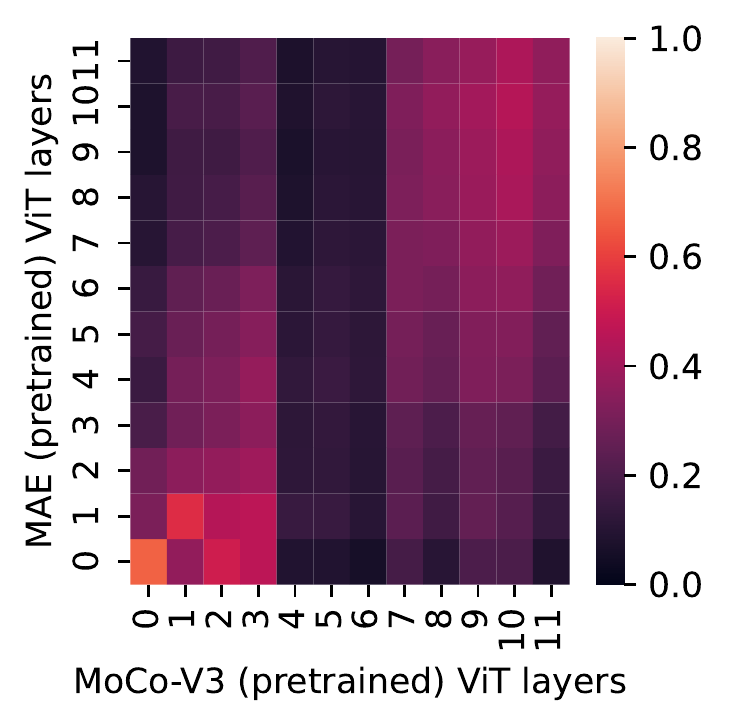}
    \caption{Self-Attention (PT)}\label{fig:imagel2}
\end{subfigure}
   \hfill
\begin{subfigure}{.3\linewidth}
    \centering
    \includegraphics[width=\linewidth]{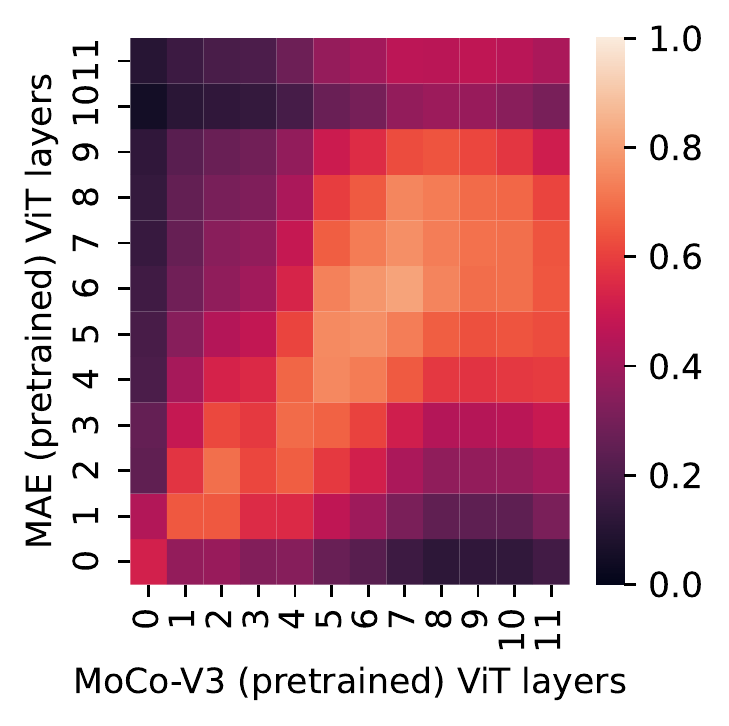}
    \caption{Fully-Connected (PT)}\label{fig:imagel3}
\end{subfigure}
\begin{subfigure}{.3\linewidth}
    \centering
    \includegraphics[width=\linewidth]{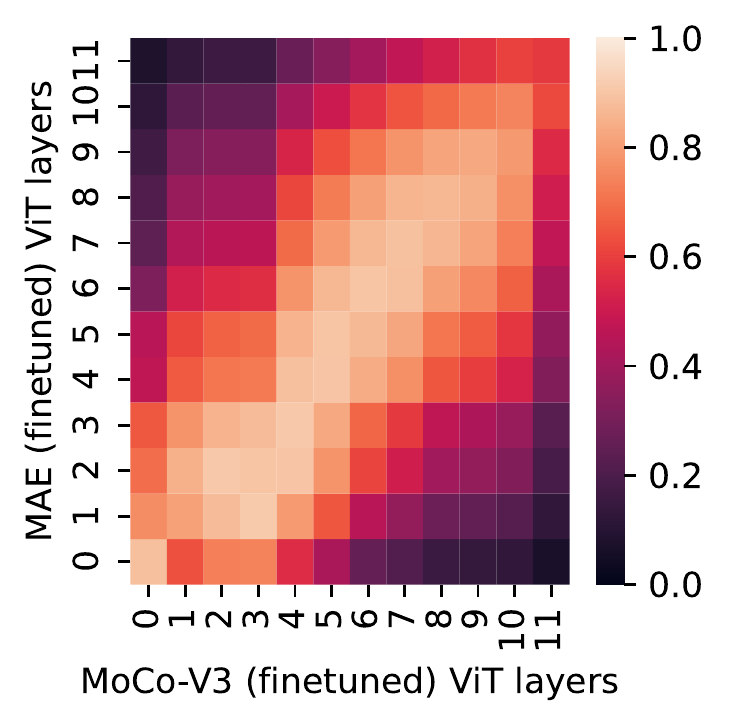}
    \caption{Layer Norm  (FT)}
    \label{fig:imagel4}
\end{subfigure}
    \hfill
\begin{subfigure}{.3\linewidth}
    \centering
    \includegraphics[width=\linewidth]{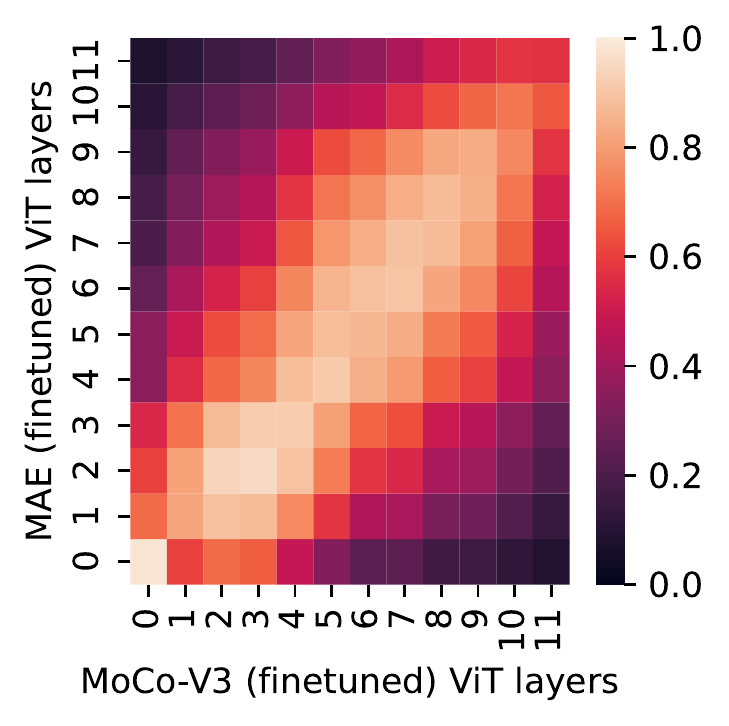}
    \caption{Self-Attention (FT)}\label{fig:imagel5}
\end{subfigure}
\hfill
\begin{subfigure}{.3\linewidth}
    \centering
    \includegraphics[width=\linewidth]{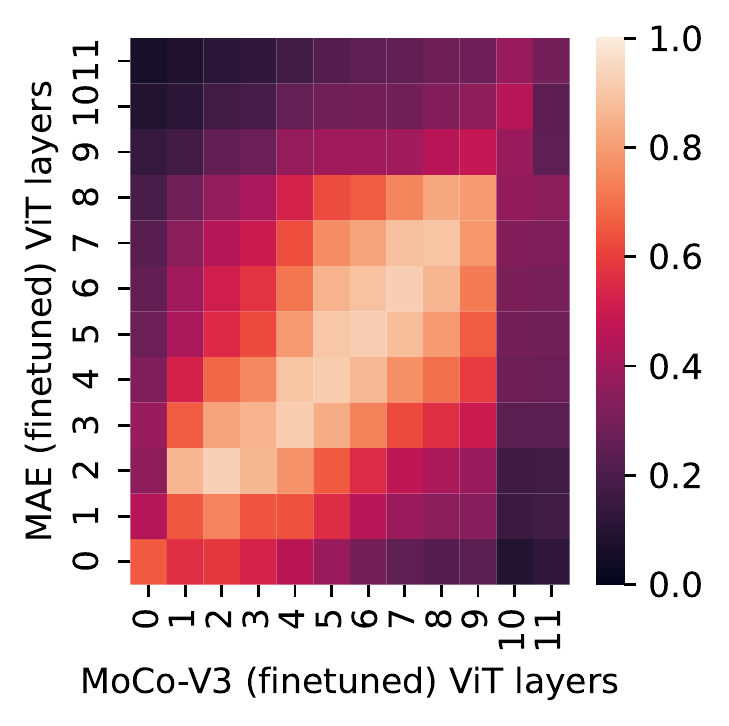}
    \caption{Fully-Connected (FT)}\label{fig:imagel6}
\end{subfigure}
\caption{CKA similarity between MoCo-V3 and MAE before and after fine-tuning by layer type.}
\label{fig:image-app}
\end{figure}

In addition to Fig. \ref{fig:imageft} we include additional comparisons of layer-wise CKA similarity between MoCo-V3 and MAE layers before and after fine-tuning in \ref{fig:image-app}. We can observed that the similarity between the fully-connected layers (MLP-FC1) increases for the initial and intermediate ViT layers but decreases for the later layers. However, the similarity between multi-head self-attention layers (MHSA-QKV) and layer normalization layers after attention (LayerNorm) of both models increases remarkably post fine-tuning. There is also a strong linear correspondence (layers at similar depth learn similar features) as well as strong block correspondence (groups of layers learn similar features) in the initial and intermediate MHSA-QKV and LayerNorm layers after fine-tuning.
%%%%%%%%%%%%%%%%%%%%%%%%%%%%%%%%%%%%%%%%%%%%%%%%%%%%%%%%%%%%%%%%%%%%%%%%%%%%%%%
%%%%%%%%%%%%%%%%%%%%%%%%%%%%%%%%%%%%%%%%%%%%%%%%%%%%%%%%%%%%%%%%%%%%%%%%%%%%%%%

\clearpage

\section{Rank Prediction Consistency across pre-training objectives}
\label{app:KTrank}

\begin{figure}[ht]
\centering
\includegraphics[width=\columnwidth]{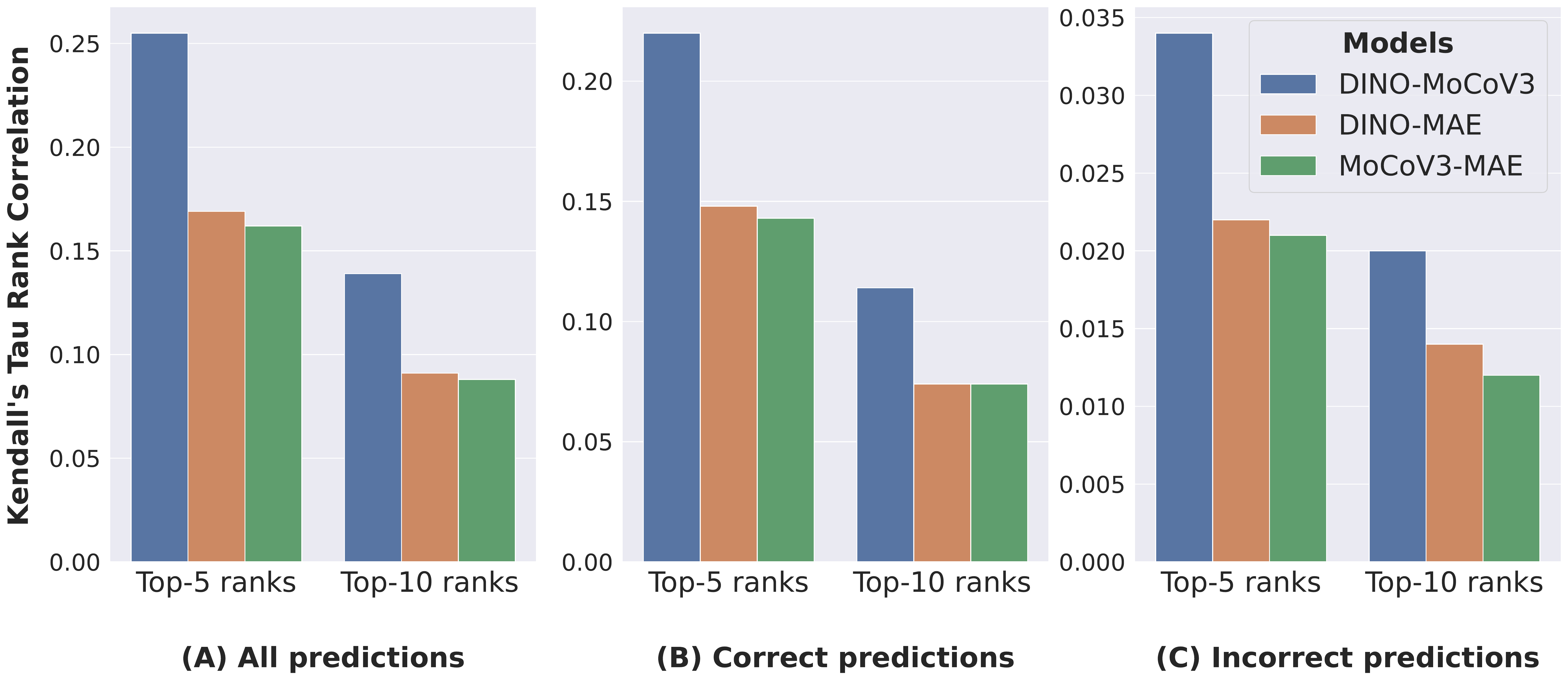}
\caption{Kendall's Tau rank correlation of linear probe ranks (Top-5 and Top-10 ranks averaged across ImageNet val set). \textbf{\contrastive models generate more similar rankings across all predictions } (5A), \textbf{and are also incorrect in similar ways} (5C).}
\label{fig:fig_kt}
\end{figure}

We consider whether the differences in representational similarity and in class separability across \contrastive and \reconstruction models translates to the class predictions made by these models. While we have observed higher linear and k-NN transfer performance in \contrastive models in Section \ref{sec:res-representation}, we do not know whether similar representations and performance are driven by consistent object classification results, or inconsistent results across different classes.  In order to evaluate this, we consider the Kendall's Tau rank correlation coefficient of the top-5 and top-10 class predictions made across the ImageNet validation set from MoCo-V3, DINO, and MAE in Fig. \ref{fig:fig_kt}. We observe that the ranking predictions generated by MoCo-V3 and DINO are consistently more correlated across all predictions, as well as both correct and incorrect predictions. We also calculate the F-1 score of top-1 predictions for DINO and MoCo-V3 (0.93) and confirm that it is higher than the F-1 score for MAE and MoCo-V3 (0.88). Our results verify that not only does the training objective determine representation content, similar pre-training objectives lead to features that represent different object classes similarly leading to class predictions which are also right and wrong in similar ways. 

\clearpage

\section{Comparisons of Reconstruction-Based and Joint-Embedding Learning with Masked Siamese Networks}
\label{app:msn}

We also compare the representations in \contrastive and \reconstruction ViTs to a training procedure that incorporates elements of both: Masked Siamese Networks (MSN) \citep{msn}. Masked Siamese Networks use a joint embedding approach similar to DINO \cite{caron2021emerging} as their objective, however they also sample input patches from the image like MAE \cite{he2022masked} for learning their anchor view embeddings in the joint embedding framework.

\begin{figure}[ht]
\centering
\begin{subfigure}{.3\linewidth}
    \centering
    \includegraphics[width=\linewidth]{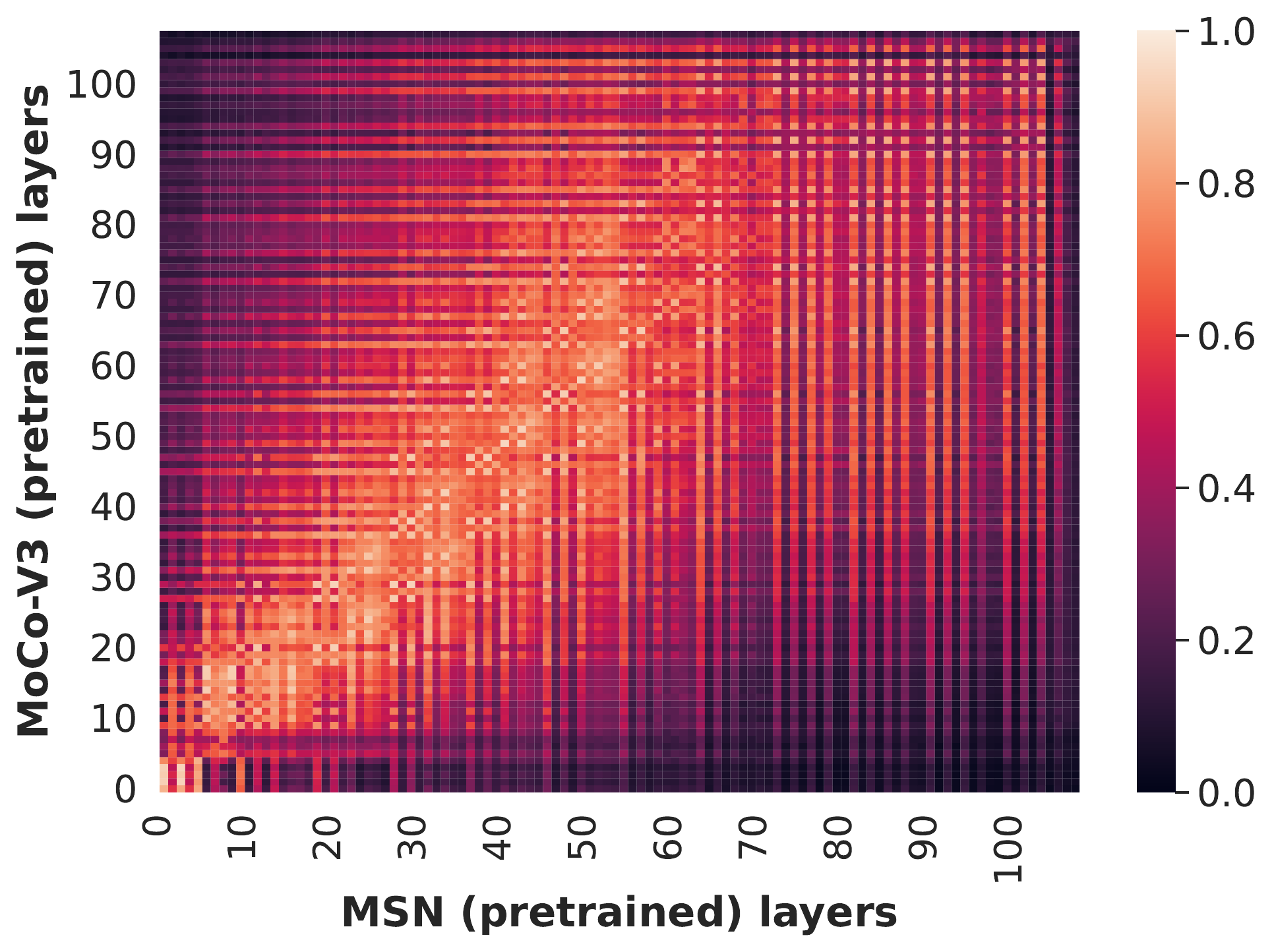}
    \caption{MoCo-V3 (PT) vs MSN (PT)}
    \label{fig:imagemsn1}
\end{subfigure}
    \hfill
\begin{subfigure}{.3\linewidth}
    \centering
    \includegraphics[width=\linewidth]{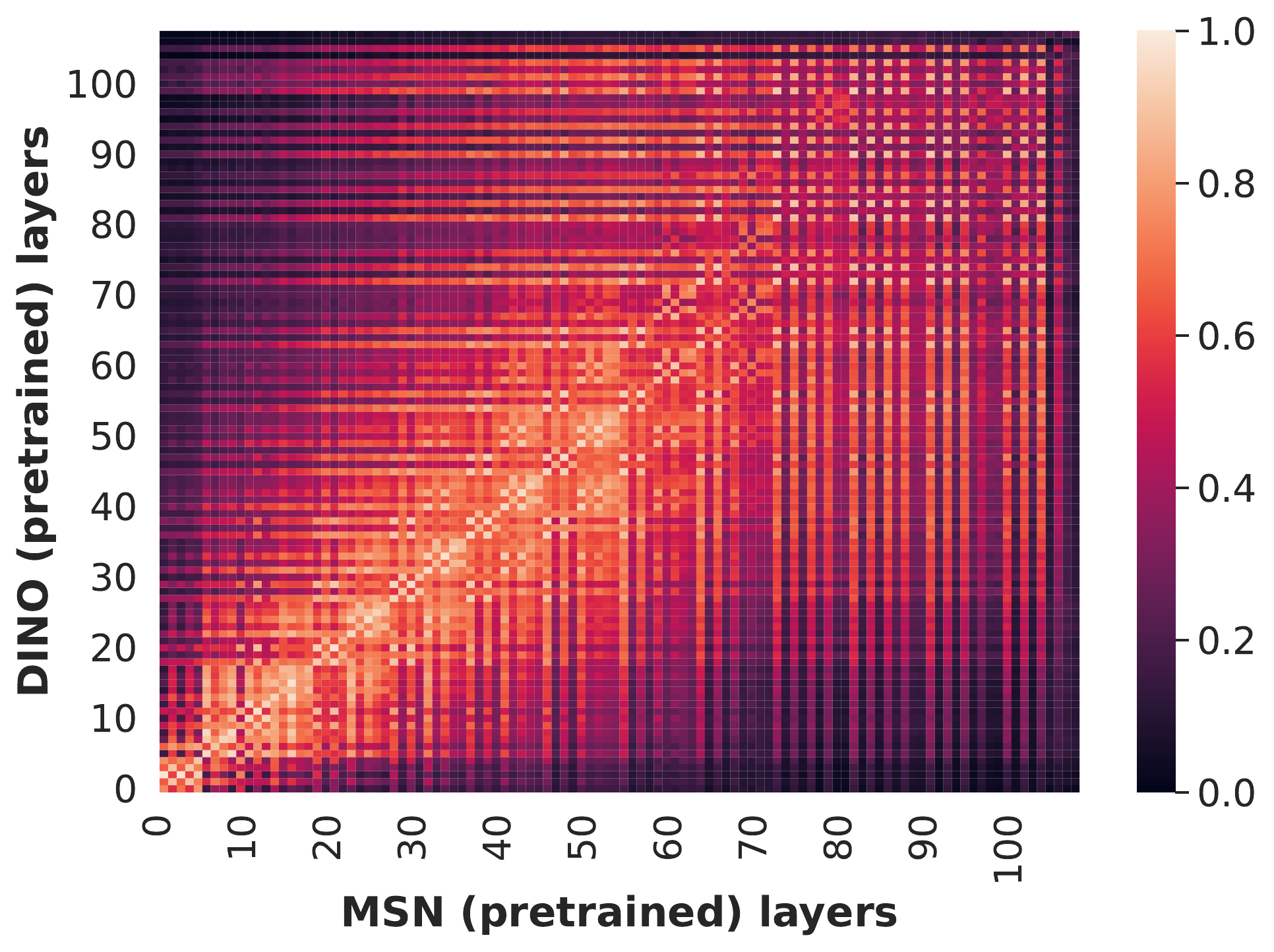}
    \caption{DINO (PT) vs MSN (PT)}\label{fig:imagemsn2}
\end{subfigure}
   \hfill
\begin{subfigure}{.3\linewidth}
    \centering
    \includegraphics[width=\linewidth]{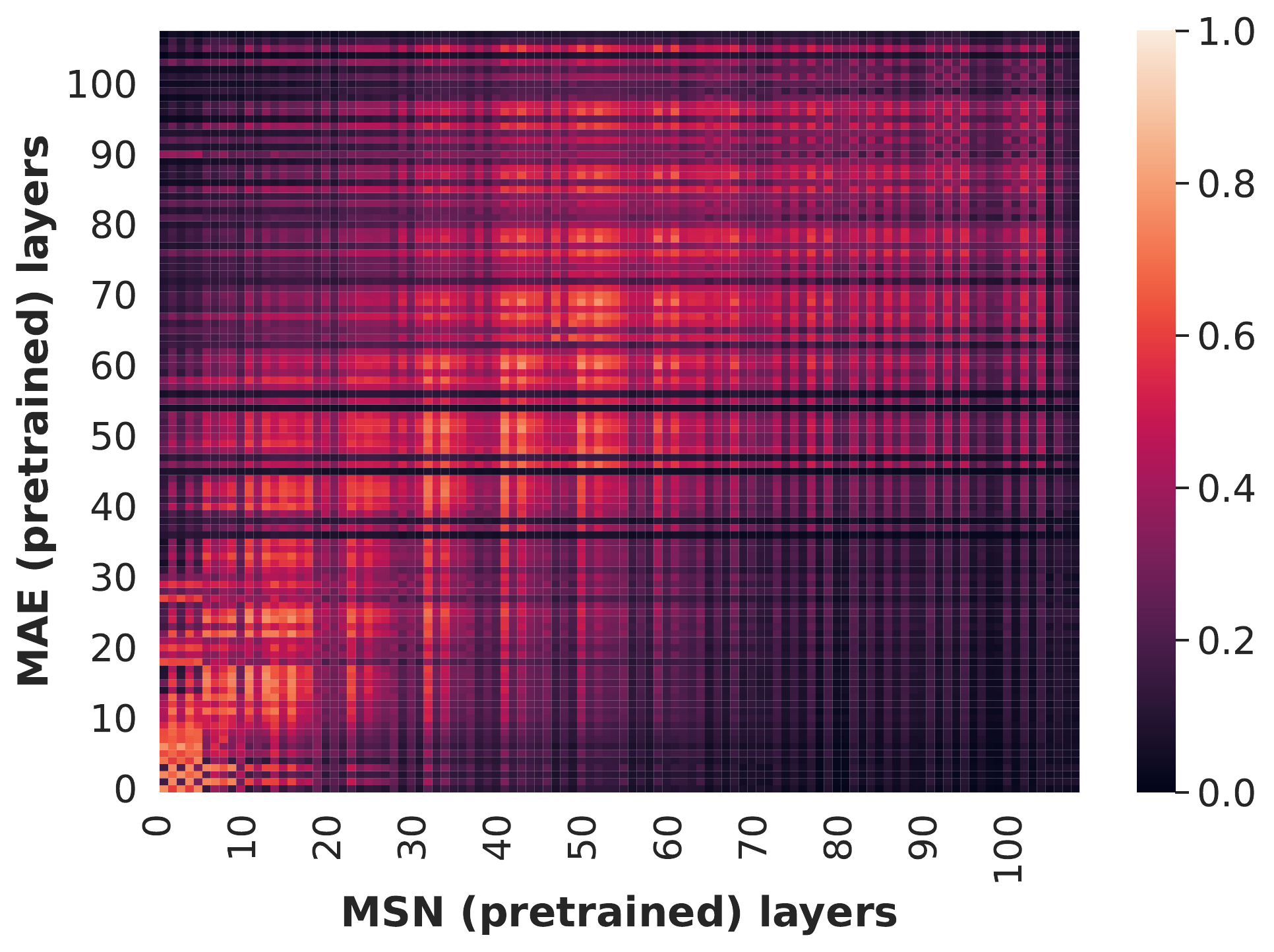}
    \caption{MAE (PT) vs MSN (PT)}\label{fig:imagemsn3}
\end{subfigure}
\begin{subfigure}{.3\linewidth}
    \centering
    \includegraphics[width=\linewidth]{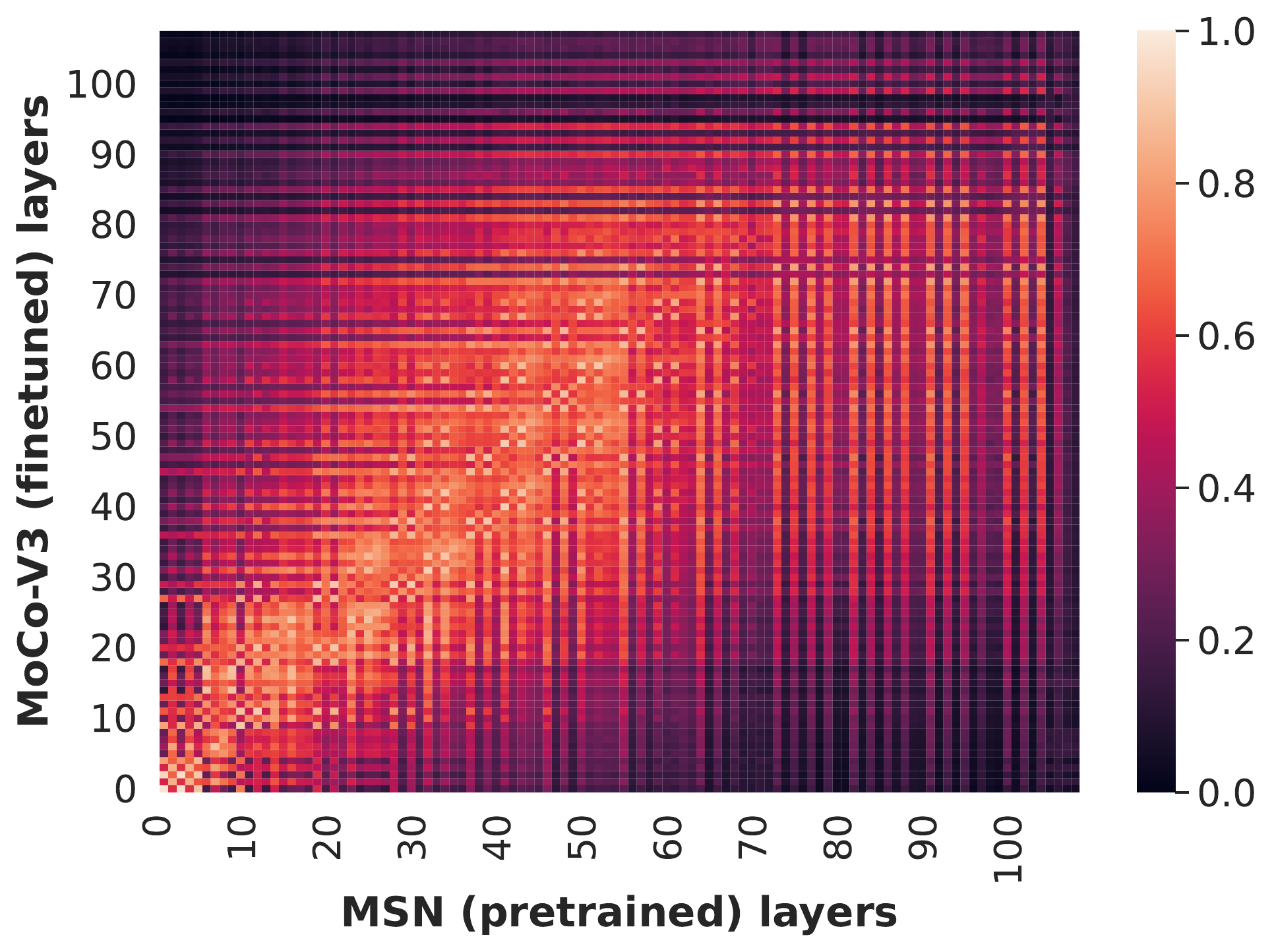}
    \caption{MoCo-V3 (FT) vs MSN (PT)}
    \label{fig:imagemsn4}
\end{subfigure}
    \hfill
\begin{subfigure}{.3\linewidth}
    \centering
    \includegraphics[width=\linewidth]{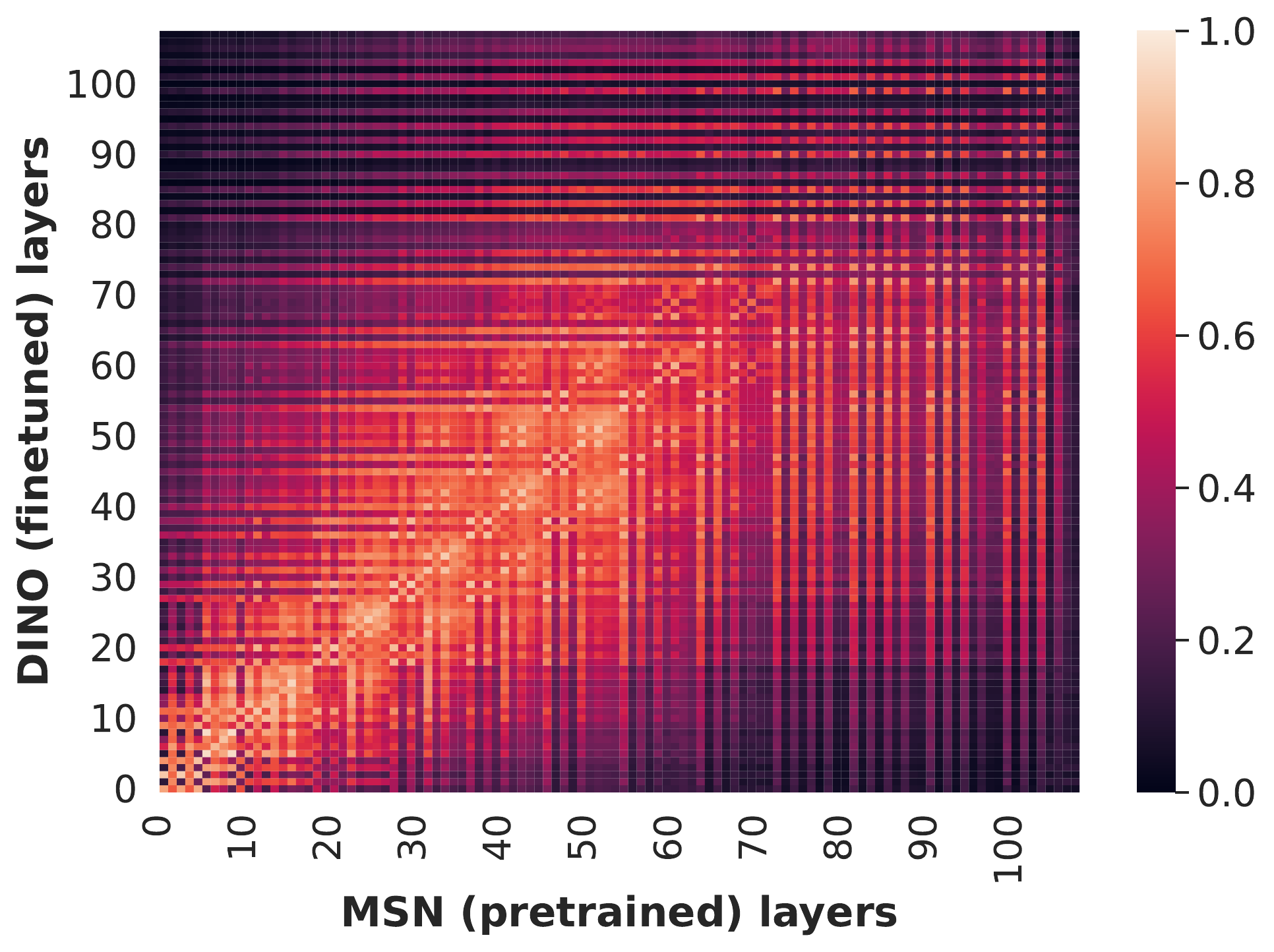}
    \caption{DINO (FT) vs MSN (PT)}
    \label{fig:imagemsn5}
\end{subfigure}
\hfill
\begin{subfigure}{.3\linewidth}
    \includegraphics[width=\linewidth]{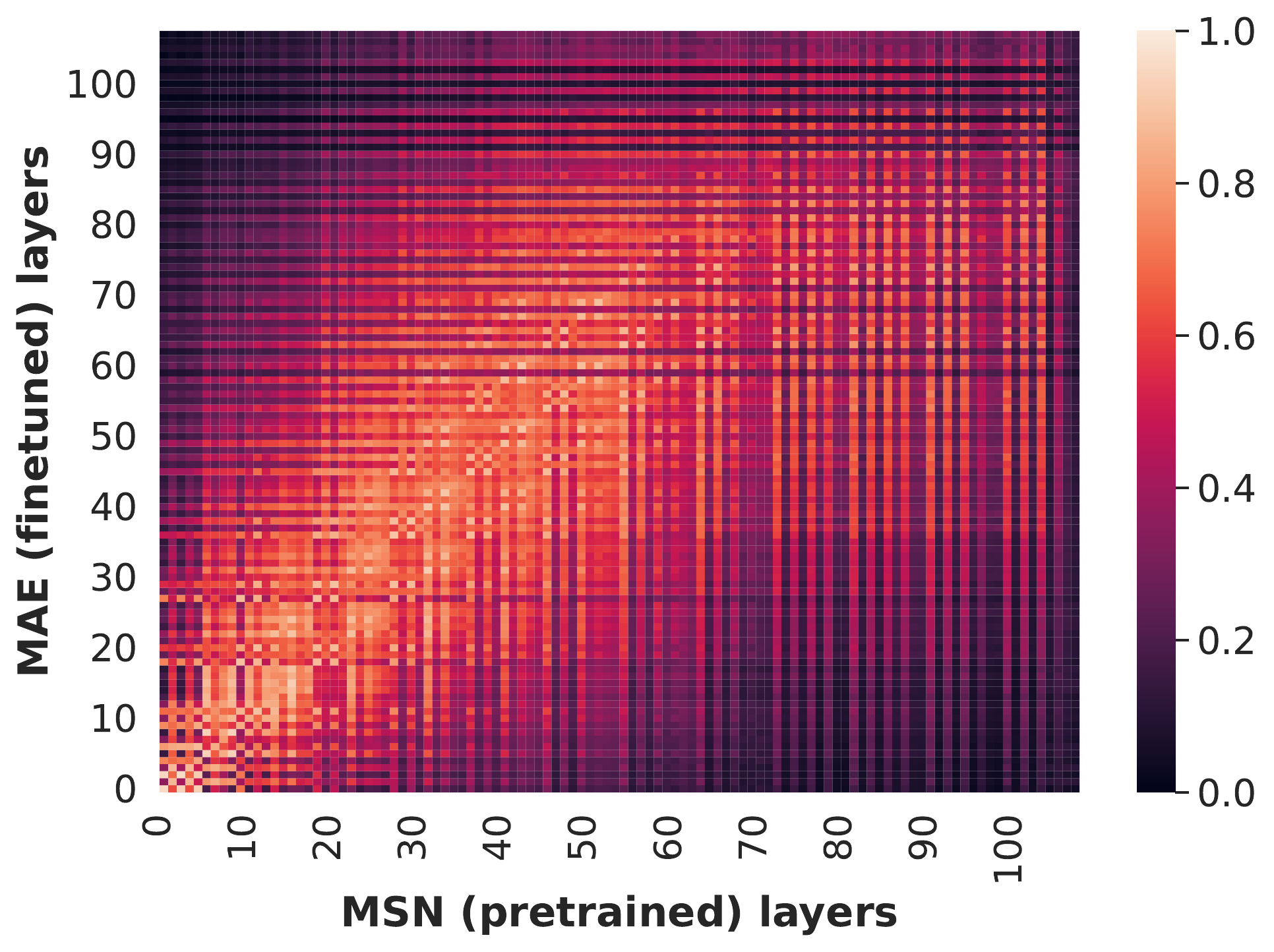}
    \caption{MAE (FT) vs MSN (PT)}
    \label{fig:imagemsn6}
\end{subfigure}
\caption{CKA similarity between pre-trained and fine-tuned \contrastive and \reconstruction models and a pre-trained MSN model.}
\label{fig:imagemsn}
\end{figure}

We hypothesize that since the training objective of MSN does not invoke reconstruction-based losses, the representations learned will be similar to joint-embedding approaches despite their use of masking-based feature learning. Indeed, our representation similarity analysis in Figure \ref{fig:imagemsn} shows that pre-trained MSN representations are much more similar to pre-trained \contrastive representations (DINO, MoCo-V3) than \reconstruction representation (MAE). Thus, we conclude that the reconstruction based objective plays a much stronger role in the features learned by MAE versus the modelling of masked image features which MSN shares with MAEs, while the \contrastive objective of modelling similarity between pairs of views dominates features learned by MSN.

\begin{figure}[ht]
\centering
\begin{subfigure}{.3\linewidth}
    \centering
    \includegraphics[width=\linewidth]{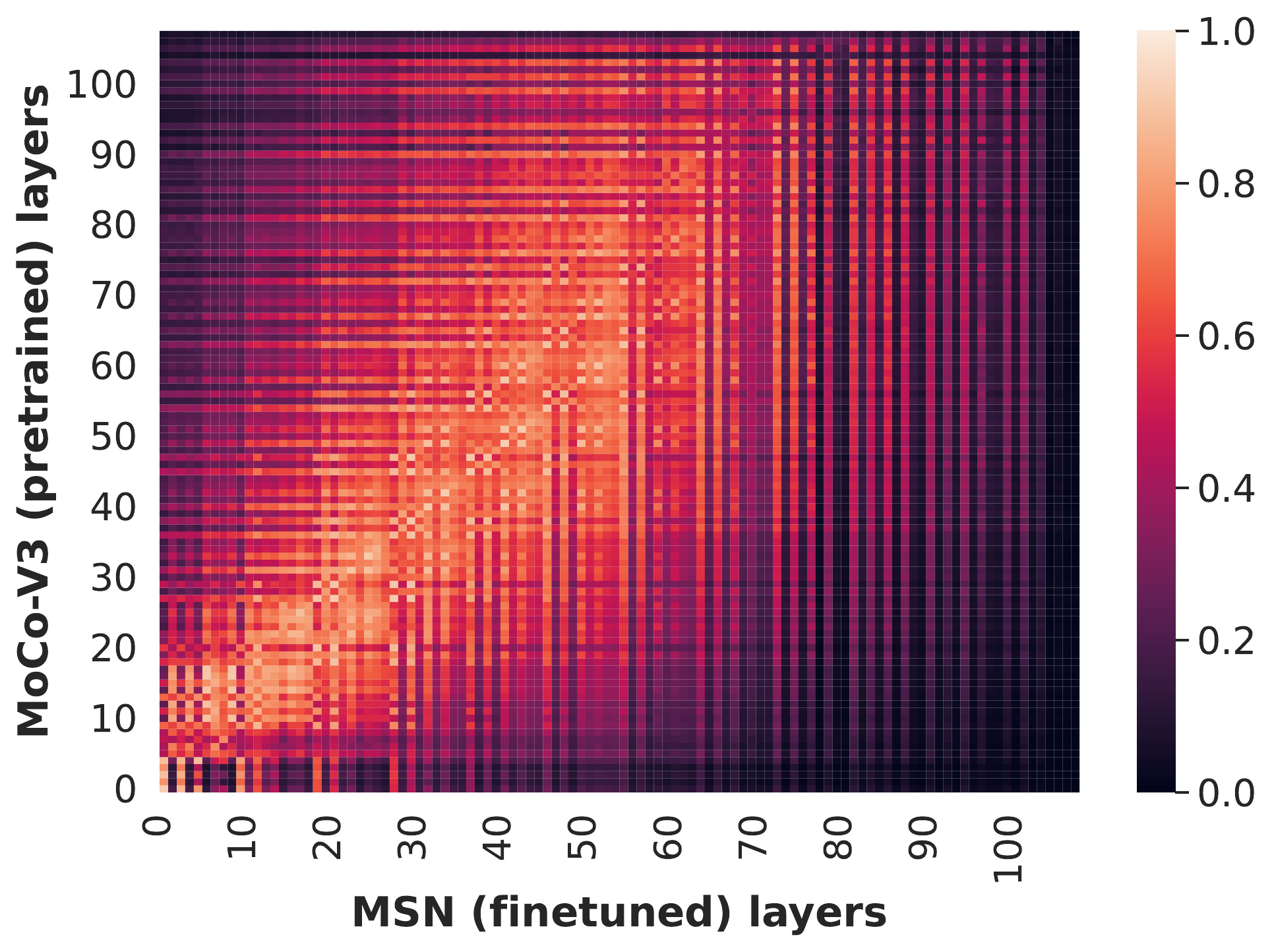}
    \caption{MoCo-V3 (PT) vs MSN (FT)}
    \label{fig:imagemsnft1}
\end{subfigure}
    \hfill
\begin{subfigure}{.3\linewidth}
    \centering
    \includegraphics[width=\linewidth]{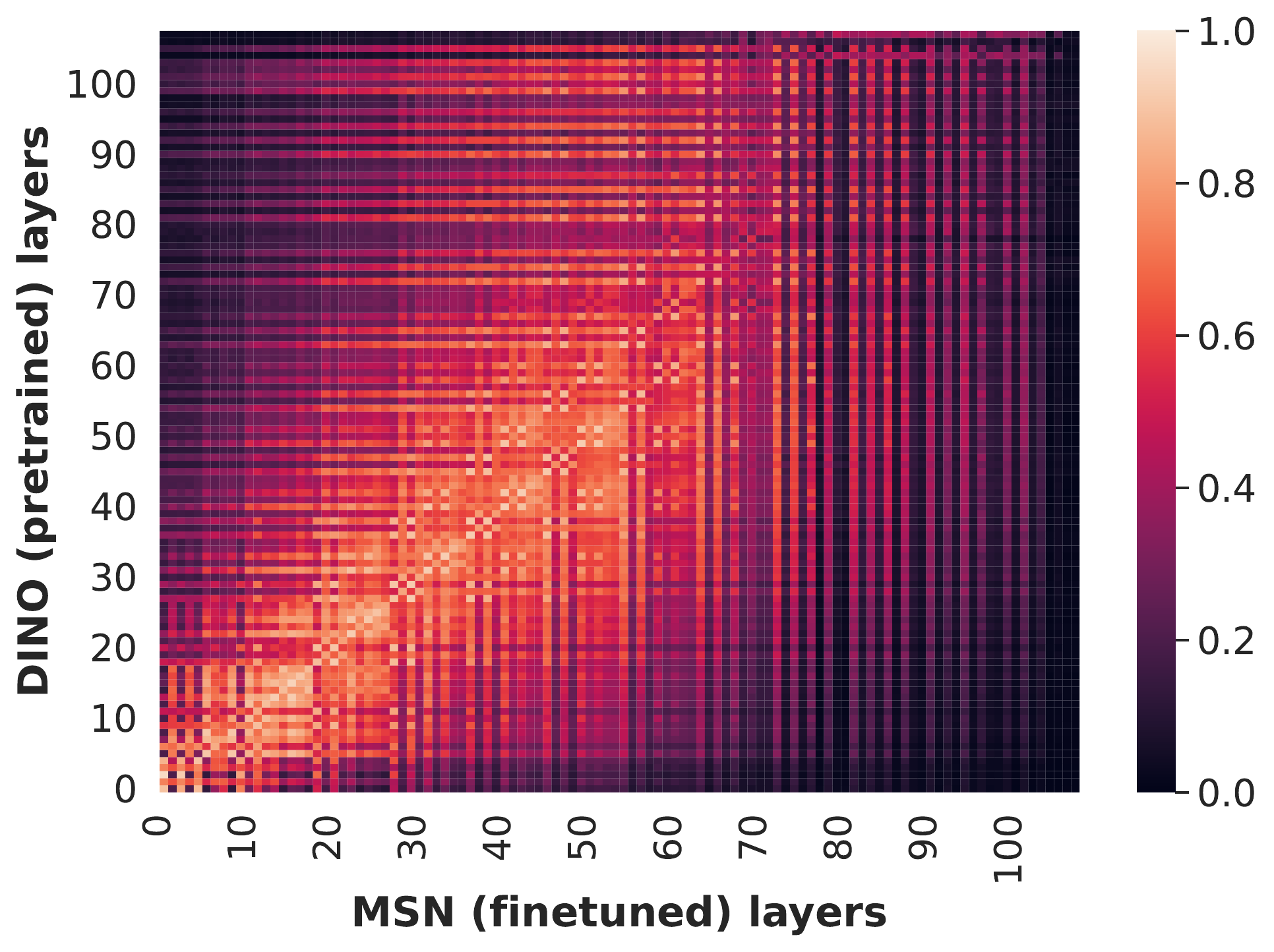}
    \caption{DINO (PT) vs MSN (FT)}\label{fig:imagemsnft2}
\end{subfigure}
   \hfill
\begin{subfigure}{.3\linewidth}
    \centering
    \includegraphics[width=\linewidth]{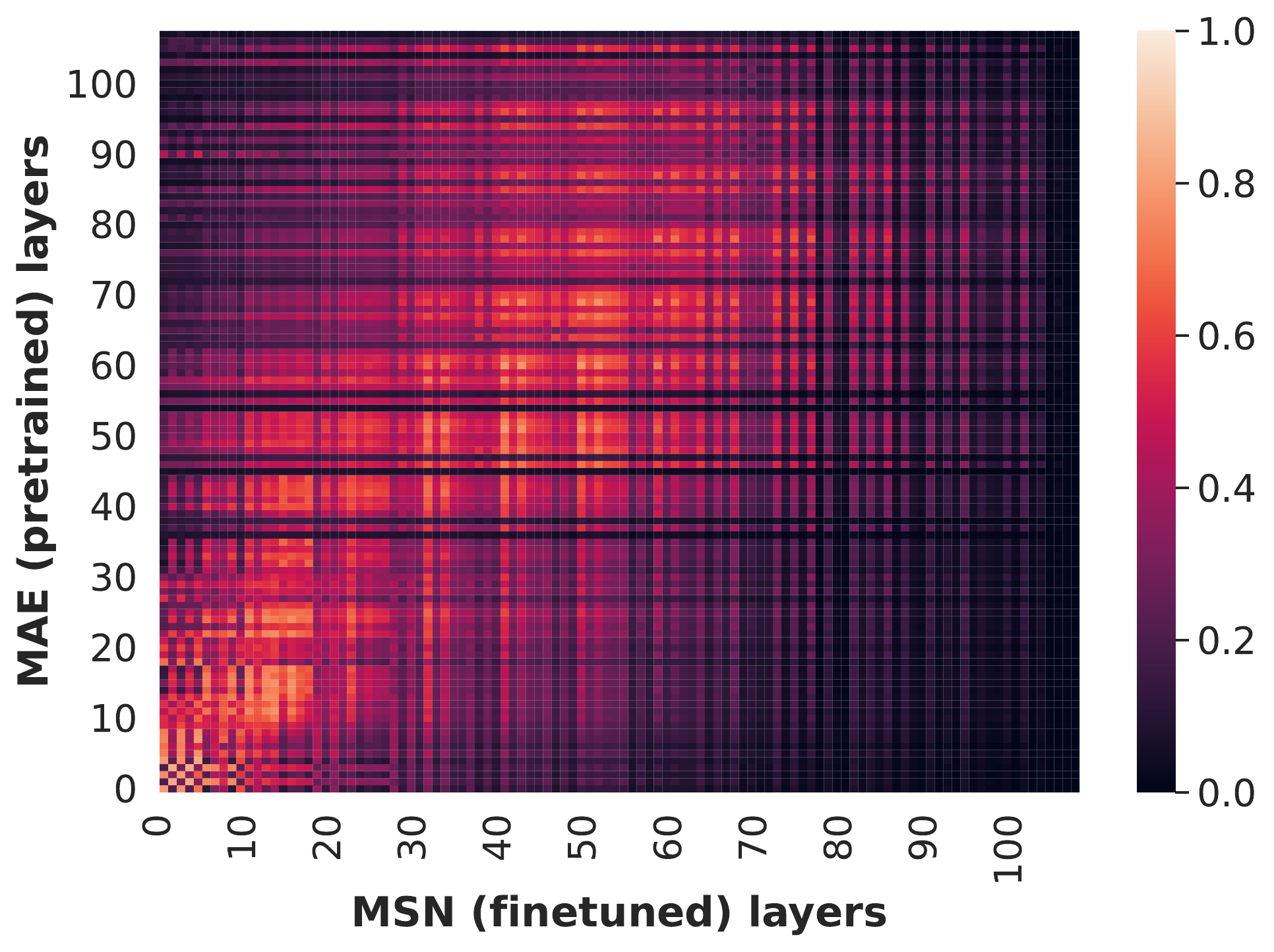}
    \caption{MAE (PT) vs MSN (FT)}\label{fig:imagemsnft3}
\end{subfigure}
\begin{subfigure}{.3\linewidth}
    \centering
    \includegraphics[width=\linewidth]{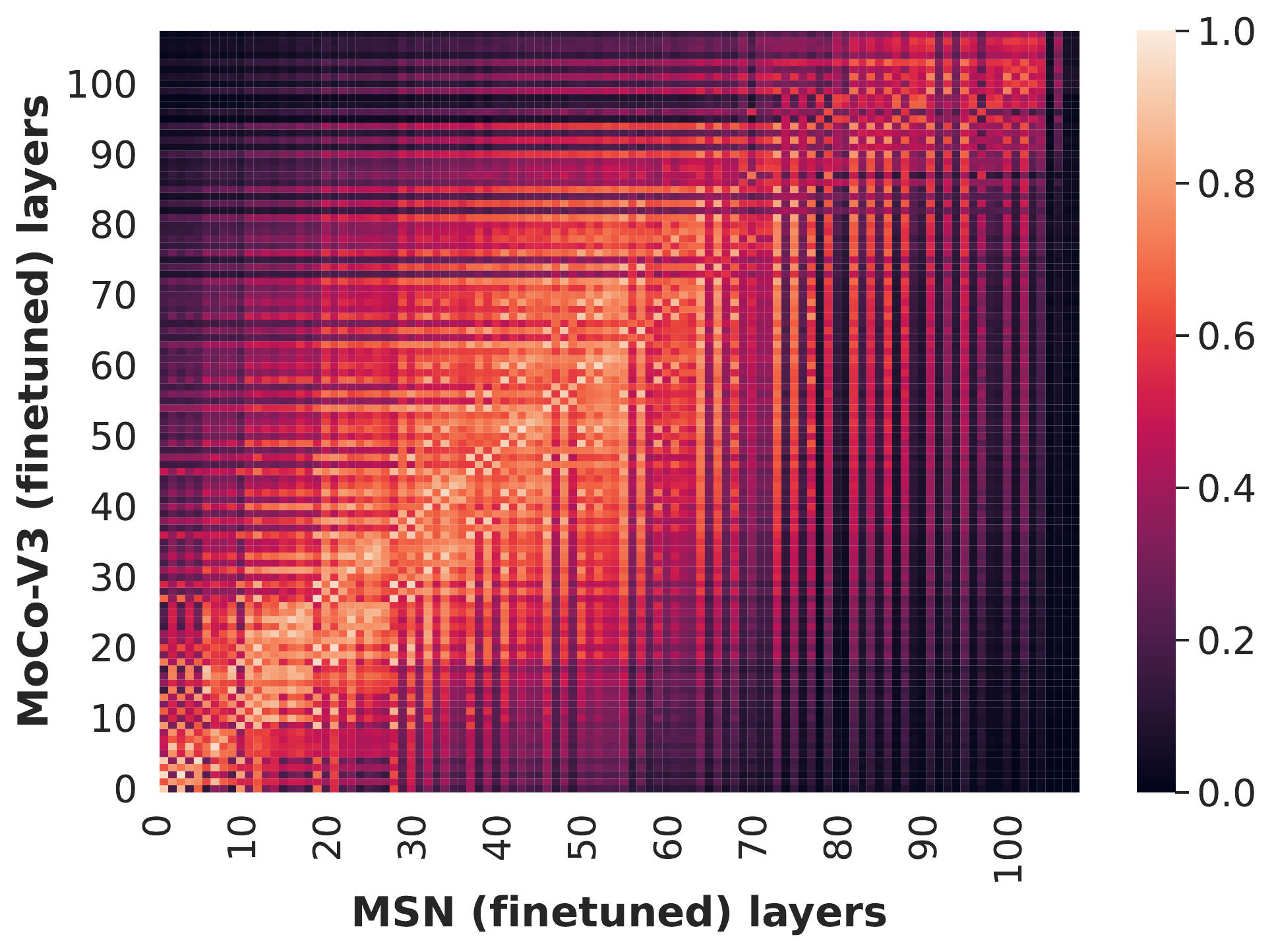}
    \caption{MoCo-V3 (FT) vs MSN (FT)}
    \label{fig:imagemsnft4}
\end{subfigure}
    \hfill
\begin{subfigure}{.3\linewidth}
    \centering
    \includegraphics[width=\linewidth]{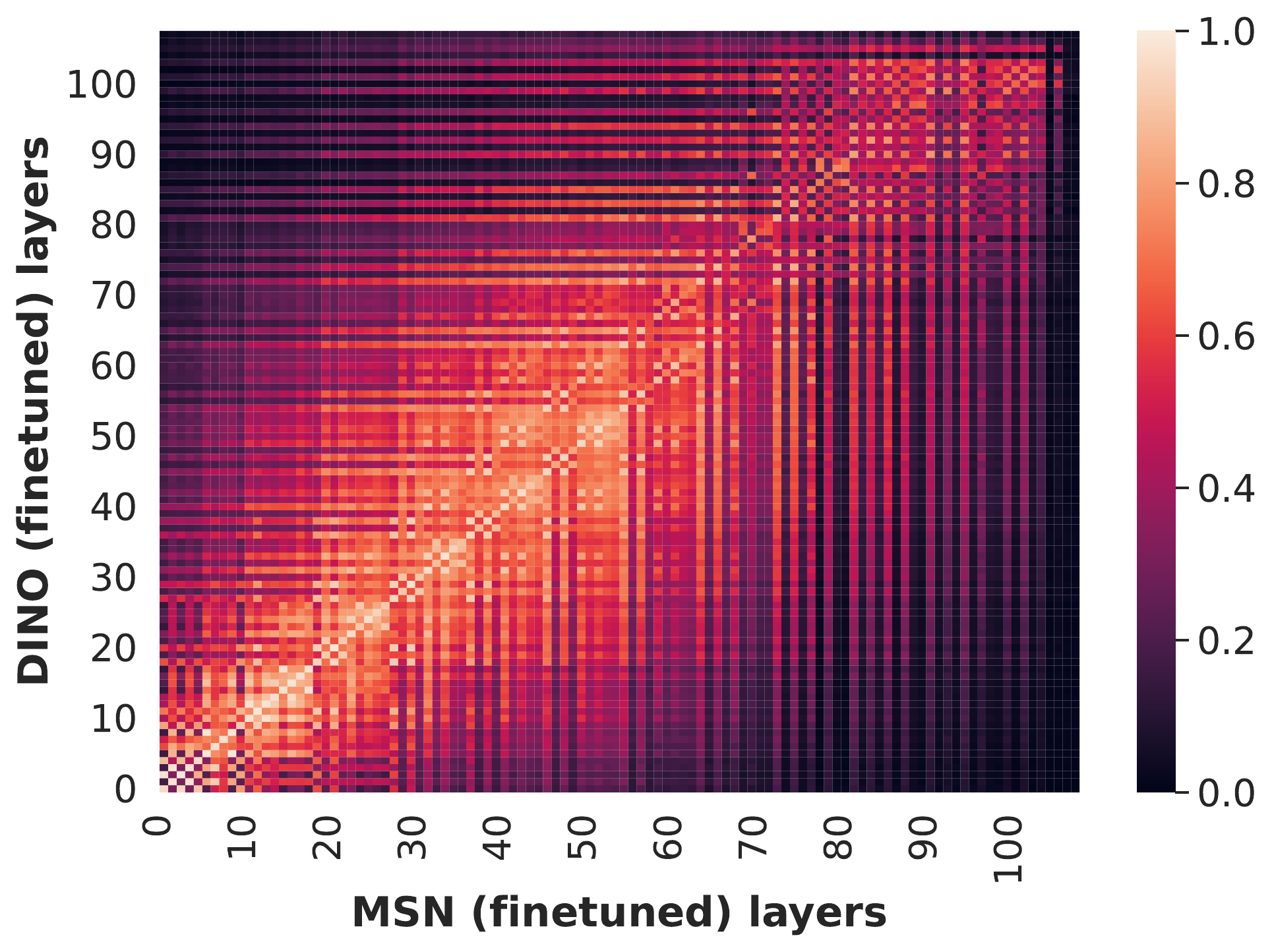}
    \caption{DINO (FT) vs MSN (FT)}
    \label{fig:imagemsnft5}
\end{subfigure}
\hfill
\begin{subfigure}{.3\linewidth}
    \centering
    \includegraphics[width=\linewidth]{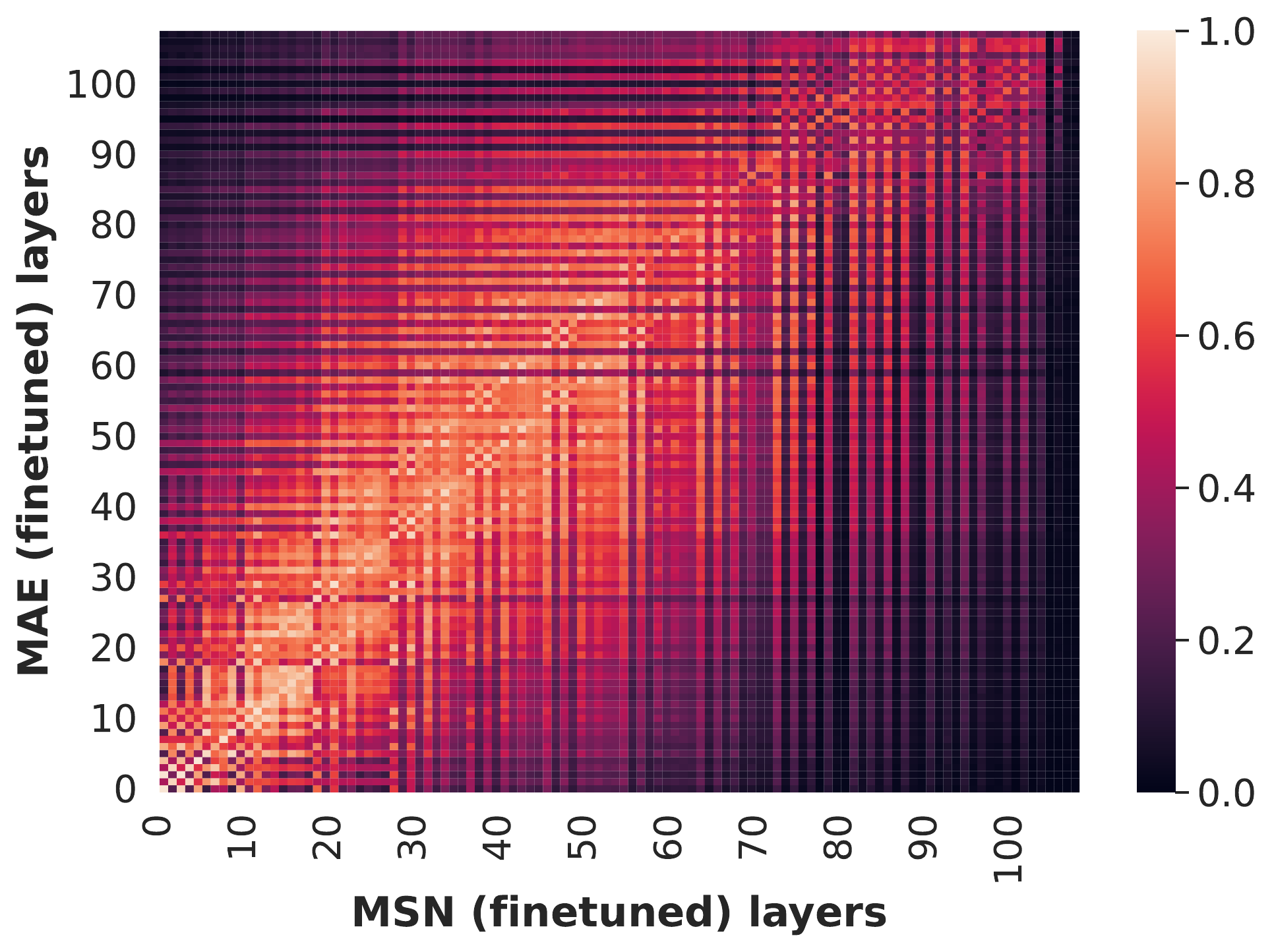}
    \caption{MAE (FT) vs MSN (FT)}
    \label{fig:imagemsnft6}
\end{subfigure}
\caption{CKA similarity between pre-trained and fine-tuned \contrastive and \reconstruction models and a fine-tuned MSN model.}
\label{fig:imagemsnft}
\end{figure}

Fine-tuning ViTs pre-trained with MSN also gives results consistent with fine-tuning \contrastive models as outlined in Section \ref{sec:res-ft}. Fine-tuned MSN models continue to remain similar to pre-trained and fine-tuned \contrastive ViTs, as well as fine-tuned \reconstruction ViTs, as shown in in Figure \ref{fig:imagemsnft}.

\clearpage

\section{Additional RCDM examples} \label{sec:app-rcdm}

%\ari{4x2 PT and FT RCDM for all four models, }
We visualize RCDM\footnote{We train RCDM on the face-blurred version of ImageNet \citep{yang2021imagenetfaces}, which enhances privacy.\label{refnote2}} samples conditioned on a fine-tuned MAE model below to demonstrate that fine-tuning imparts new invariances (like horizontal flip invariance) that improve transfer performance on classification.

\begin{figure}[ht]

\begin{subfigure}{0.49\linewidth}
    \centering
   \includegraphics[width=\columnwidth]{figures/ICML/RCDM_mae2.pdf}
    \caption{MAE (PT)}
\end{subfigure}
    \hfill
\begin{subfigure}{0.49\linewidth}
    \centering
   \includegraphics[width=\columnwidth]{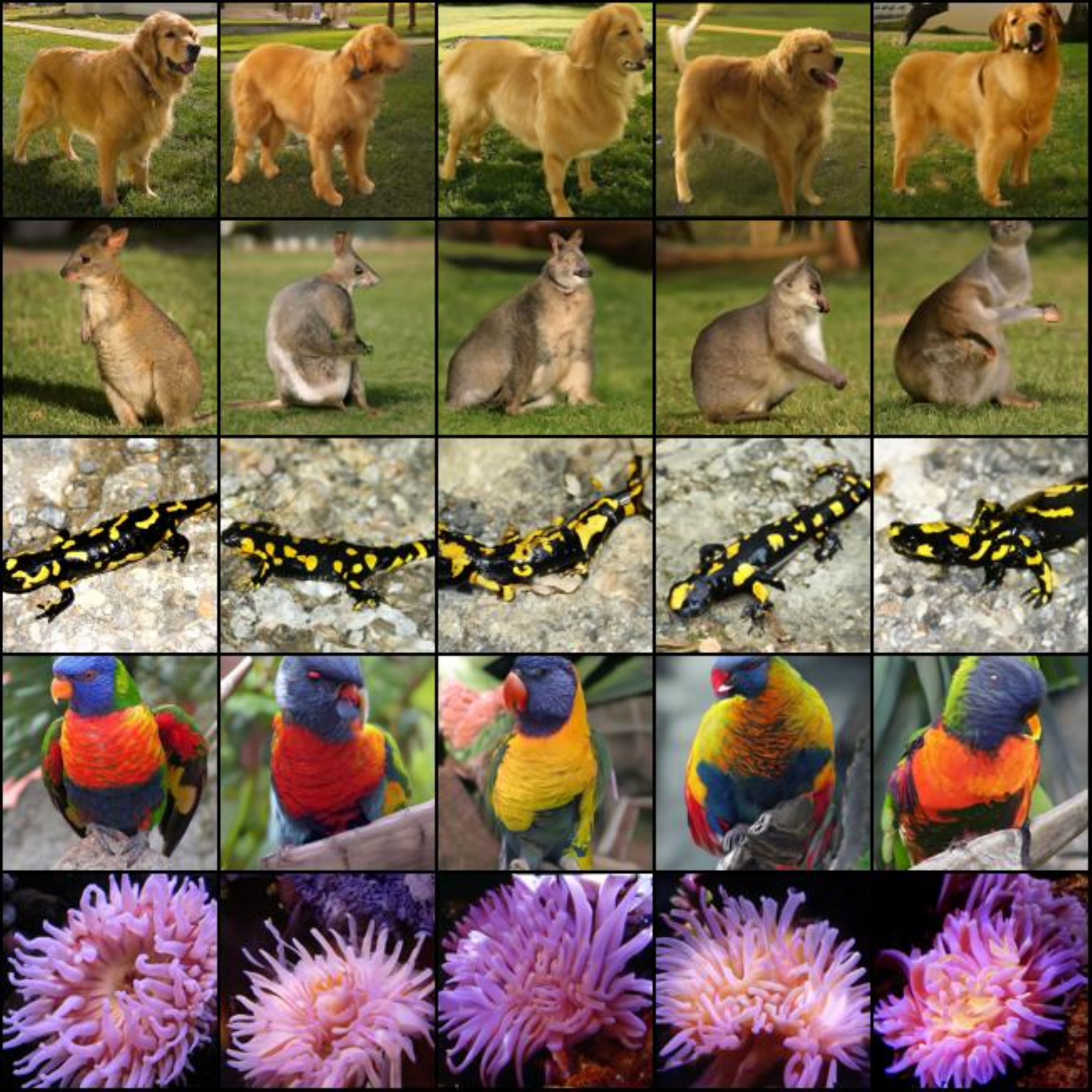}
    \caption{MAE (FT)}
\end{subfigure}
\caption{Visualization of samples generated from an RCDM\textsuperscript{\ref{refnote2}} conditioned on fine-tuned MAE and trained on the face-blurred version of ImageNet \citep{yang2021imagenetfaces}. \textbf{Unlike pre-trained MAEs, fine-tuned MAE features generate objects oriented differently (horizontally) to the source image, demonstrating that these features are invariant to horizontal flips.}}
\label{fig:rcdm_mae_ft}
\end{figure}

We also provide additional RCDM samples visualized for each of the 4 models (Pretrained: MAE, DINO, MoCo-V3 and Finetuned: MAE) for readers to identify additional invariances.

\begin{figure}
\centering
\begin{subfigure}[t]{0.48\linewidth}
    \centering
    \includegraphics[width=\linewidth]{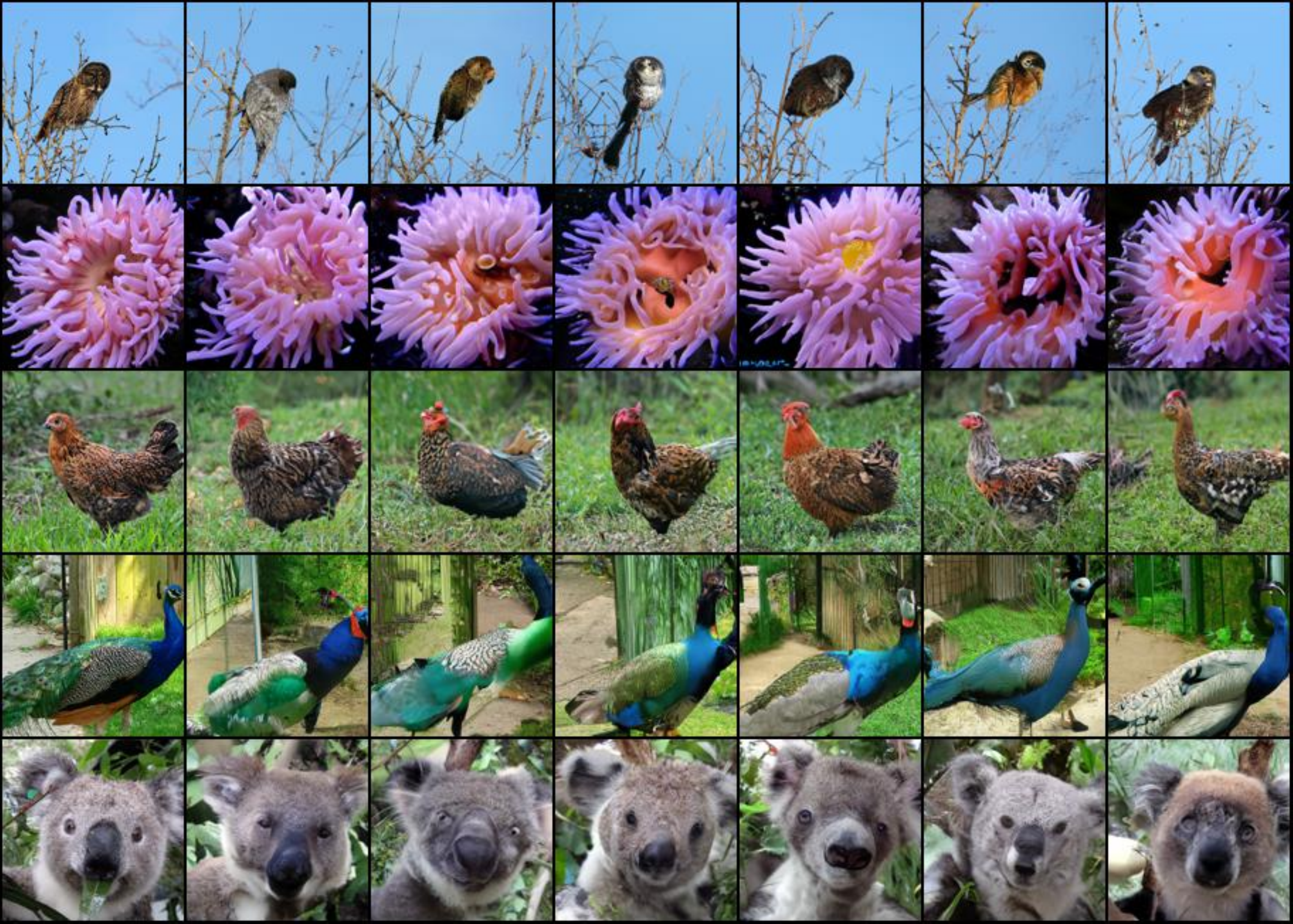}
    \caption{MAE (PT)}
    \label{fig:rcdm-extra-1}
\end{subfigure}
\hfill
\begin{subfigure}[t]{0.48\linewidth}
    \centering
    \includegraphics[width=\linewidth]{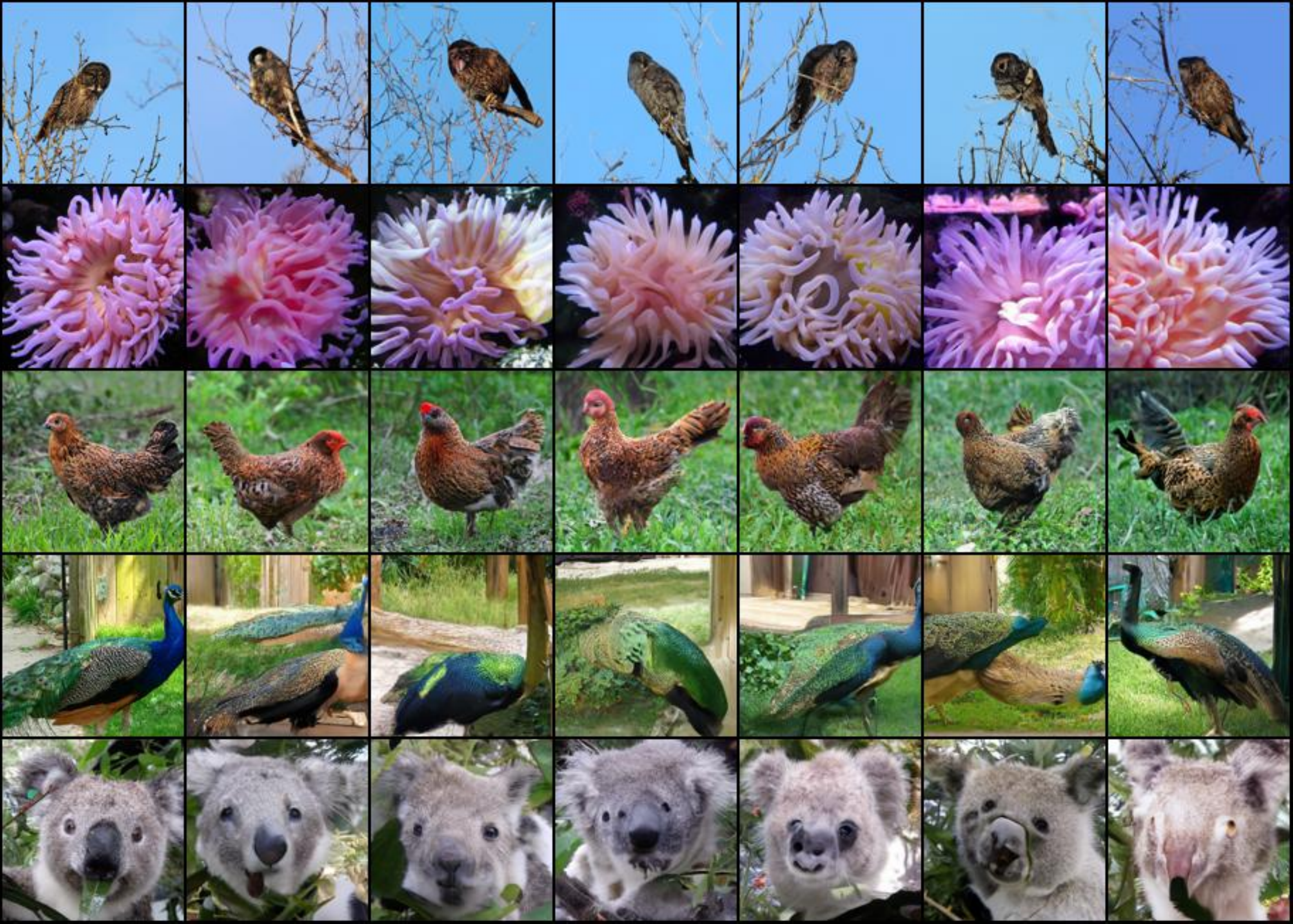}
    \caption{MoCo-V3 (PT)}
    \label{fig:rcdm-extra-2}
\end{subfigure}
\begin{subfigure}[t]{0.48\linewidth}
    \centering
    \includegraphics[width=\linewidth]{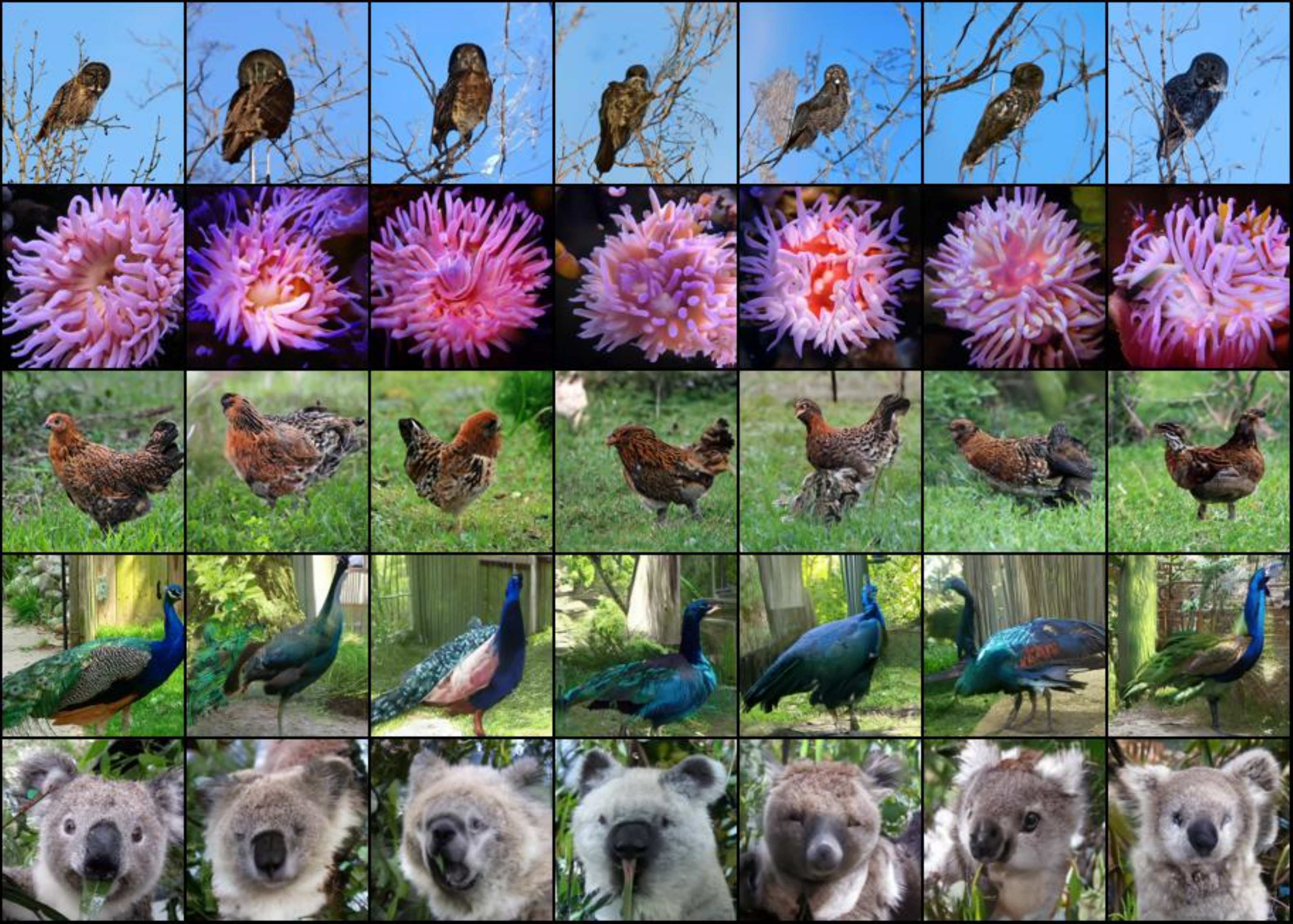}
    \caption{MAE (FT)}
    \label{fig:rcdm-extra-3}
\end{subfigure}
\hfill
\begin{subfigure}[t]{0.48\linewidth}
    \centering
    \includegraphics[width=\linewidth]{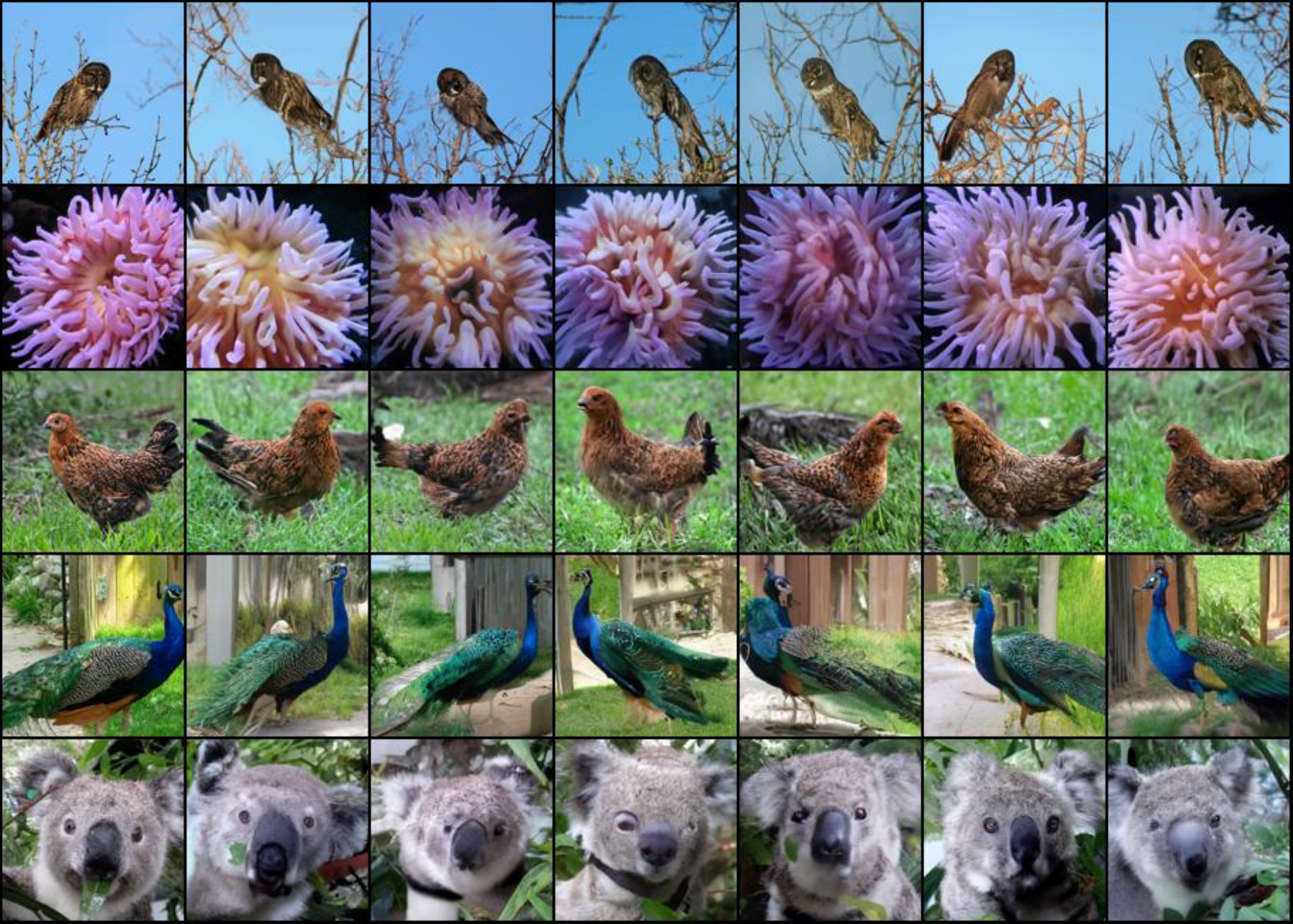}
    \caption{DINO (PT)}
    \label{fig:rcdm-extra-4}
\end{subfigure}
\caption{Additional RCDM\textsuperscript{\ref{refnote2}} samples generated using representations from pre-trained MAE, DINO, MoCo-V3 and fine-tuned MAE when training RCDM on the face-blurred version of ImageNet \citep{yang2021imagenetfaces}.}
\label{fig:rcdm-extra}
\end{figure}

\end{document}